\documentclass[10pt,twocolumn,letterpaper]{article}

\usepackage[pagenumbers]{wacv} %

\usepackage{siunitx}
\usepackage[table]{xcolor}
\usepackage{multirow} %
\usepackage{booktabs}       %
\usepackage{colortbl}       %
\usepackage{pifont} %
\usepackage{pdflscape}

\definecolor{best_blue}{HTML}{749ECE}
\definecolor{worst_red}{HTML}{FF8080}

\definecolor{fig_blue}{HTML}{1f77b4}
\definecolor{fig_orange}{HTML}{ff7f0e}

\definecolor{markred}{HTML}{CD0f01} %
\definecolor{markgreen}{HTML}{008836} %
\newcommand{\cmark}{\textcolor{markgreen}{\ding{51}}}
\newcommand{\xmark}{\textcolor{markred}{\ding{55}}} %

\definecolor{wacvblue}{rgb}{0.21,0.49,0.74}
\usepackage[pagebackref,breaklinks,colorlinks,allcolors=wacvblue]{hyperref}

\def\confName{WACV}
\def\confYear{2026}

\title{ICONIC-444: A 3.1-Million-Image Dataset for OOD Detection Research}

\author{
Gerhard Krumpl\thanks{Correspondence: \tt gerhard.krumpl@icg.tugraz.at} \textsuperscript{$\,$1,2} \qquad
Henning Avenhaus\textsuperscript{2} \qquad 
Horst Possegger\textsuperscript{1} \\ [0.4em]
\textsuperscript{1}Institute of Visual Computing, Graz University of Technology, Austria\\
\textsuperscript{2}KESTRELEYE GmbH, Austria\\
}

\begin{document}
\maketitle

\begin{abstract}
Current progress in out-of-distribution (OOD) detection is limited by the lack of large, high-quality datasets with clearly defined OOD categories across varying difficulty levels (near- to far-OOD) that support both fine- and coarse-grained computer vision tasks.
To address this limitation, we introduce ICONIC-444 (Image Classification and OOD Detection with Numerous Intricate Complexities), a specialized large-scale industrial image dataset containing over 3.1 million RGB images spanning 444 classes tailored for OOD detection research.
Captured with a prototype industrial sorting machine, ICONIC-444 closely mimics real-world tasks.
It complements existing datasets by offering structured, diverse data suited for rigorous OOD evaluation across a spectrum of task complexities. 
We define four reference tasks within ICONIC-444 to benchmark and advance OOD detection research and provide baseline results for 22 state-of-the-art post-hoc OOD detection methods.

\end{abstract}

\section{Introduction}

\begin{figure}[t]
\centering
\includegraphics[width=\linewidth]{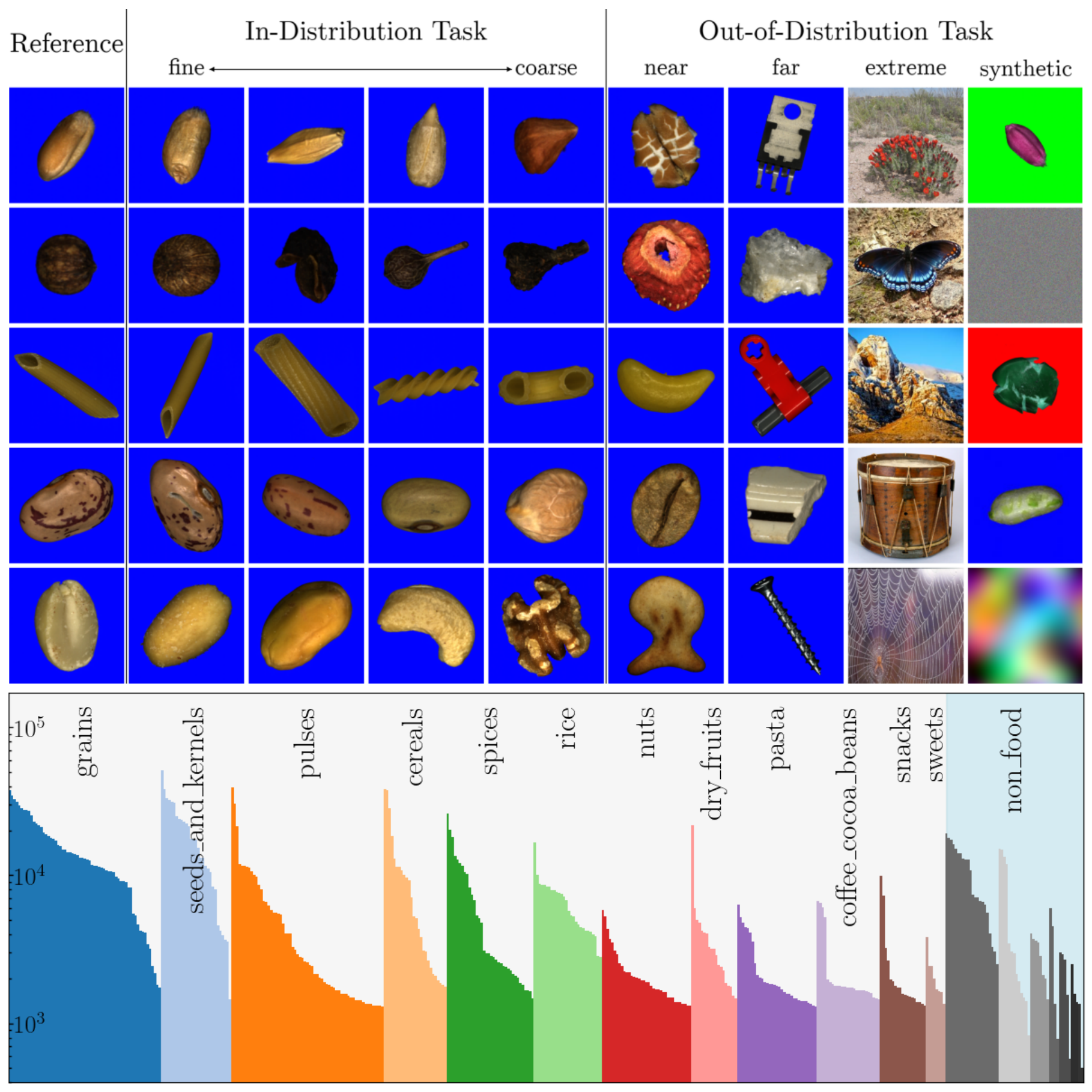}
\caption{
Overview of ICONIC-444 and its application in creating tasks for image classification and OOD detection evaluation.
\textbf{Top panel:} An image grid illustrating the progression from fine-grained to coarse-grained ID classes alongside OOD samples of varying difficulty levels, near, far, extreme (from external datasets such as ImageNet~\cite{deng2009_imagenet}, iNaturalist~\cite{Horn_2018_inat}, Places365~\cite{zhou2017places}, and Textures~\cite{cimpoi14dtd}), as well as synthetic images, providing diverse OOD detection benchmarks.
\textbf{Bottom panel:} A histogram on a logarithmic scale showing the distribution of images per class.
Classes are organized left to right by category (Food, Non-Food), group size (total images per group), and then by the number of images per class within each group.
}
\label{fig:teaser}
\end{figure}

In real-world applications, machine learning (ML) models frequently encounter unexpected inputs that differ from their training data, such as entirely new object classes, corrupted data, sensor failures, or insufficient illumination.
These out-of-distribution (OOD) samples can cause deep learning models to make highly confident yet incorrect predictions of in-distribution (ID) classes, thereby undermining both the reliability of and the trust in the system~\cite{Nguyen2014, moosavidezfooli17, hein2019relu}. 
The consequences of such errors range from minor inconveniences to critical failures---for instance, misclassifying foreign materials as good products in food sorting or misidentifying objects in autonomous driving. 
Therefore, detecting and properly managing OOD samples is crucial to extend the usability and safety of ML models in real-world applications \cite{hendrycks2021_ml_safety, hendrycks2022_ml_safety, mohseni2021_ml_safety}.

While the specific actions taken when handling OOD inputs---such as entering a failure state, requesting human intervention, or preventing incorrect classifications---depend on the application, the fundamental task of detecting OOD inputs in the first place is currently unsolved.
Existing OOD detection methods have not yet achieved the reliability required for routine deployment in real-world scenarios~\cite{zhang2023openood, yang2022openood, hendrycks2021_ml_safety}.
In OOD research, a key barrier is that the most commonly used datasets were not designed with OOD detection in mind.

Standard evaluation protocols for OOD detection, \eg~\cite{zhang2023openood, hendrycks2016_ood_msp, sun2021_ood_react, wang2022_ood_vim}, usually train models on ID datasets (or subsets) and test them on OOD samples from different datasets (or excluded classes).
Widely-used datasets include CIFAR~\cite{krizhevsky09cifar}, ImageNet~\cite{deng2009_imagenet}, and fine-grained visual classification datasets like FGVC-Aircraft~\cite{maji13fine-grained} and CUB-200~\cite{WahCUB_200_2011}.
While instrumental in establishing OOD detection research, these benchmarks have substantial limitations, including contamination in OOD test sets~\cite{bitterwolf2023_ninco}, limited scale, and insufficient granularity to effectively test OOD methods' robustness.
Additionally, these datasets fail to replicate the conditions and complexities of real-world industrial environments, making them ill-suited for evaluating the applicability of OOD detection methods in such settings.

We propose the industrial object sorting task as an ideal real-world scenario for advancing OOD detection methods, combining a controlled setting with the task's inherent complexity.
This environment can provide both fine- and coarse-grained classification problems with varying numbers of ID classes and a broad spectrum of OOD samples, from subtle variations to entirely unrelated and visually distinct images (see \cref{fig:teaser}). 
ICONIC-444 addresses the aforementioned issues with current OOD evaluation protocols and provides a strong foundation for OOD detection research and the statistical evaluation of OOD detection methods.
Our contribution to this field is threefold:
\begin{itemize} 
\item 
We introduce \textbf{ICONIC-444}, a specialized, large-scale industrial image dataset\footnote{The ICONIC-444  dataset is publicly available at \texttt{\href{https://github.com/gkrumpl/iconic-444}{https://github.com/gkrumpl/iconic-444}}.} designed for contamination-free evaluation of OOD detection methods, comprising 444 classes and over 3.1 million images.

\item We define four distinct ID tasks that mirror real-world classification challenges, creating a robust framework for evaluating OOD detection methods across varying complexities.

\item We conduct a comprehensive benchmark of post-hoc OOD detection methods across diverse tasks and network architectures, demonstrating that robust OOD detection remains an unsolved challenge---even for seemingly simple OOD scenarios such as far- and extreme-OOD.
\end{itemize}

\section{Related Work}
\label{sec:related_work}
In the following, we highlight the strengths and limitations of existing OOD detection benchmarks (\cref{sec:rw:existing_ood_datasets}) and summarize OOD detection methods (\cref{sec:rw:ooddmethods}).

\subsection{Existing OOD Detection Datasets}
\label{sec:rw:existing_ood_datasets}
Current OOD detection benchmarks are typically built on well-established image classification datasets~\cite{yang2021_ood_survey, yang2022openood, zhang2023openood}.
Initially, small, low-resolution datasets, such as CIFAR-10/100~\cite{krizhevsky09cifar} and MNIST~\cite{lecun-mnisthandwrittendigit-2010}, were used, with OOD evaluations often conducted on SVHN~\cite{Netzer11_svhn}, LSUN~\cite{yu2016lsun}, Places365~\cite{zhou2017places}, Textures~\cite{cimpoi14dtd}, and TinyImageNet~\cite{Torralba2008_tinyimagenet}.

More recently, larger, higher-resolution datasets have become the standard for OOD detection, with ImageNet-1k~\cite{deng2009_imagenet} emerging as the default ID dataset, \eg as used by~\cite{wang2022_ood_vim, djurisic2023ash, sun2022dice, sun2021_ood_react, yang2022openood}. 
In this context, SSB~\cite{vaze2022openset}, iNaturalist~\cite{Horn_2018_inat}, ImageNet-O~\cite{hendrycks2021_nae}, OpenImage-O~\cite{wang2022_ood_vim}, Places365~\cite{zhou2017places}, and Textures~\cite{cimpoi14dtd} are frequently used as OOD datasets, providing greater class diversity, improved resolution, and a broader range of visual concepts, thereby addressing some limitations of smaller datasets.

To evaluate OOD detection in fine-grained tasks~\cite{ahmed2019semantic, Techapanurak2021, Zhang_2023_WACV, vaze2022openset}, datasets such as CUB-200~\cite{WahCUB_200_2011}, FGVC-Aircraft~\cite{maji13fine-grained}, and Stanford Cars~\cite{Krause2013_stanford_cars} are commonly used, employing a \emph{holdout class} method.
In this setting, some categories are designated as ID and included in the training set, while the remaining categories are held out and treated as OOD during testing.
This approach, often referred to as open-set recognition (OSR)~\cite{yang2021_ood_survey}, assesses the model's ability to detect OOD samples within a visually similar domain. 

Industrial anomaly detection datasets such as MVTec AD~\cite{Bergmann_mvtecad_1, Bergmann_mvtecad_2} and VisA~\cite{yang2022_visa} bring in a real-world context similar to ours, focusing on industrial objects and defect detection. 
However, with their binary classification format (good vs. defective) and limited class diversity, these datasets are less suited for evaluating OOD detection in the wild.

Nonetheless, commonly used datasets face several limitations that can compromise the reliability and generalizability of OOD evaluations. 
For smaller datasets (\eg, CIFAR-10/100, MNIST), issues like limited diversity, low resolution, and contamination undermine their effectiveness as OOD benchmarks~\cite{hendrycks2019_ood_mls, Krumpl2024_ats}.
In larger-scale settings, the ImageNet OOD benchmark contains \emph{categorical contamination}, where OOD classes overlap with ID classes directly or as subsets (\eg, \textit{hayfield} overlapping with \textit{hay}), and \emph{incidental contamination}, where ID objects appear unintentionally within OOD categories (e.g., \textit{airplanes} in \textit{sky} images)~\cite{bitterwolf2023_ninco}. 
This distorts performance metrics, resulting in an inaccurate assessment of model effectiveness and obscuring true model behavior.
To mitigate this, NINCO~\cite{bitterwolf2023_ninco} and ImageNet-OOD~\cite{yang2024imagenetood} have been proposed as OOD test sets for the ImageNet benchmark.
While NINCO provides a cleaner evaluation, it remains relatively small (5879 images, including 64 classes). 
ImageNet-OOD is larger (31807 images from 637 classes) but sourced from ImageNet-21k, which is frequently used to pretrain modern architectures~\cite{dosovitskiy2021vit, Liu_2022_CVPR}.
Consequently, pretrained models may have already seen similar images, artificially inflating OOD detection performance.
Moreover, neither dataset provides explicit control over ID and OOD granularity, which is essential when evaluating fine-grained OOD detection~\cite{Zhang_2023_WACV}.

Similar limitations apply to fine-grained datasets like CUB-200, FGVC-Aircraft, and Stanford Cars. 
Despite offering higher-resolution images, their small sample sizes restrict statistical robustness, and their OOD samples either closely resemble ID holdout classes or are too distant, lacking intermediate granularity.
Furthermore, due to limited training data, these datasets often require ImageNet-pretrained models~\cite{Zhang_2023_WACV, Humblot-Renaux_2024_CVPR, vaze2022openset, Heliang2017_mannfg}, resulting in similar issues of OOD familiarity and performance inflation.

Zhang \etal~\cite{zhang2024openood} show that OOD detection performance is highly task-dependent, with no single method proving universally superior. 
This is compounded by the challenge of ensuring models are evaluated on truly unseen data, as many existing benchmarks suffer from label contamination or unintentional feature exposure during pre-training, which compromises evaluation integrity. 
These issues underscore the need for a diverse suite of methodologically clean benchmarks covering distinct domains. 
ICONIC-444 is introduced as a crucial complement to this suite, providing a large-scale, contamination-free, object-centric benchmark from a specialized industrial domain. 
It's 444 fine-grained classes and structured OOD hierarchy allow the community to reliably evaluate methods on challenges not covered by current resources.

\subsection{OOD Detection Methods}
\label{sec:rw:ooddmethods}
For a comprehensive survey, we refer the interested reader to~\cite{yang2021_ood_survey}.
Here, we focus on OOD detection methods that are \emph{post-hoc} and \emph{sample-free} and, thus, especially valuable in real-world scenarios, as they avoid the need for model retraining and do not require access to OOD samples (which are often unavailable in practice)~\cite{yang2021_ood_survey, zhang2023openood}. 
They can be broadly categorized into \emph{score-based methods} (\eg~\cite{hendrycks2016_ood_msp, hendrycks2019_ood_mls, shiyu17, liu2020_ood_ebo, huang2021_gradnorm, wang2022_ood_vim, Bendale2016_openmax, Ren2021_rmds, zhang2023_she, Gua2017_tempscaling}) and \emph{model enhancement methods} (\eg ~\cite{sun2021_ood_react, djurisic2023ash, sun2022dice, Liu2023_GEN, xu2024scaling}).

\textbf{Score-based methods} derive scalar scores from the model's outputs or intermediate features to distinguish between ID and OOD samples. 
From the perspective of utilized information, the maximum softmax probability (MSP)~\cite{hendrycks2016_ood_msp} and maximum logit score (MLS)~\cite{hendrycks2019_ood_mls} derive scores directly from the model output.
From the feature space, the Mahalanobis distance score (MDS)~\cite{lee2018_mahala} uses the minimum Mahalanobis distance from class centroids to obtain a score for distinguishing between ID and OOD samples.
KNN~\cite{sun2022knnood} adopts a non-parametric approach by calculating the nearest neighbor distances to ID samples in the feature space.
GRAM~\cite{sastry20a_gram} and ATS~\cite{Krumpl2024_ats} use intermediate layer activations for scoring, with GRAM leveraging the Gram matrix and ATS deriving sample-specific temperatures for calibration. 
GradNorm~\cite{huang2021_gradnorm} takes a different approach by using the norm of the gradients to minimize the Kullback-Leibler (KL) divergence between the model predictions and a uniform distribution.

\textbf{Model enhancement methods} modify the inference process to improve OOD detection without altering the underlying model. 
Examples include ReAct~\cite{sun2021_ood_react}, which clips penultimate layer activations during inference, DICE~\cite{sun2022dice}, which sparsifies weights in the final classification layer, and ASH~\cite{djurisic2023ash}, which rectifies and scales the activations in the penultimate layer.
While these methods cannot independently detect OOD samples and require a score function, their promising results on large-scale benchmarks, such as ImageNet, have led to their widespread evaluation~\cite{zhang2023openood}.

\begin{table}
\centering
\footnotesize
\renewcommand{\arraystretch}{0.95}
\setlength{\tabcolsep}{4pt}
\resizebox{\linewidth}{!}{
\begin{tabular}{lccccc}
\toprule
\multirow{2}{*}{\textbf{Dataset}} &
\multirow{2}{*}{\textbf{Classes}} &
\multirow{2}{*}{\textbf{Samples}} &
\multicolumn{3}{c}{\textbf{Designed for}} \\
\cmidrule(lr){4-6}
& & & IC & AD & OODD \\
\midrule
CIFAR-10~\cite{krizhevsky09cifar}                       & \phantom{00}10                                       & \phantom{00}60000                                 & \cmark & \xmark & \xmark    \\
CIFAR-100~\cite{krizhevsky09cifar}                      & \phantom{0}100                                       & \phantom{00}60000                                 & \cmark & \xmark & \xmark    \\
ImageNet-1k~\cite{deng2009_imagenet}                    & 1000                                                 & 1431167                               & \cmark & \xmark & \xmark    \\
FGVC-Aircraft~\cite{maji13fine-grained}                 & \phantom{0}102                                       & \phantom{00}10200                                 & \cmark & \xmark & \xmark    \\
Stanford Cars~\cite{Krause2013_stanford_cars}           & \phantom{0}196                                       & \phantom{00}16185                                 & \cmark & \xmark & \xmark    \\
CUB-200~\cite{WahCUB_200_2011}                          & \phantom{0}200                                       & \phantom{00}11788                                 & \cmark & \xmark & \xmark    \\
North American Birds~\cite{horn2015_northamerikanbirds} & \phantom{0}555                                       & \phantom{00}48562                                 & \cmark & \xmark & \xmark    \\
MVTec AD~\cite{Bergmann_mvtecad_1, Bergmann_mvtecad_2} & \phantom{00}15                                        & \phantom{000}5354                                  & \xmark & \cmark & \xmark    \\
VisA~\cite{yang2022_visa}                              & \phantom{00}12                                        & \phantom{00}10821                                 & \xmark & \cmark & \xmark    \\
\midrule
\rowcolor{gray!10}\textbf{ICONIC-444 (Ours)}            & \phantom{0}444                                       & 3132406                               & \cmark & \xmark & \cmark\\
\bottomrule

\end{tabular}
}
\caption{
Commonly used ID datasets for OOD detection evaluation, along with related anomaly detection datasets.
The "Designed for" column indicates the initially intended application of each dataset: image classification (IC), anomaly detection (AD), or out-of-distribution detection (OODD). 
For exclusive AD datasets, defective samples are not counted as separate classes.
}
\label{tab:dataset_sizes}
\end{table}

\section{Dataset}

Our ICONIC-444 (\textbf{I}mage \textbf{C}lassification and \textbf{O}OD Detection with \textbf{N}umerous \textbf{I}ntricate \textbf{C}omplexities) dataset is designed to complement existing benchmarks by addressing common limitations, namely, limited class diversity, lack of controllable complexity for both ID and OOD tasks, absence of smooth transitions between near- and far-OOD samples, ID-contaminated OOD test sets, small sample sizes, and low resolution.
As detailed in \cref{sec:ds:descp}, ICONIC-444 provides over $3.1$ million high-quality, high-resolution images across 444 classes targeted for OOD detection.
This dataset supports both coarse- and fine-grained ID scenarios of various difficulties (depending on the chosen ID classes) and
a well-defined, contamination-free OOD hierarchy (\mbox{near-,} far-, extreme-, and synthetic). 

To facilitate standardized evaluations, we define four benchmark tasks (\cref{sec:dataset_description:bmtasks}). 
ICONIC-444 complements existing datasets with its focus on controlled, industrial application scenarios while superseding most existing OOD detection datasets in either size (amount of images available), resolution, cleanliness (individual classes being clearly defined and separated, data curated to a high-quality standard), or breadth of scenarios which can be represented, as shown in~\cref{tab:dataset_sizes}.
This makes it an ideal resource to support future research and conduct robust statistical evaluations.

\subsection{Description}
\label{sec:ds:descp}
Industrial object sorting serves as an ideal testbed, as it presents unique challenges in balancing the reliable rejection of true OOD samples with the retention of valuable ID items. 
This inspired us to leverage this setting to collect our ICONIC-444, organized into two categories, 19 groups, and 444 classes, as illustrated in \cref{fig:teaser} (example images for each class are included in the supplemental material).
The \textit{Food} category comprises 387 classes divided into 12 groups, while the \textit{Non-Food} category includes 57 classes across seven groups.
Individual classes contain between 1303 and 51283 images in the food category and between 659 and 19175 images in the non-food category.
This long-tailed distribution is intentional and reflects real-world environments.
Classes were selected based on the following three criteria:

\emph{Size suitability:} The objects within each class fall within a specified size range (a few millimeters to centimeters) to ensure compatibility with our acquisition setup.

\emph{Distinctness:} Classes are clearly defined and distinct, meaning each item belongs exclusively and unambiguously to a single class.

\emph{Available in bulk:} Classes are selected based on their feasibility for bulk acquisition to ensure we can capture the intra-class variance and enable large-scale data collection.

ICONIC-444 encompasses fine- to coarse-grained complexity levels, developed with input from domain experts in food sorting, agriculture, and waste recycling. 
While all classes are mutually exclusive, distinctions are often fine-grained or subtle; for example, different breeds of wheat, although distinct, may appear visually similar to non-experts.
This applies, for example, to 12 types of wheat and multiple breeds of rye, barley, and triticale, along with other visually similar classes within the rice, beans, and coffee groups. 
In contrast, coarser distinctions exist in groups like \textit{seeds and kernels}, while cross-group classification across broader categories, such as \textit{spices}, \textit{sweets}, and \textit{snacks}, enables even more coarse-grained tasks suitable for varying levels of complexity.

To further increase complexity and add additional fine-grained classes, manual defects were introduced by damaging, painting, or burning items from certain classes.
For instance, pasta samples were altered with drilled holes, breakages, and small colored dots to simulate damage and contamination. 
A total of 53 classes exist with such manually induced defects. 
These items add to not only the dataset’s applicability for OOD detection but also open opportunities for anomaly detection tasks in the future, further enhancing the usefulness of ICONIC-444.

\paragraph{Data acquisition.}
The data was acquired using an industrial free-fall sorting machine prototype, equipped with a 4K RGB line-scan camera.
This setup captures images at a resolution of $56\,\mu\text{m/pixel}$, providing highly detailed images surpassing the resolution of typical human vision when viewed even from close up. 
The high resolution enables models to capture the fine details necessary for class differentiation, particularly in fine-grained classification.

For bulk data acquisition, objects are placed on a vibrating conveyor that separates and feeds them onto a chute, which stabilizes and aligns the objects before they enter free fall. 
The line-scan camera captures the objects mid-air as a continuous stream.
A blue LED backlight provides a consistent blue background, while two white LED lights illuminate the objects from the front. 
The constant blue background facilitates automatic detection and cropping, resulting in images with resolutions ranging from $256 \times 256$ to $1024 \times 1024$ pixels, depending on the maximum expected object size for each class.
A video showcasing our acquisition setup is available in the supplementary material.

Variability within the individual classes comes naturally:
Due to the vast majority of our data being natural products or processed natural products, they come with inherent variability in their appearance. 
Even man-made materials often exhibit high variability, such as the broken edges in shredded glass.
Only very few classes (such as coins, plastic building blocks, or electronic components) present a high degree of standardization, though they often vary by type (5-, 10-, 20-cent-coins, \etc) or show age-induced wear and tear. 
Additional variability is introduced by the different ways in which items fall through the machine, such as rotation and slight differences in falling speed, enhancing intra-class diversity.
These variabilities closely resemble those found in real-world sorting due to our acquisition setup being almost identical to operational sorting machines.

\begin{figure*}[t]
\centering
\includegraphics[width=\linewidth]{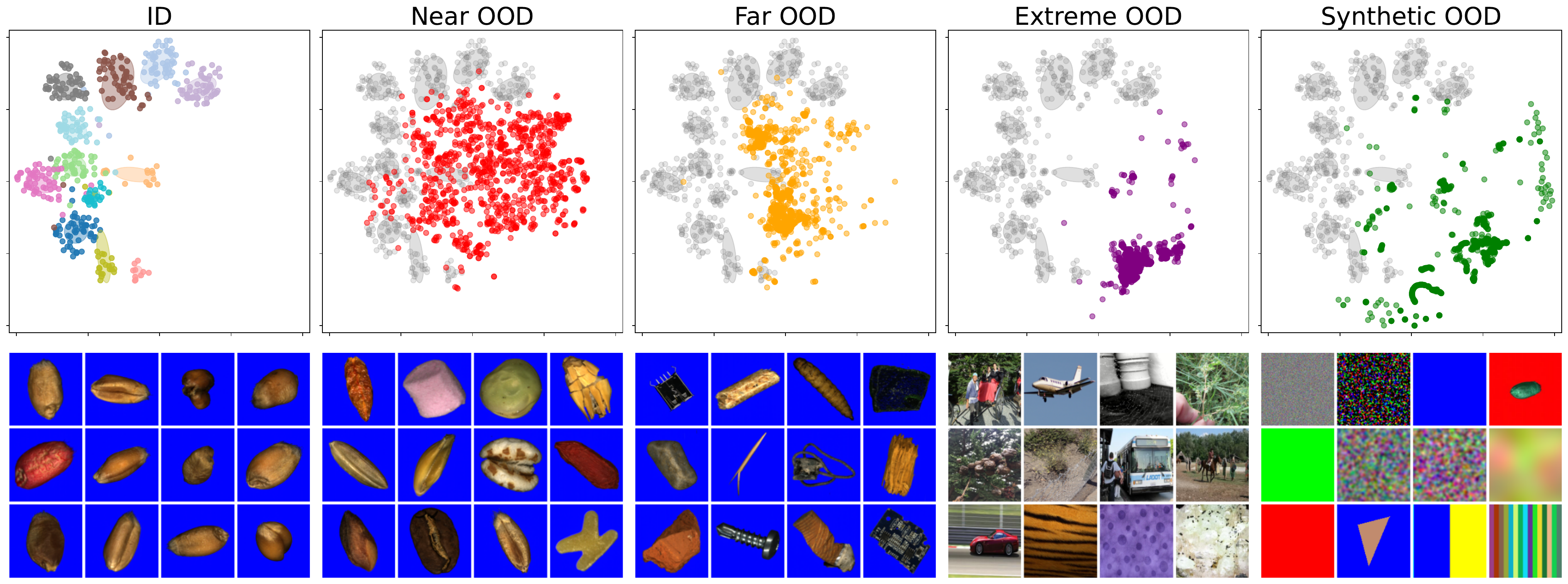}
\caption{
Visualization of data samples (bottom row) from the Wheat task, showing the 12 ID classes and all OOD categories (near, far, extreme, and synthetic) and their embedded representations (top) extracted from the ResNet18 penultimate layer. 
The t-SNE visualizations (top) include only a limited subset of samples to enhance readability. 
The bottom row shows representative images from each category, highlighting the sample diversity. 
For ID data, one sample per class is shown; for OOD data, only a small subset of classes is depicted.
}
\label{fig:tsne}
\end{figure*}

\paragraph{Data quality.}
To ensure the highest possible quality of the data, we focused on the following three key areas: 

\emph{Sourcing of materials}: When sourcing, we ensured that pure, uncontaminated materials were acquired. 
Most food classes were store-bought and food-grade, which, given food safety regulations, already guarantees a high degree of purity.
Most grains and seeds were sourced directly from suppliers or plant breeders, who themselves have quality protocols to keep their breeds clean from contamination.

\emph{Data acquisition}: We adhered to strict protocols when taking data: Items from only one class were handled at a time, with the machine being inspected and cleaned after every acquisition run. 
Each image in ICONIC-444 represents a distinct physical object captured only once.

\emph{Data cleaning}: 
Post-acquisition, all samples were semi-automatically checked to filter out objects that were too small (potential dust and debris), too dark (outside well-illuminated regions), or were touching other objects.
Additionally, we leveraged image classification models in a cross-validation setup (training and evaluating on separate subsets) to flag suspicious samples for manual inspection. 
Further details about the acquisition protocol and the cleaning and quality control procedures are provided in the supplementary material (Section A).

We define the benchmark tests using ID classes from the food category due to their real-world relevance in safety- and sustainability-critical applications. 
Their high quality as food-grade items and high intra-class but low inter-class variance (depending on the ID task) make them well-suited for fine-grained classification. 
In contrast, non-food classes such as glass, stones, \etc serve as OOD samples, mirroring real-world conditions. 
Quality control follows the same rigorous standards across all classes to ensure dataset integrity.

The breadth and volume of data acquired, combined with the intense quality control and well-defined class and group structure, make the ICONIC-444 dataset ideal for creating a variety of tasks (see \cref{sec:dataset_description:bmtasks}), serving as a comprehensive framework for evaluating both OOD detection methods and image classification. 
It offers a wide range of complexities and, thanks to the sheer amount of data available, allows for robust statistical evaluations.

\begin{table}
\centering
\footnotesize
\resizebox{\linewidth}{!}{
\begin{tabular}{lllcccc}
\toprule
\multirow{2}{*}{\textbf{Task}} & \multicolumn{1}{c}{\multirow{2}{*}{\textbf{ID}}} & \multicolumn{5}{c}{\textbf{OOD}} \\
\cmidrule(lr){3-7}
& &  \multicolumn{1}{c}{Near} & Far & Extreme & Synthetic & Validation \\
\midrule
Almond   & \phantom0\phantom07 (\phantom0\phantom09.4k/\phantom0\phantom01.3k/\phantom0\phantom02.7k) & 380 (190.0k)        &        &          &         &         \\
Wheat    & \phantom012 (\phantom083.2k/\phantom011.8k/\phantom023.7k)      & 375 (187.5k)        & 51     & 1522     & 25      & 6       \\
Kernels  & \phantom029 (365.3k/\phantom052.1k/104.4k)    & 358 (179.0k)        &(25.5k) & (102.1k) & (12.5k) & (10.0k) \\
Food-grade   & 324 (291.6k/\phantom032.4k/\phantom097.2k)             & \phantom063 (\phantom031.5k) &        &          &         &         \\
\bottomrule

\end{tabular}
}
\caption{
Summary of the four benchmark tasks. 
For the ID tasks, we report the number of classes along with the number of train, validation, and test images (\#Classes [Train/Val/Test]).
For the OOD sets, we report the number of classes and total images.
}
\label{tab:dataset_splits}
\end{table}

\subsection{Benchmark Tasks}
\label{sec:dataset_description:bmtasks}
To demonstrate the utility of ICONIC-444 in evaluating OOD detection methods, we define four benchmark tasks that encompass a wide range of complexities, as summarized in \cref{tab:dataset_splits}.
For each ID task, we further define corresponding OOD test sets, which consist of subsets of the ICONIC-444 dataset, public datasets, and synthetically generated samples. 
Each task differs in ID class count and dataset size and provides different levels of complexity and granularity for classification models and OOD detection.

The first ID task, \textbf{Almond}, comprises seven almond-related classes within the \textit{nuts} group from the \textit{Food} category, driven by a real-world food-sorting application. 
This task offers a moderate classification challenge with classes that are visually similar yet distinct. 
The second task, \textbf{Wheat}, focuses on the classification of 12 different wheat varieties, representing a fine-grained classification problem due to the subtle visual differences among wheat types. 
The third task, \textbf{Kernels}, encompasses all 29 classes from the \textit{seeds and kernels} group, increasing the number of classes and covering medium-grained distinctions.
The final task, \textbf{Food-grade}, includes all 324 non-damaged food classes---manually introduced defects, nut shells, and substandard products (such as rancid or damaged items) are excluded.

In all tasks, OOD samples are defined as classes not included in the ID task. 
Recognizing that OOD data come in different types---varying in complexity and severity---and that requirements for detecting different types of OOD may vary between use cases, we group our OOD samples into four categories: near-OOD, far-OOD, extreme-OOD, and synthetic-OOD.
\textbf{Near-OOD} samples are drawn from the \textit{Food} category but are not part of the ID classes for the specific task, varying accordingly for each task. 
\textbf{Far-OOD} samples are consistently sourced from the \textit{Non-Food} category across all tasks. 
\textbf{Extreme-OOD} samples are included from four public datasets: ImageNet~\cite{deng2009_imagenet}, iNaturalist~\cite{Horn_2018_inat} (curated version from \cite{huang2021mos}), Places365~\cite{zhou2017places}, and Textures~\cite{cimpoi14dtd}, which are semantically unrelated to ICONIC-444, and provide a broader range for OOD evaluation. 

Following prior works~\cite{hendrycks2016_ood_msp, hendrycks2019oe, hein2019relu, bitterwolf2023_ninco}, we also add \textbf{synthetic-OOD} samples to test specific weaknesses in OOD detection, including camera failures, data corruption, and transmission errors.
We included 25 synthetic-OOD types comprising both pre-existing and newly proposed variations.
Details and example images of these synthetic OOD samples are provided in the supplementary material.
 
For the near- and far-OOD test sets, 500 samples are randomly selected per class for a balanced representation.
Additionally, six classes are removed from the far-OOD set to form an OOD validation set, with 10k samples randomly drawn from these classes. 
This setup enables hyperparameter tuning for OOD detection methods that require it to ensure optimal performance.
The synthetic-OOD set includes 500 generated samples per type, creating a comprehensive OOD test set covering a wide range of complexities.

For a visual impression of the imposed complexities, \cref{fig:tsne} shows examples of the ID and OOD classes of the Wheat task.
The corresponding t-SNE plot shows that near-OOD samples, which have subtle distinctions from ID classes, span the largest area and overlap significantly with several ID clusters, indicating their high semantic similarity (an observation consistent with findings in~\cite{Zhang_2023_WACV}). 
As the semantic similarity of OOD samples decreases (\ie, far- and extreme-OOD), the spanned area contracts, and samples cluster more densely, with a noticeable increase in distance from ID clusters.
Synthetic-OOD data, typically appearing further from ID clusters, aligns with this trend, though some classes, such as color channel permuted images, show high semantic similarity and are positioned closer to ID clusters.

\section{Experiments}
\label{sec:experiments}

We conduct extensive experiments on ICONIC-444 to provide baseline results and insights for state-of-the-art post-hoc OOD detection methods.
\cref{sec:exp:setup} describes the experimental setup, while \cref{sec:exp:analysis} presents the detailed analysis.

\subsection{Experimental Setup}
\label{sec:exp:setup}
\paragraph{Setup.} 
We evaluate the four ID tasks---Almond, Wheat, Kernels, and Food-grade---as detailed in \cref{sec:dataset_description:bmtasks}, utilizing separate training, validation, and test splits derived from the ICONIC-444 dataset. 
We provide baseline results for two architectures: Traditional Convolutional Neural Networks, represented by ResNet18~\cite{He2015_resnet}, and transformer architectures, represented by the Compact Transformer (CCT) model~\cite{hassani202_cct}.
To ensure robustness and account for variability in training, each ID task is trained with three different random seeds. 
For OOD detection methods that require ID data to derive statistics or parameters, we utilize the ID training split. 
Additionally, for methods needing hyperparameter tuning, we leverage both the ID and OOD validation sets.
Details on more complex architectures, including ConvNeXt~\cite{Liu_2022_CVPR}, Vision Transformer~\cite{dosovitskiy2021vit} (ViT), and CLIP~\cite{radford2021_clip}, are in the supplementary material, along with further details on training and hyperparameter search.

\paragraph{Evaluation metric.}
Following established evaluation protocols for OOD detection~\cite{hendrycks2016_ood_msp, hendrycks2019_ood_mls, sun2021_ood_react, sun2022dice, djurisic2023ash}, we assess the performance using the False Positive Rate at 95$\%$ True Positive Rate (FPR95) and the Area Under the Receiver Operating Characteristic Curve (AUROC) as evaluation metrics. 
We furthermore propose the False Positive Rate at 99$\%$ True Positive Rate (FPR99) as an additional metric: 
Rejecting 5$\%$ of good data is typically unacceptable in real-world applications, and keeping the goal of making OOD detection viable for such applications in mind, we need stricter performance criteria. 
Note that FPR99 requires a large evaluation dataset to yield statistically meaningful results.
ICONIC-444 is the first dataset to provide a sufficient number of images for such statistically relevant analyses.
We strongly encourage the additional use of FPR99 when evaluating future advances in OOD detection methods.

\paragraph{Test time and evaluation.}
For all tasks, we preprocess the images by resizing them to $256 \times 256$ pixels and then center-cropping them to $224 \times 224$ pixels to ensure consistency across inputs. 
To evaluate OOD detection performance on our dataset, we employ 22 post-hoc methods from the OpenOOD benchmark~\cite{yang2022openood, zhang2023openood}, a comprehensive open-source OOD detection benchmark library.

\begin{table}
\centering
\footnotesize
\resizebox{\linewidth}{!}{
\begin{tabular}{lcccc}
\toprule
\multirow{2}{*}{\textbf{Task}} & \multicolumn{2}{c}{\textbf{ResNet}} & \multicolumn{2}{c}{\textbf{CCT}} \\
                                 & Accuracy & ECE & Accuracy & ECE \\
\midrule
Almond & 95.53$^{\scriptsize\pm \phantom{0}0.11}$ & \phantom{0}1.33$^{\scriptsize\pm \phantom{0}0.18}$ & 93.63$^{\scriptsize\pm \phantom{0}0.50}$ & \phantom{0}4.75$^{\scriptsize\pm \phantom{0}0.37}$ \\
Wheat & 94.72$^{\scriptsize\pm \phantom{0}0.06}$ & \phantom{0}1.51$^{\scriptsize\pm \phantom{0}0.11}$ & 93.77$^{\scriptsize\pm \phantom{0}0.45}$ & \phantom{0}4.30$^{\scriptsize\pm \phantom{0}0.36}$ \\
Kernels & 96.38$^{\scriptsize\pm \phantom{0}0.14}$ & \phantom{0}0.78$^{\scriptsize\pm \phantom{0}0.07}$ & 95.75$^{\scriptsize\pm \phantom{0}0.20}$ & \phantom{0}3.10$^{\scriptsize\pm \phantom{0}0.16}$ \\
Food-grade & 94.31$^{\scriptsize\pm \phantom{0}0.31}$ & \phantom{0}0.44$^{\scriptsize\pm \phantom{0}0.06}$ & 93.18$^{\scriptsize\pm \phantom{0}0.04}$ & \phantom{0}3.62$^{\scriptsize\pm \phantom{0}0.04}$ \\
\bottomrule

\end{tabular}
}
\caption{
Performance across four ID tasks on the test set.
For each architecture and task, we report the mean Accuracy and Expected Calibration Error (ECE), averaged over three random seeds and expressed as percentages.
}
\label{tab:acc_ece}
\end{table}

\begin{table*}
\centering\footnotesize
\setlength{\extrarowheight}{1.5pt}
\resizebox{\textwidth}{!}{
\begin{tabular}{lccccccccccccccc}
\toprule
\multirow{3}{*}{\textbf{Method}} & \multicolumn{12}{c}{\textbf{OOD-Dataset}} \\
\cmidrule(lr){2-13}
 & \multicolumn{3}{c}{\textbf{Near}} & \multicolumn{3}{c}{\textbf{Far}} & \multicolumn{3}{c}{\textbf{Extreme}} & \multicolumn{3}{c}{\textbf{Synthetic}} & \multicolumn{3}{c}{\textbf{Average}} \\
\cmidrule(lr){2-4} \cmidrule(lr){5-7} \cmidrule(lr){8-10} \cmidrule(lr){11-13} \cmidrule(lr){14-16}
 & FPR95 $\downarrow$ & FPR99 $\downarrow$ & AUROC $\uparrow$
 & FPR95 $\downarrow$ & FPR99 $\downarrow$ & AUROC $\uparrow$
 & FPR95 $\downarrow$ & FPR99 $\downarrow$ & AUROC $\uparrow$
 & FPR95 $\downarrow$ & FPR99 $\downarrow$ & AUROC $\uparrow$
 & FPR95 $\downarrow$ & FPR99 $\downarrow$ & AUROC $\uparrow$ \\
\midrule
GRAM~\cite{sastry20a_gram} & $\cellcolor[HTML]{CCDCED} \textbf{35.74}$ & $\cellcolor[HTML]{EAF0F8} \textbf{54.59}$ & $\cellcolor[HTML]{C5D7EB} \textbf{88.13}$ & $\cellcolor[HTML]{ABC4E1} \textbf{19.75}$ & $\cellcolor[HTML]{CEDDEE} \textbf{36.89}$ & $\cellcolor[HTML]{A2BEDE} \textbf{94.03}$ & $\cellcolor[HTML]{749ECE} \phantom{0}\textbf{0.00}$ & $\cellcolor[HTML]{749ECE} \phantom{0}\textbf{0.00}$ & $\cellcolor[HTML]{749ECE} \textbf{99.99}$ & $\cellcolor[HTML]{759FCE} \phantom{0}\textbf{0.44}$ & $\cellcolor[HTML]{7BA3D0} \phantom{0}\textbf{2.40}$ & $\cellcolor[HTML]{759ECE} \textbf{99.88}$ & $\cellcolor[HTML]{9CBADC} \textbf{13.98}$ & $\cellcolor[HTML]{B3CAE4} \textbf{23.47}$ & $\cellcolor[HTML]{97B7DA} \textbf{95.51}$ \\
ATS~\cite{Krumpl2024_ats} & $\cellcolor[HTML]{F6F9FC} 66.27$ & $\cellcolor[HTML]{FEFEFF} 80.76$ & $\cellcolor[HTML]{F0F5FA} 76.35$ & $\cellcolor[HTML]{E2EBF5} 48.90$ & $\cellcolor[HTML]{F7F9FC} 67.99$ & $\cellcolor[HTML]{D0DEEE} 86.04$ & $\cellcolor[HTML]{76A0CF} \phantom{0}\underline{0.81}$ & $\cellcolor[HTML]{7BA3D0} \phantom{0}\underline{2.25}$ & $\cellcolor[HTML]{759FCE} \underline{99.84}$ & $\cellcolor[HTML]{99B8DB} \underline{13.07}$ & $\cellcolor[HTML]{ABC4E1} \underline{20.02}$ & $\cellcolor[HTML]{97B6DA} 95.59$ & $\cellcolor[HTML]{C6D7EB} 32.26$ & $\cellcolor[HTML]{D9E4F2} \underline{42.76}$ & $\cellcolor[HTML]{BED2E8} 89.45$ \\
VIM~\cite{wang2022_ood_vim} & $\cellcolor[HTML]{DDE7F3} 45.37$ & $\cellcolor[HTML]{F6F9FC} 66.79$ & $\cellcolor[HTML]{D2DFEF} 85.53$ & $\cellcolor[HTML]{C6D7EB} 32.42$ & $\cellcolor[HTML]{ECF2F8} 56.64$ & $\cellcolor[HTML]{A5C0DF} \underline{93.57}$ & $\cellcolor[HTML]{A1BDDE} 15.92$ & $\cellcolor[HTML]{C4D6EA} 31.44$ & $\cellcolor[HTML]{8BAED6} 97.12$ & $\cellcolor[HTML]{9CBADC} 14.07$ & $\cellcolor[HTML]{B4CBE4} 23.87$ & $\cellcolor[HTML]{91B3D8} \underline{96.35}$ & $\cellcolor[HTML]{BBCFE7} \underline{26.94}$ & $\cellcolor[HTML]{DCE7F3} 44.69$ & $\cellcolor[HTML]{A8C2E0} \underline{93.14}$ \\
RMDS~\cite{Ren2021_rmds} & $\cellcolor[HTML]{DDE7F3} 45.27$ & $\cellcolor[HTML]{F4F7FB} \underline{64.29}$ & $\cellcolor[HTML]{DDE7F3} 82.96$ & $\cellcolor[HTML]{D8E4F1} 42.53$ & $\cellcolor[HTML]{F0F5FA} 60.64$ & $\cellcolor[HTML]{D6E2F0} 84.61$ & $\cellcolor[HTML]{B6CCE5} 24.66$ & $\cellcolor[HTML]{C7D8EB} 33.01$ & $\cellcolor[HTML]{BED1E8} 89.60$ & $\cellcolor[HTML]{BBCFE7} 27.07$ & $\cellcolor[HTML]{CEDDEE} 36.65$ & $\cellcolor[HTML]{C2D5EA} 88.71$ & $\cellcolor[HTML]{CBDBED} 34.88$ & $\cellcolor[HTML]{E2EAF5} 48.65$ & $\cellcolor[HTML]{CDDCEE} 86.47$ \\
MDS~\cite{lee2018_mahala} & $\cellcolor[HTML]{E1EAF4} 48.23$ & $\cellcolor[HTML]{F7FAFC} 68.30$ & $\cellcolor[HTML]{DFE9F4} 82.33$ & $\cellcolor[HTML]{CEDDEE} 36.92$ & $\cellcolor[HTML]{EEF3F9} 58.31$ & $\cellcolor[HTML]{C7D8EB} 87.75$ & $\cellcolor[HTML]{ACC5E2} 20.35$ & $\cellcolor[HTML]{CCDCED} 35.52$ & $\cellcolor[HTML]{AEC7E3} 92.19$ & $\cellcolor[HTML]{B4CBE4} 23.74$ & $\cellcolor[HTML]{C6D7EB} 32.54$ & $\cellcolor[HTML]{BDD1E8} 89.65$ & $\cellcolor[HTML]{C6D7EB} 32.31$ & $\cellcolor[HTML]{E2EBF5} 48.67$ & $\cellcolor[HTML]{C6D7EB} 87.98$ \\
KNN~\cite{sun2022knnood} & $\cellcolor[HTML]{DAE6F2} \underline{43.75}$ & $\cellcolor[HTML]{F5F8FC} 65.54$ & $\cellcolor[HTML]{D7E3F1} 84.32$ & $\cellcolor[HTML]{C8D9EC} 33.64$ & $\cellcolor[HTML]{EEF3F9} 57.85$ & $\cellcolor[HTML]{ACC5E2} 92.47$ & $\cellcolor[HTML]{A8C3E0} 18.93$ & $\cellcolor[HTML]{D2E0EF} 38.97$ & $\cellcolor[HTML]{91B3D8} 96.32$ & $\cellcolor[HTML]{ACC5E2} 20.26$ & $\cellcolor[HTML]{CBDBED} 35.14$ & $\cellcolor[HTML]{A2BEDE} 93.96$ & $\cellcolor[HTML]{BFD2E9} 29.14$ & $\cellcolor[HTML]{E3EBF5} 49.37$ & $\cellcolor[HTML]{B1C9E4} 91.77$ \\
ODIN~\cite{shiyu17} & $\cellcolor[HTML]{E0EAF4} 47.70$ & $\cellcolor[HTML]{F6F9FC} 66.19$ & $\cellcolor[HTML]{D9E4F2} 83.84$ & $\cellcolor[HTML]{C4D6EA} \underline{31.44}$ & $\cellcolor[HTML]{E6EEF6} \underline{51.61}$ & $\cellcolor[HTML]{B5CBE5} 91.14$ & $\cellcolor[HTML]{BBD0E7} 27.29$ & $\cellcolor[HTML]{D5E2F0} 40.63$ & $\cellcolor[HTML]{CCDCED} 86.73$ & $\cellcolor[HTML]{BFD2E9} 29.01$ & $\cellcolor[HTML]{D3E1F0} 39.51$ & $\cellcolor[HTML]{D7E3F1} 84.45$ & $\cellcolor[HTML]{C9D9EC} 33.86$ & $\cellcolor[HTML]{E3ECF5} 49.49$ & $\cellcolor[HTML]{CDDCEE} 86.54$ \\
GEN~\cite{Liu2023_GEN} & $\cellcolor[HTML]{EDF2F9} 57.27$ & $\cellcolor[HTML]{FDFDFE} 77.53$ & $\cellcolor[HTML]{E1EAF4} 81.66$ & $\cellcolor[HTML]{DEE8F3} 46.29$ & $\cellcolor[HTML]{F7F9FC} 68.15$ & $\cellcolor[HTML]{C9D9EC} 87.36$ & $\cellcolor[HTML]{E1EAF4} 48.21$ & $\cellcolor[HTML]{F3F7FB} 63.07$ & $\cellcolor[HTML]{E2EAF5} 81.57$ & $\cellcolor[HTML]{E2EBF5} 48.92$ & $\cellcolor[HTML]{F4F7FB} 64.82$ & $\cellcolor[HTML]{EAF0F8} 78.94$ & $\cellcolor[HTML]{E4ECF5} 50.17$ & $\cellcolor[HTML]{F7FAFC} 68.39$ & $\cellcolor[HTML]{DFE9F4} 82.38$ \\
DICE~\cite{sun2022dice} & $\cellcolor[HTML]{F0F4FA} 60.00$ & $\cellcolor[HTML]{FDFDFE} 77.82$ & $\cellcolor[HTML]{ECF2F8} 77.87$ & $\cellcolor[HTML]{E0EAF4} 47.64$ & $\cellcolor[HTML]{F8FAFD} 69.66$ & $\cellcolor[HTML]{D3E1F0} 85.13$ & $\cellcolor[HTML]{E6EEF6} 51.73$ & $\cellcolor[HTML]{F5F8FB} 65.41$ & $\cellcolor[HTML]{FBFCFD} 70.42$ & $\cellcolor[HTML]{E2EBF5} 48.95$ & $\cellcolor[HTML]{F2F6FA} 62.30$ & $\cellcolor[HTML]{FBFCFD} 70.41$ & $\cellcolor[HTML]{E6EEF6} 52.08$ & $\cellcolor[HTML]{F7FAFC} 68.80$ & $\cellcolor[HTML]{F1F5FA} 75.96$ \\
ReAct~\cite{sun2021_ood_react} & $\cellcolor[HTML]{F0F5FA} 60.46$ & $\cellcolor[HTML]{FDFEFE} 80.07$ & $\cellcolor[HTML]{E3ECF5} 81.04$ & $\cellcolor[HTML]{E0EAF4} 47.69$ & $\cellcolor[HTML]{FAFBFD} 71.59$ & $\cellcolor[HTML]{C9D9EC} 87.41$ & $\cellcolor[HTML]{E2EBF5} 48.98$ & $\cellcolor[HTML]{F4F7FB} 64.46$ & $\cellcolor[HTML]{E0E9F4} 82.04$ & $\cellcolor[HTML]{E3ECF5} 49.52$ & $\cellcolor[HTML]{F5F8FC} 66.02$ & $\cellcolor[HTML]{E3EBF5} 81.27$ & $\cellcolor[HTML]{E6EEF6} 51.66$ & $\cellcolor[HTML]{F9FBFD} 70.54$ & $\cellcolor[HTML]{DDE7F3} 82.94$ \\
EBO~\cite{liu2020_ood_ebo} & $\cellcolor[HTML]{F0F4FA} 59.77$ & $\cellcolor[HTML]{FDFDFE} 78.72$ & $\cellcolor[HTML]{EEF3F9} 77.59$ & $\cellcolor[HTML]{E2EAF5} 48.57$ & $\cellcolor[HTML]{F8FAFD} 69.76$ & $\cellcolor[HTML]{D6E2F0} 84.65$ & $\cellcolor[HTML]{EBF1F8} 55.54$ & $\cellcolor[HTML]{F6F9FC} 66.70$ & $\cellcolor[HTML]{FCFDFE} 69.26$ & $\cellcolor[HTML]{E9EFF7} 53.80$ & $\cellcolor[HTML]{F7FAFC} 68.90$ & $\cellcolor[HTML]{FCFDFE} 69.28$ & $\cellcolor[HTML]{EAF0F8} 54.42$ & $\cellcolor[HTML]{F9FBFD} 71.02$ & $\cellcolor[HTML]{F3F7FB} 75.19$ \\
SHE~\cite{zhang2023_she} & $\cellcolor[HTML]{F0F5FA} 60.57$ & $\cellcolor[HTML]{FDFEFE} 80.02$ & $\cellcolor[HTML]{F6F9FC} 73.77$ & $\cellcolor[HTML]{E6EEF6} 51.46$ & $\cellcolor[HTML]{FAFCFD} 72.99$ & $\cellcolor[HTML]{E3EBF5} 81.28$ & $\cellcolor[HTML]{E4ECF5} 49.99$ & $\cellcolor[HTML]{F7F9FC} 67.52$ & $\cellcolor[HTML]{FCFDFE} 69.09$ & $\cellcolor[HTML]{E3ECF5} 49.78$ & $\cellcolor[HTML]{F4F7FB} 63.97$ & $\cellcolor[HTML]{FDFDFE} 67.88$ & $\cellcolor[HTML]{E8EFF7} 52.95$ & $\cellcolor[HTML]{F9FBFD} 71.13$ & $\cellcolor[HTML]{F7F9FC} 73.00$ \\
OpenMax~\cite{Bendale2016_openmax} & $\cellcolor[HTML]{DBE6F2} 44.06$ & $\cellcolor[HTML]{FDFEFE} 80.23$ & $\cellcolor[HTML]{CDDCEE} \underline{86.50}$ & $\cellcolor[HTML]{CEDDEE} 36.38$ & $\cellcolor[HTML]{FEFEFF} 80.28$ & $\cellcolor[HTML]{B9CEE6} 90.45$ & $\cellcolor[HTML]{ABC4E1} 20.10$ & $\cellcolor[HTML]{F0F5FA} 60.63$ & $\cellcolor[HTML]{9AB9DB} 95.18$ & $\cellcolor[HTML]{BBD0E7} 27.29$ & $\cellcolor[HTML]{F4F7FB} 64.33$ & $\cellcolor[HTML]{A8C2E0} 93.17$ & $\cellcolor[HTML]{C5D7EB} 31.96$ & $\cellcolor[HTML]{FAFBFD} 71.37$ & $\cellcolor[HTML]{B4CBE4} 91.33$ \\
SCALE~\cite{xu2024scaling} & $\cellcolor[HTML]{F1F5FA} 60.90$ & $\cellcolor[HTML]{FEFEFF} 80.74$ & $\cellcolor[HTML]{E8EFF7} 79.46$ & $\cellcolor[HTML]{DFE9F4} 46.63$ & $\cellcolor[HTML]{F8FAFD} 70.20$ & $\cellcolor[HTML]{CCDCED} 86.74$ & $\cellcolor[HTML]{E7EEF7} 52.66$ & $\cellcolor[HTML]{F5F8FB} 65.37$ & $\cellcolor[HTML]{EFF4F9} 76.83$ & $\cellcolor[HTML]{E8EFF7} 52.99$ & $\cellcolor[HTML]{F8FAFD} 69.43$ & $\cellcolor[HTML]{F4F7FB} 74.47$ & $\cellcolor[HTML]{E8EFF7} 53.29$ & $\cellcolor[HTML]{FAFBFD} 71.44$ & $\cellcolor[HTML]{E9EFF7} 79.38$ \\
MLS~\cite{hendrycks2019_ood_mls} & $\cellcolor[HTML]{EFF4F9} 59.07$ & $\cellcolor[HTML]{FDFEFE} 79.12$ & $\cellcolor[HTML]{EBF1F8} 78.32$ & $\cellcolor[HTML]{E2EAF5} 48.56$ & $\cellcolor[HTML]{F9FBFD} 71.14$ & $\cellcolor[HTML]{D3E1F0} 85.14$ & $\cellcolor[HTML]{EAF1F8} 55.00$ & $\cellcolor[HTML]{F6F9FC} 66.89$ & $\cellcolor[HTML]{FCFDFE} 69.44$ & $\cellcolor[HTML]{E9F0F7} 54.18$ & $\cellcolor[HTML]{F8FAFD} 69.32$ & $\cellcolor[HTML]{FCFDFE} 69.14$ & $\cellcolor[HTML]{E9F0F7} 54.20$ & $\cellcolor[HTML]{FAFBFD} 71.62$ & $\cellcolor[HTML]{F2F6FB} 75.51$ \\
KLM~\cite{hendrycks2019_ood_mls} & $\cellcolor[HTML]{F1F5FA} 61.36$ & $\cellcolor[HTML]{FEFEFF} 80.25$ & $\cellcolor[HTML]{F0F4FA} 76.51$ & $\cellcolor[HTML]{E6EEF6} 51.45$ & $\cellcolor[HTML]{FAFBFD} 72.20$ & $\cellcolor[HTML]{D4E1F0} 84.90$ & $\cellcolor[HTML]{E2EAF5} 48.52$ & $\cellcolor[HTML]{F5F8FB} 64.91$ & $\cellcolor[HTML]{DAE5F2} 83.59$ & $\cellcolor[HTML]{EAF0F8} 54.58$ & $\cellcolor[HTML]{F8FAFD} 70.51$ & $\cellcolor[HTML]{EAF1F8} 78.68$ & $\cellcolor[HTML]{E9EFF7} 53.98$ & $\cellcolor[HTML]{FAFBFD} 71.97$ & $\cellcolor[HTML]{E4ECF5} 80.92$ \\
ASH-s~\cite{djurisic2023ash} & $\cellcolor[HTML]{F2F6FB} 62.47$ & $\cellcolor[HTML]{FEFFFF} 83.09$ & $\cellcolor[HTML]{EAF0F8} 78.92$ & $\cellcolor[HTML]{E3ECF5} 49.56$ & $\cellcolor[HTML]{FBFCFE} 74.63$ & $\cellcolor[HTML]{D1DFEF} 85.70$ & $\cellcolor[HTML]{EBF1F8} 55.66$ & $\cellcolor[HTML]{FAFCFD} 72.99$ & $\cellcolor[HTML]{F2F6FB} 75.31$ & $\cellcolor[HTML]{ECF2F8} 56.93$ & $\cellcolor[HTML]{FCFDFE} 76.21$ & $\cellcolor[HTML]{F7F9FC} 73.08$ & $\cellcolor[HTML]{EBF1F8} 56.15$ & $\cellcolor[HTML]{FCFDFE} 76.73$ & $\cellcolor[HTML]{EBF1F8} 78.25$ \\
GradNorm~\cite{huang2021_gradnorm} & $\cellcolor[HTML]{FBFCFD} 73.72$ & $\cellcolor[HTML]{FFFFFF} 88.56$ & $\cellcolor[HTML]{FBFCFE} 69.71$ & $\cellcolor[HTML]{F8FAFD} 69.32$ & $\cellcolor[HTML]{FEFFFF} 84.27$ & $\cellcolor[HTML]{F6F9FC} 73.69$ & $\cellcolor[HTML]{EBF1F8} 56.18$ & $\cellcolor[HTML]{F5F8FC} 65.91$ & $\cellcolor[HTML]{FFFFFF} 63.75$ & $\cellcolor[HTML]{F0F4FA} 60.16$ & $\cellcolor[HTML]{F8FAFD} 69.87$ & $\cellcolor[HTML]{FFFFFF} 61.34$ & $\cellcolor[HTML]{F5F8FB} 64.84$ & $\cellcolor[HTML]{FCFDFE} 77.15$ & $\cellcolor[HTML]{FDFEFE} 67.12$ \\
TempScale~\cite{Gua2017_tempscaling} & $\cellcolor[HTML]{F5F8FC} 66.08$ & $\cellcolor[HTML]{FFFFFF} 84.48$ & $\cellcolor[HTML]{E5EDF6} 80.44$ & $\cellcolor[HTML]{ECF2F8} 56.63$ & $\cellcolor[HTML]{FDFDFE} 77.54$ & $\cellcolor[HTML]{CDDCEE} 86.50$ & $\cellcolor[HTML]{EAF0F8} 54.44$ & $\cellcolor[HTML]{F9FBFD} 71.28$ & $\cellcolor[HTML]{E3EBF5} 81.24$ & $\cellcolor[HTML]{EEF3F9} 58.52$ & $\cellcolor[HTML]{FCFDFE} 75.53$ & $\cellcolor[HTML]{ECF2F8} 78.08$ & $\cellcolor[HTML]{EFF4F9} 58.92$ & $\cellcolor[HTML]{FCFDFE} 77.21$ & $\cellcolor[HTML]{E2EAF5} 81.57$ \\
MSP~\cite{hendrycks2016_ood_msp} & $\cellcolor[HTML]{F7FAFC} 68.81$ & $\cellcolor[HTML]{FFFFFF} 87.78$ & $\cellcolor[HTML]{E7EEF7} 79.90$ & $\cellcolor[HTML]{F0F5FA} 60.65$ & $\cellcolor[HTML]{FEFFFF} 82.50$ & $\cellcolor[HTML]{D0DEEE} 85.90$ & $\cellcolor[HTML]{EDF2F9} 57.20$ & $\cellcolor[HTML]{FCFDFE} 75.84$ & $\cellcolor[HTML]{E3ECF5} 81.02$ & $\cellcolor[HTML]{F2F6FA} 62.36$ & $\cellcolor[HTML]{FDFEFE} 79.69$ & $\cellcolor[HTML]{EDF2F9} 77.65$ & $\cellcolor[HTML]{F2F6FA} 62.26$ & $\cellcolor[HTML]{FEFEFF} 81.45$ & $\cellcolor[HTML]{E3ECF5} 81.12$ \\
ASH-b~\cite{djurisic2023ash} & $\cellcolor[HTML]{FFFFFF} 86.50$ & $\cellcolor[HTML]{FFFFFF} 95.60$ & $\cellcolor[HTML]{FEFFFF} 64.45$ & $\cellcolor[HTML]{FEFEFF} 80.47$ & $\cellcolor[HTML]{FFFFFF} 92.57$ & $\cellcolor[HTML]{FEFEFF} 66.32$ & $\cellcolor[HTML]{FEFFFF} 82.30$ & $\cellcolor[HTML]{FFFFFF} 86.63$ & $\cellcolor[HTML]{FFFFFF} 56.08$ & $\cellcolor[HTML]{FEFEFF} 81.82$ & $\cellcolor[HTML]{FFFFFF} 88.67$ & $\cellcolor[HTML]{FFFFFF} 53.50$ & $\cellcolor[HTML]{FEFFFF} 82.77$ & $\cellcolor[HTML]{FFFFFF} 90.87$ & $\cellcolor[HTML]{FFFFFF} 60.09$ \\
RankFeat~\cite{song2022rankfeat} & $\cellcolor[HTML]{FFFFFF} 92.35$ & $\cellcolor[HTML]{FEFFFF} 97.69$ & $\cellcolor[HTML]{FFFFFF} 50.49$ & $\cellcolor[HTML]{FFFFFF} 89.03$ & $\cellcolor[HTML]{FFFFFF} 95.33$ & $\cellcolor[HTML]{FFFFFF} 53.74$ & $\cellcolor[HTML]{FFFFFF} 92.55$ & $\cellcolor[HTML]{FEFFFF} 96.19$ & $\cellcolor[HTML]{FFFFFF} 35.92$ & $\cellcolor[HTML]{FFFFFF} 93.83$ & $\cellcolor[HTML]{FEFFFF} 97.21$ & $\cellcolor[HTML]{FFFFFF} 36.05$ & $\cellcolor[HTML]{FFFFFF} 91.94$ & $\cellcolor[HTML]{FEFFFF} 96.60$ & $\cellcolor[HTML]{FFFFFF} 44.05$ \\

\bottomrule
\end{tabular}
}
\caption{Average OOD detection performance per method and OOD category (near, far, extreme, and synthetic).
Results are reported as mean over three random seeds, two architectures (ResNet and CCT), and four tasks (Almond, Wheat, Kernels, and Food-grade).
Arrows ($\uparrow$/$\downarrow$) indicate whether higher or lower values are better.
Cells are color-coded from \textcolor{best_blue}{\textbf{blue}} (high performance) to white (low performance). 
Additionally, the \textbf{best} and \underline{second-best} results in each column are highlighted in bold and underlined, respectively. 
All values are reported as percentages, and methods are sorted based on their average False Positive Rate at 99\% True Positive Rate (FPR99) score.
}
\label{tab:ood_benchmark_avg_seed_arch}
\end{table*}

\subsection{Analysis}
\label{sec:exp:analysis}
\paragraph{ID Performance.}

As shown in \cref{tab:acc_ece}, both architectures achieve high accuracy ($93.18\%$ to $96.38\%$), indicating robust classification capabilities across all four ID tasks. 
Additionally, their Expected Calibration Error (ECE)~\cite{Naeini2015_ece, Gua2017_tempscaling} is consistently low ($0.44\%$ to $4.75\%$), indicating reliable confidence estimates.
Notably, ResNet models slightly outperform CCT models in both accuracy and calibration. 
These results provide a solid basis for the OOD evaluation.

\paragraph{Comparison of OOD detection methods.}

\cref{tab:ood_benchmark_avg_seed_arch} summarizes the average OOD performance across all ID tasks and OOD categories.
Our benchmark reveals a clear and systematic performance hierarchy: methods utilizing features from intermediate or penultimate layers (\eg, GRAM, ATS, ViM, KNN) consistently outperform both widely used logit-based baselines (\eg, MSP, MLS) and modern model enhancement techniques (\eg, SCALE, ASH, DICE). 
GRAM, in particular, stands out as the best-performing method across all OOD categories and metrics.
Nevertheless, its FPR remains prohibitively high for near- and far-OOD samples, limiting its real-world applicability.

This performance hierarchy points to a direct link between dataset characteristics and method suitability. 
We find that training on ICONIC-444's controlled, low-variance, and fine-grained industrial data (object-centric nature)
creates more statistically coherent feature spaces, a conclusion supported by our in-depth feature space analysis in the supplement (App.~C.3, Supp.~Fig. 8).
This structure intrinsically favors feature-based methods, while the high data variability of general-purpose (scene-centric) benchmarks like ImageNet creates a scenario where model enhancement techniques, which are often developed and validated on ImageNet itself, are more robust.
Our cross-benchmark analysis empirically confirms this divergence, positions ICONIC-444 as a methodologically clean complement for existing large-scale benchmarks (\cf Supp.~Fig.~7), providing the first controlled environment to assess robustness in fine-grained, industrial settings---a critical aspect of OOD performance not otherwise covered.
However, the benchmark remains challenging; alarmingly, the FPRs of most methods exceed 30\% on extreme- and synthetic-OOD samples at 99\% TPR level, with only GRAM, ATS, and ViM achieving lower FPRs in these "easy" categories---highlighting that current methods remain far from fully reliable.
As detailed in the supplement (Section C--E), this challenge persists even for larger and more complex architectures like ConvNeXT~\cite{Liu_2022_CVPR}, ViTs~\cite{dosovitskiy2021vit}, and CLIP~\cite{radford2021_clip}, which also suffer from high FPRs and do not offer a clear performance advantage, especially on the more difficult near- and far-OOD categories.
These results reinforce findings from Zhang \etal~\cite{zhang2024openood} that increased model complexity does not necessarily improve OOD detection performance.

\begin{figure}
\centering
\includegraphics[width=\linewidth]{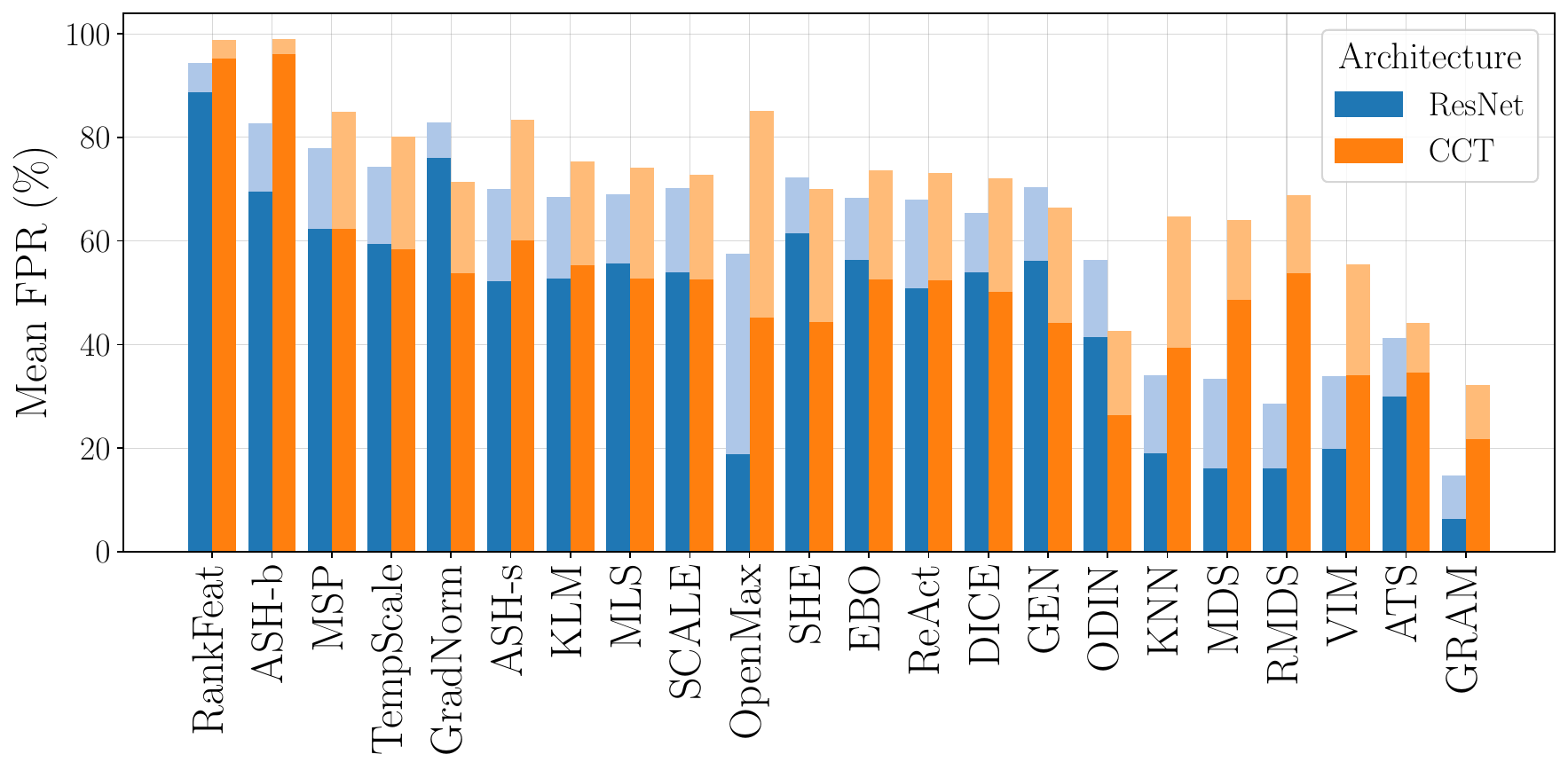}
\caption{
Comparison of the mean FPR95 and FPR99 for each OOD detection method, averaged over three random seeds, all four tasks, and all four OOD categories.
Darker shades present the FPR at $95\%$ TPR, while the lighter shades show the gap to the more challenging (but highly relevant in practice) FPR at $99\%$ TPR.
}
\label{fig:mean_fpr_95_99_with_arch}
\end{figure}

\cref{fig:mean_fpr_95_99_with_arch} shows a detailed comparison of FPR95 and FPR99 across both architectures.
The performance differences reveal architecture-specific behaviors in OOD detection methods:
Methods like MDS, RMDS, ODIN, OpenMax, and KNN exhibit substantial performance variations, with ResNet18 generally outperforming CCT.
Also, for GRAM, ResNet18 achieves more than double the performance of CCT in the FPR99 setting but still consistently demonstrates top-tier results on both architectures. 
Model enhancement methods such as ASH, SCALE, DICE, and the baseline methods MLS, EBO, and MSP show relatively stable performance across architectures with minimal variation.
Interestingly, the average performance drop from FPR95 to FPR99 across all methods is more pronounced for CCT ($19.04\%$) compared to ResNet18 ($14.45\%$).

\paragraph{Practical Importance of FPR99.}
In safety-critical or resource-intensive applications, OOD detectors must strictly control false positives while maintaining very high true positive rates.
Thus, FPR99, which evaluates OOD detection performance at 99\% TPR, is therefore crucial: any drop in TPR wastes valuable ID samples, while high FPR99 allows dangerous or undesirable OOD samples (\eg, contaminants, foreign objects) to pass through, posing serious safety and sustainability risks.

High AUROC values alone can be misleading in practical applications. 
While AUROC values (90\%+) may indicate good overall separability, they do not guarantee low false positive rates at critical operating points. 
ICONIC-444 proves to be a particularly challenging benchmark, with AUROC ranges comparable to ImageNet-based benchmarks (near-OOD: [50.99\%, 81.36\%]; far-OOD: [53.93\%, 96.53\%])~\cite{zhang2024openood, xu2024scaling}, despite the fact that ImageNet OOD test sets contain ID contamination, artificially deflating AUROC scores. 
Our results clearly indicate that existing methods struggle with maintaining low false positive rates at high true positive rates (FPR95 and especially FPR99).
This shows that reliable OOD detection remains unsolved, underscoring ICONIC-444's importance for future research.

\begin{figure}
\centering
\includegraphics[width=\linewidth]{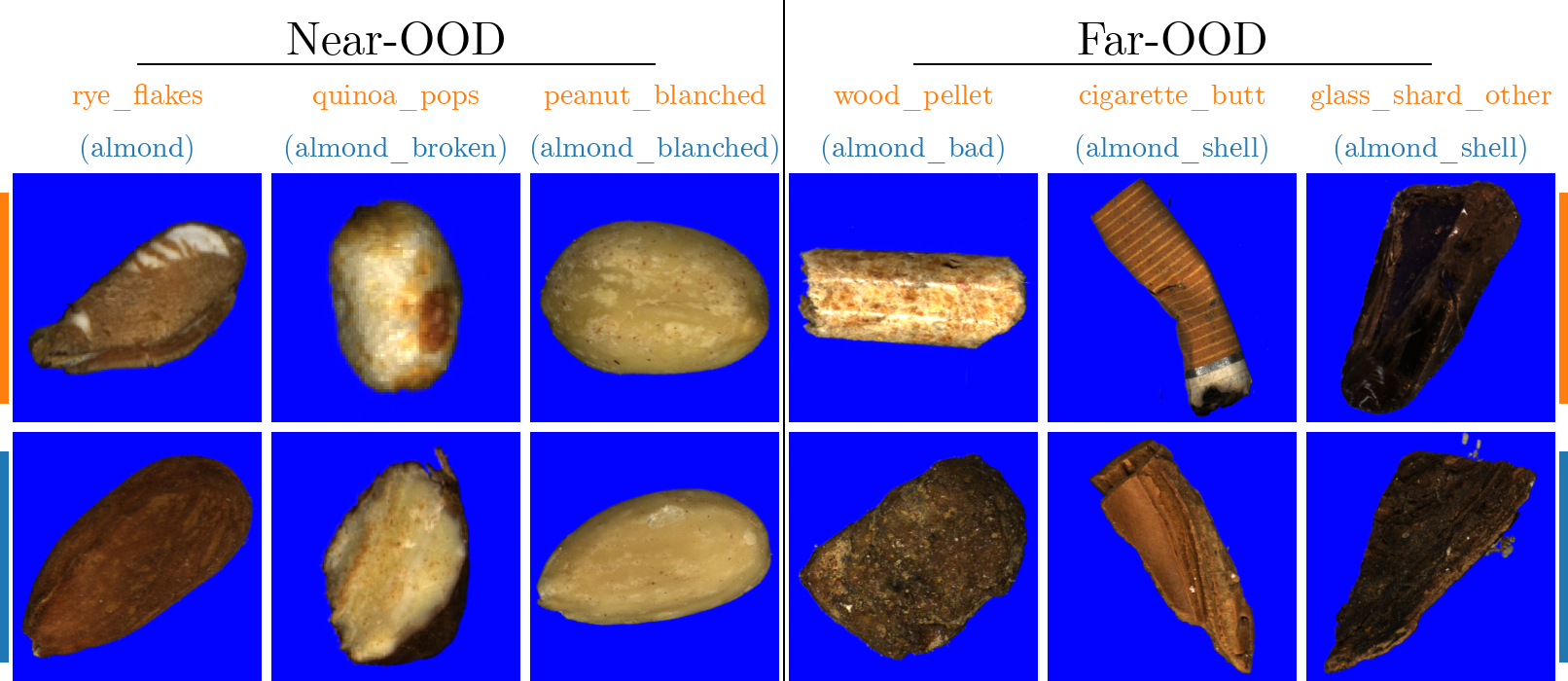}
\caption{
Difficult near- and far-OOD samples for GRAM~\cite{sastry20a_gram} on the Almond task (ResNet18).
The first row shows example images from the \textcolor{fig_orange}{OOD classes} that are most frequently misclassified as ID, while the second row presents the closest (in feature space) \textcolor{fig_blue}{ID (almond) classes} causing these confusions.
}
\label{fig:difficutl_ood_classes}
\end{figure}

\paragraph{Examples of Difficult OOD Cases.}
\cref{fig:difficutl_ood_classes} shows samples from difficult near- and far-OOD classes for GRAM (best-performing OOD detector in terms of mean FPR) on the Almond task.
For each hard OOD sample, we also show the nearest ID sample (via penultimate-layer feature distance).
The visualizations reveal two key sources of confusion: visual similarity (appearance, texture) and structural similarity (overall shape).
GRAM especially struggles with near-OOD classes, such as \textit{rye flakes} (93.4\% FPR95) and \textit{peanut blanched} (79.4\% FPR95). 
Similar patterns emerge for far-OOD samples, where visually dissimilar yet structurally related objects like \textit{glass shard other} are confused with \textit{almond shell} (31.6\% FPR95).
From a human perspective, these confusions are understandable due to subtle visual distinctions; however, the method also struggles with clearly distinguishable classes (\eg, \textit{wood pellet}, 90.6\% FPR95; \textit{cigarette butt}, 57.6\% FPR95), underscoring that substantial unresolved issues persist in robust OOD detection.

\section{Conclusion}
\label{sec:conclusion}
In this paper, we introduce ICONIC-444, a structured image classification dataset created in a controlled industrial environment, specifically with out-of-distribution (OOD) detection research and evaluation in mind. 
ICONIC-444 is designed to complement existing OOD benchmarks by introducing a new domain with well-defined, contamination-free class boundaries, enabling seamless construction of ID and OOD splits without overlap and allowing researchers to develop and evaluate methods across a wide range of scenarios.
Our extensive experiments on multiple state-of-the-art OOD detection methods demonstrate performance variability based on data, architecture, and OOD complexity, highlighting limitations in applicability. 
We believe the ICONIC-444 dataset will help our research community gain valuable insights and enhance the practical use of OOD detection methods in diverse, real-world settings.

\textbf{Acknowledgments}
We acknowledge support from KESTRELEYE GmbH, which provided the core infrastructure and resources throughout the project.
We also thank Ioannis Weber-Konstantellos for his extensive assistance during data collection, including laboratory support, data cleaning, and object preparation.

{
    \small
    \bibliographystyle{ieeenat_fullname}
    \bibliography{main}
}

\appendix
\clearpage
\maketitlesupplementary

\crefname{appendix}{Appendix}{Appendices}
\Crefname{appendix}{Appendix}{Appendices}
\crefalias{section}{appendix}

In the following, we provide detailed insights into our dataset (\cref{sec:supp:daqd}), covering its acquisition setup, process, and quality assurance protocols. 
Additionally, we outline the experimental setup (\cref{sec:supp:ed}) and provide detailed evaluations (\cref{sec:supp:detailed_results}) that complement the results reported in the main manuscript.
We extend our analysis to larger and more complex architectures (\cref{sec:supp:eval_modern_arch}), explore OOD detection performance using a vision language model (VLM) (\cref{sec:supp:eval_clip}), and conduct ablation studies
regarding the correlation between ID and OOD performance (\cref{sec:supp:io_ood_cor}) and the impact of data corruption (\cref{sec:supp:data_corruption}).

\section{Dataset and Quality Details}
\label{sec:supp:daqd}
Ensuring the highest quality of recordings and maintaining class purity is critical to our dataset's utility and, thus, a high priority during data acquisition. 
We follow a stringent acquisition protocol and implement comprehensive data cleaning and quality control measures to ensure the data meets rigorous standards for accuracy and consistency.

\begin{figure}[t]
 \centering
    \begin{subfigure}[b]{\linewidth}
        \centering
        \includegraphics[width=\linewidth]{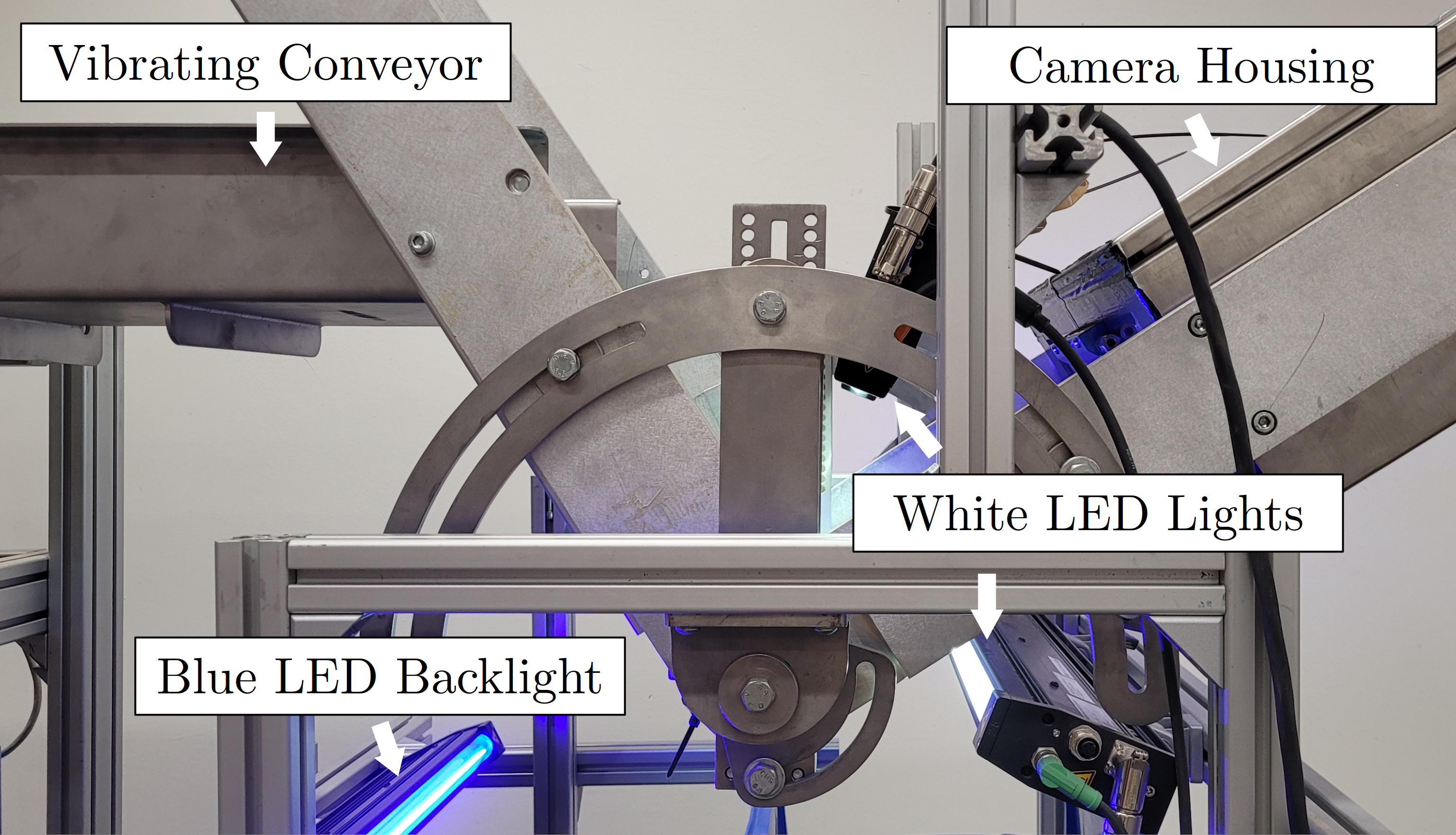}
        \caption{}
    \end{subfigure}
    \hfill
    \vspace{1pt} %
    \begin{subfigure}[b]{\linewidth}
        \centering
        \includegraphics[width=\linewidth]{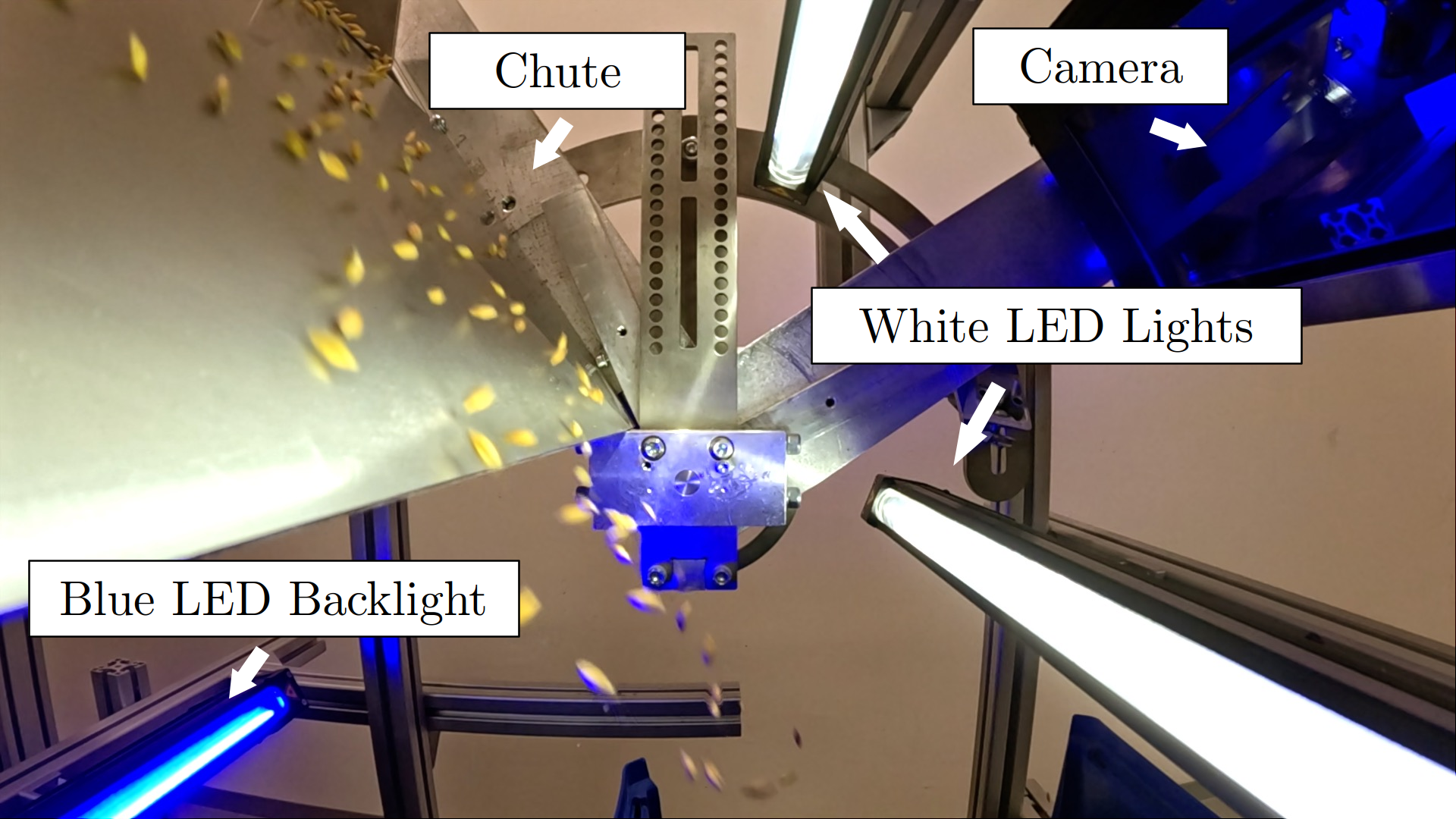}
        \caption{}
    \end{subfigure}
\caption{
Views of the sorting machine prototype.
\textbf{(a)} External overview, highlighting the vibrating conveyor, blue LED backlight, white LED lights, and camera housing.
\textbf{(b)} Internal perspective, illustrating the blue LED backlight, white LED lights, chute, and camera within the machine's main structure.
}
\label{fig:machine_setup}
\end{figure}

\subsection{System Setup}
Our data is acquired using a sorting machine prototype designed to closely mimic real-world applications. 
The setup (see \cref{fig:machine_setup}) includes key components such as a vibrating conveyor, chute, camera, and lighting configuration, all optimized for capturing high-quality images of food objects, which serve as the in-distribution (ID) classes in our defined benchmark tasks.

The chute is adjusted to a 60-degree slope, and the camera is positioned at a 90-degree angle to the chute, ensuring it is perpendicular to the object's motion for optimal image capture. 
The camera's scan line is located just beyond the chute endpoint, capturing the objects as they enter free fall. 
The lighting system ensures uniform illumination and minimizes shadows.
A blue LED backlight, aligned with the camera's scan line, provides consistent contrast, while white LED lights at 45-degree angles above and below the camera, and focused on the scan line, illuminate the objects from multiple directions.
The selection of a blue backlight is a deliberate, industry-driven choice.  
Uniform, narrow-band blue illumination provides high, stable contrast against the color spectrum of virtually all food items, supporting reliable and computationally efficient foreground--background segmentation in high-throughput bulk sorting.
Camera settings, such as exposure, gains, aperture, and focus, are calibrated to capture high-resolution images with balanced brightness and contrast, tailored to the reflective properties and appearances of food items.

We capture our non-food classes using the same setup, which may result in overexposure, underexposure, reflections, or other artifacts, particularly for transparent objects like glass. 
However, we deem this typical for out-of-distribution data: since the system is optimized for food sorting, such effects are inevitable when OOD objects fall through the machine, making them perfectly normal and expected for OOD data.
Including these classes in our ICONIC-444, we ensure the dataset reflects real-world constraints, where systems tailored to specific tasks naturally encounter such artifacts.
\cref{fig:iconic_1_3,fig:iconic_2_3,fig:iconic_3_3} show sample images representing all classes included in ICONIC-444.

To clearly understand our setup and acquisition process, we also include a supplementary video showing the sorting machine prototype and sample recordings.

\subsection{Acquisition Protocol}

\paragraph{Setup verification.} 
We use a white balance target as a reference to verify camera calibration and ensure consistent illumination across all recordings. 
The target's intensity profiles are compared against initially saved references to detect deviations in lighting, camera settings, or hardware configuration.
These checks ensure uniformity across acquisition sessions, preventing data quality shifts due to environmental factors or subtle hardware changes.

\paragraph{Acquisition process.}
We follow a sequential acquisition process to mitigate the risk of environmental contamination and ensure the integrity of all classes.
Each acquisition session is organized into groups, with one class prepared and recorded at a time.
This approach is essential for fine-grained classes within the same group, where cross-contamination between classes is nearly impossible to detect and rectify post-acquisition.
Objects are placed on a vibrating conveyor and manually inspected to ensure the removal of rare foreign materials as an additional precaution to maintain the dataset's high cleanliness standards. 
After completing the acquisition, we thoroughly clean the machine to prevent cross-contamination between different classes.
During this procedure, we inspect the system for residual particles (\eg, dust or debris) that could affect subsequent classes.

\subsection{Data Cleaning and Quality Control}

We implement a comprehensive data cleaning and quality control process after acquisition to remove poorly acquired data and further improve the quality of our dataset.
These procedures are designed to exclude contaminants and eliminate issues such as objects outside the illumination range or objects that are out of focus.

Certain classes, for example in the dried fruits, spices, or pulses groups, naturally contain contaminants like broken stems or pieces of skin that can separate from the main objects. 
These classes require more stringent cleaning procedures to ensure that any environmental or natural contaminants are systematically removed.
The following describes the steps we implemented in the data cleaning and quality control process to achieve these goals.

Our semi-automatic data cleaning process is performed on a class-by-class basis and follows a structured, iterative approach.
The goal is to ensure that each class in the dataset is free from contamination:

For each class, we first extract a set of handcrafted features, such as area, the mean and standard deviation of each color channel, bounding box dimensions, aspect ratio, and object brightness. 
Additionally, we compute feature vectors from the penultimate layer of a ResNet50 model pre-trained on ImageNet to capture high-level semantic information.

The cleaning process begins by identifying extreme values (both minimum and maximum) for these handcrafted features.
A class-specific threshold is set for selecting a fixed percentage of these extreme cases, which are then manually inspected.
This process is iterative, continuing until no outliers or contaminants remain. 
Throughout this step, images are labeled as either \textit{accepted} or \textit{rejected}. 

Next, we use the feature vectors to calculate the \textit{k}-th nearest neighbor distance for all non-\emph{rejected} images.
Again, thresholds are determined on a class-by-class basis, with a percentage of images having the largest distances manually reviewed. 
This step is repeated until no further outliers are detected.

Once the initial cleaning is complete, we train a random forest classifier for each class using the manually labeled \textit{accepted} and \textit{rejected} images based on the extracted feature vectors. 
This classifier is then used to predict the status of images that were not manually inspected. 
We manually review all images classified as \textit{rejected} to ensure no contaminations remain.

For classes where manual defects are introduced, all images are manually inspected to ensure the intended defects are accurately captured and visible in the image.

To further ensure the dataset is free from critical cross-contaminations and mislabels, we conduct a final verification step aimed at minimizing human error, especially in closely related classes. 
We organize classes into subgroups (\eg, pasta, beans, nuts) and perform a cross-validation procedure by splitting the data $50/50$ within each subgroup, training a ResNet18 on one half, and evaluating it on the other. 
This allows us to detect potential cross-contaminations, particularly in closely related classes such as \textit{nuts acorn} and \textit{nuts acorn damaged} or \textit{kernel sunflower peeled} and \textit{kernel sunflower white}. 
Incorrectly classified images are manually reviewed.
Note that this is only possible for classes where manual inspection can reveal the true class of the object---for example, distinguishing between different breeds of wheat is not feasible for us. 
However, due to the quality protocols of the producers from which we sourced them, we are very confident in the high purity of these classes.

Finally, all data undergoes an independent review by individuals not involved in the acquisition or cleaning process. 
This additional verification step helps identify and quantify errors introduced during cleaning, ensuring the dataset is of high quality and reliability.
\cref{fig:contamination_samples} shows classes with natural contaminants, such as broken stems from dried fruits or shells from beans and lentils, which were flagged and rejected during the cleaning process. 

\subsection{Environmental Noise}
A further test we conduct focuses on detecting potential background noise and changes in the data acquisition setup between classes. Such changes could be caused by environmental factors, such as light sources or shadows, and might allow models to distinguish between objects based on non-object features.

For this test, we focus on the Almond task and create three variations of the dataset. 
The first version consists of the original images, serving as the control set. 
In the second version, we segment the objects and replace the background with a uniform blue color to eliminate any potential background patterns caused by environmental effects. 
In the third version, we mask the objects entirely, replacing them with a uniform blue color to test whether the model is still able to make correct predictions based solely on background patterns.

\begin{table}
\centering
\footnotesize
\begin{tabular}{lccc}
\toprule
\multirow{2}{*}{\textbf{Trainset}} & \multicolumn{3}{c}{\textbf{Testset}} \\
\cmidrule(lr){2-4}
& Almond & Almond w/o Bg & Almond w/o Fg \\
\midrule
Almond & 95.60 & 91.35 & 13.97 \\ 
Almond w/o Bg & 95.79 & 94.97 & 13.97 \\ 
Almond w/o Fg & 18.81 & 18.81 & 18.81 \\ 
\bottomrule

\end{tabular}
\caption{
ResNet18 model accuracy (in percentage) for the Almond task, trained and evaluated on three dataset versions: with background (Almond), with the background removed (w/o Bg), and with the foreground removed (w/o Fg). 
Each model was trained on a specific version and tested across all three to evaluate the influence of background noise on classification performance.
}
\label{tab:almond_bg_pattern_analysis}
\end{table}

We train ResNet18 models on each version of the dataset using the same training parameters as described in the main manuscript and evaluate their performance across all three dataset variations. 
\cref{tab:almond_bg_pattern_analysis} shows that models trained on the original Almond dataset and the dataset with the background removed (w/o Bg) achieve similarly high accuracy when tested on both the original and w/o Bg test sets. 
This indicates that the model does not use background features to classify objects.

As expected, both models achieve only around $14\%$ accuracy on the w/o Fg (foreground removed) dataset, confirming that the informative content is solely in the foreground, with no meaningful background patterns. 
The model trained on the w/o Fg dataset performs poorly across all test sets, further demonstrating that background information offers no useful guidance for predictions.
The cross-entropy loss stays near $\log(N)$ (where $N$ is the number of ID classes), indicating the model is merely guessing and unable to learn meaningful representations. 

Tests on other ID data, such as the wheat classes, show similar results, confirming that our data is free from information leakage in the background and that data quality is consistent across the different classes in our dataset.

\subsection{Limitations}
Please note that it is virtually impossible to ensure correct labeling for every single image in our dataset. 
Manually checking over $3.1$ million images is impractical and would not guarantee proper labeling due to human error and the difficulty of visually distinguishing closely related classes, such as different grain varieties.

To estimate the labeling accuracy, we conduct an additional manual inspection of 144 images per class across all 444 classes, totaling 63936 images (approximately $2\%$ of the dataset).
This inspection is performed by individuals not involved in the data acquisition or cleaning process to minimize bias. 
Across all classes, we identify a total of 25 images containing unexpected objects, which indicates that the fraction of mislabeled data within our dataset is well below $0.05\%$.
These unexpected objects are not due to cross-contamination between different classes --- thanks to our stringent sourcing and acquisition protocols, cross-contamination is extremely rare or nonexistent. 
Instead, the samples found are primarily natural contaminants (\eg, dry fruit skin with high visual similarity to the actual dry fruit).

\subsection{Societal and Ethical Impact}
Our dataset has been constructed with careful attention to ethical considerations.
No personal data were collected, processed, or included in the construction of this dataset. 
While comprehensive, the dataset cannot consider all possible real-world scenarios, and it tends to focus on foods commonly used in specific regions of the world. 
OOD detection methods tested on our dataset may not account for every potential OOD instance encountered in practice. 
However, we consider the impact of these limitations relatively minor because we took great care of sourcing a wide variety of different products for ICONIC-444.

\section{Experimental Details}
\label{sec:supp:ed}

\subsection{Implementation Details}
\paragraph{Software setup.}
Our implementation builds on the OpenOOD framework~\cite{yang2022openood, zhang2024openood}, extended by~\cite{Humblot-Renaux_2024_CVPR} (modified GRAM~\cite{sastry20a_gram} implementation and added support for CompactTransformer~\cite{hassani202_cct}).
OpenOOD currently serves as the most comprehensive framework for evaluating state-of-the-art OOD detection methods. 
Our main modifications include: (i) integrating our four benchmark tasks into the framework, (ii) adding support for evaluating custom-defined OOD groups, (iii) enabling the evaluation of custom metrics, (iv) incorporating ATS~\cite{Krumpl2024_ats} as an additional post-hoc OOD detection method, (v) adding the option to generate synthetic OOD samples for in-distribution (ID) datasets (building on the codebase of Bitterwolf \etal~\cite{bitterwolf2023_ninco}), and providing support for models from the timm-repository~\cite{rw2019timm} (\eg ConvNeXt, ViTs).
For CLIP-based OOD detection methods (MCM~\cite{ming2022mcm}, GL-MCM~\cite{Miyai2023GLMCMGA}, GalLop~\cite{lafon2024gallop}), we extend their original repositories.

\paragraph{Computational environment.}
We run our experiments on a server equipped with an Intel(R) Core(TM) i9-9900X CPU @ 3.50GHz, paired with three NVIDIA GeForce RTX A4000 GPUs. 
The computational setup operates on Ubuntu 22.04, incorporating PyTorch 2.0.1, and leverages CUDA 11.8 and cuDNN 8.7.0. 

\begin{table}
\centering
\footnotesize
\setlength{\tabcolsep}{4pt} 
\resizebox{\linewidth}{!}{
\begin{tabular}{lcccc}
\textbf{Model} & \textbf{Pretrained} & \textbf{Params} & \textbf{Inference Time} & \textbf{timm name} \\
\midrule
CCT\_7\_7x2 & ---           & \phantom{0}4.5M   & \phantom{0}2.65ms    & --- \\
ResNet18    & ---           & 11.6M             & \phantom{0}2.41ms    & --- \\
ConvNeXt-P  & ImageNet-1k   & \phantom{0}9.0M   & \phantom{0}2.41ms    & convnext\_pico.d1\_in1k  \\
ConvNeXt-T  & ImageNet-12k  & 28.6M             & \phantom{0}3.69ms    & convnext\_tiny.in12k\_ft\_in1k \\         
ViT-S/16    & ImageNet-21k  & 22.1M             & \phantom{0}3.68ms    & vit\_small\_patch16\_224.augreg\_in21k\_ft\_in1k \\
ViT-B/16    & ImageNet-21k  & 86.6M             & \phantom{0}4.44ms    & vit\_base\_patch16\_224.augreg2\_in21k\_ft\_in1k \\ 
\midrule
CLIP-B/16   & openai        & 150M              & 12.71ms          & --- \\
\bottomrule

\end{tabular}
}
\caption{
Overview of the evaluated models.
}
\label{tab:model_overview}
\end{table}

\subsection{Training}
Our main evaluations from the main manuscript focus on two architectures: a Convolutional Neural Network (CNN) using ResNet18~\cite{He2015_resnet} and a Transformer using the Compact Transformer (CCT\_7\_7x2\_224)~\cite{hassani202_cct}. 
These models were selected for their efficiency, effectiveness, and practicality while balancing performance, runtime efficiency, and computational cost.
This makes them particularly suitable for real-world applications where cloud computation is not feasible due to latency constraints, such as real-time quality control, automated sorting systems, and embedded AI applications in industrial environments. 
By ensuring both research accessibility and deployment readiness, these models provide a strong foundation for OOD detection across diverse settings.

To further broaden architectural diversity, we additionally evaluate two larger and more complex architectures: ConvNeXt~\cite{Liu_2022_CVPR} and Vision Transformers~\cite{dosovitskiy2021vit} (ViT).
For ConvNeXt, we include both a compact variant (ConvNeXt-P, or "pico"), which has a parameter count comparable to ResNet18, as well as the larger ConvNeXt-T.
Additionally, we evaluate two ViT models, ViT-S and ViT-B. 
Unlike ResNet18 and CCT, these models leverage pre-trained weights for fine-tuning---to mitigate overfitting---except for ConvNeXt-P, which is trained from scratch to match the training paradigm of our baseline models.
\cref{tab:model_overview} gives an overview of the evaluated models.
All pre-trained model weights were taken from the public available timm-repository~\cite{rw2019timm}, except for the zero-shot CLIP model, which is available via github\footnote{https://github.com/openai/CLIP}.

Unless stated otherwise, all models are trained from scratch.
ResNet18 models are trained for 100 epochs using the Stochastic Gradient Descent (SGD) optimizer with an initial learning rate of \num{1e-1} and cosine annealing with a weight decay of \num{5e-3}. 
Compact Transformer models are trained for 150 epochs using the AdamW~\cite{loshchilov2018decoupled} optimizer with a learning rate of \num{5e-4} and a weight decay of \num{6e-2}. 
ConvNeXt-P follows the same setup and is trained from scratch for 100 epochs, while ConvNeXt-T, ViT-S, and ViT-B are fine-tuned for 50 epochs using AdamW with the same learning rate and weight decay as CCT.
To mitigate class imbalance, we use per-class weights in the cross-entropy loss, with $w_c=1/f_c$ where $f_c$ is the class frequency on the training split; weights are normalized to have mean $1$. 
The same scheme is applied for all training recipes and fine-tuning runs.

\paragraph{OOD detection methods.}
Since our dataset differs significantly from commonly used benchmarks like CIFAR10/100~\cite{krizhevsky09cifar} and ImageNet~\cite{deng2009_imagenet}, on which most OOD detection methods are developed, and given that some of the evaluated methods are sensitive to hyperparameters, we optimized these methods using both the ID and our specifically designed OOD validation set. 
\cref{tab:ood_hyperparameter} shows the 22 methods along with their respective hyperparameter search spaces, where applicable.

\begin{table}
\centering
\footnotesize
\resizebox{\linewidth}{!}{
\begin{tabular}{ll}
\toprule
\textbf{Method} & \multicolumn{1}{c}{\textbf{Hyperparameter Search Space}} \\
\midrule
\rowcolor{gray!10} GRAM \cite{sastry20a_gram}               &   \\ 
ViM~\cite{wang2022_ood_vim}             &   \\
\rowcolor{gray!10} KNN~\cite{sun2022knnood}                & $K \in \{50, 100, 200, 500, 1000\}$ \\
OpenMax~\cite{Bendale2016_openmax}      &   \\
\rowcolor{gray!10} ATS~\cite{Krumpl2024_ats}               &   \\
MDS~\cite{lee2018_mahala}           &   \\
\rowcolor{gray!10}
    & temperature $T \in \{1, 10, 100, 1000\}$  \\
  \multirow{-2}{*}{\cellcolor{gray!10}ODIN~\cite{shiyu17}}                   & \cellcolor{gray!10}perturbation mag. $\sigma \in \{0.0, 0.0007, 0.0014, 0.0028\}$ \\
RMDS~\cite{Ren2021_rmds}                &   \\
\rowcolor{gray!10}ReAct~\cite{sun2021_ood_react}          & percentile $\in \{70, 80, 85, 90, 95, 99\}$  \\
\multirow{2}{*}{GEN~\cite{Liu2023_GEN}} & gamma $\in \{0.01, 0.1, 0.5, 1, 2, 5, 10\}$  \\
                                        & top-M classes $\in \{2, 4, 5, 10, 15, 20, 25, 50, 100, 200, 300\}$ \\
\rowcolor{gray!10}TempScale~\cite{Gua2017_tempscaling}    &   \\
MSP~\cite{hendrycks2016_ood_msp}        &   \\
\rowcolor{gray!10}KLM~\cite{hendrycks2019_ood_mls}        &   \\
SCALE~\cite{xu2024scaling}              & percentile $\in \{65, 70, 75, 80, 85, 90, 95\}$   \\
\rowcolor{gray!10}ASH-s~\cite{djurisic2023ash}            & percentile $\in \{65, 70, 75, 80, 85, 90, 95\}$  \\
DICE~\cite{sun2022dice}                 & percentile $\in \{10, 30, 50, 70, 90\}$  \\
\rowcolor{gray!10}MLS~\cite{hendrycks2019_ood_mls}        &   \\
EBO~\cite{liu2020_ood_ebo}              &   \\
\rowcolor{gray!10}SHE~\cite{zhang2023_she}                &   \\
GradNorm~\cite{huang2021_gradnorm}      &   \\
\rowcolor{gray!10}ASH-b~\cite{djurisic2023ash}            & percentile $\in \{65, 70, 75, 80, 85, 90, 95\}$  \\
RankFeat~\cite{song2022rankfeat}        &   \\
\bottomrule
\end{tabular}
}
\caption{
Hyperparameter search space for each OOD detection method.
}
\label{tab:ood_hyperparameter}
\end{table}

\subsection{Extreme- and Synthetic-OOD Details}

To provide a comprehensive and diverse OOD test set, we add extreme and synthetic OOD samples in addition to the near- and far-OOD samples sourced from ICONIC-444:

\begin{figure*}
\centering
\includegraphics[width=\linewidth]{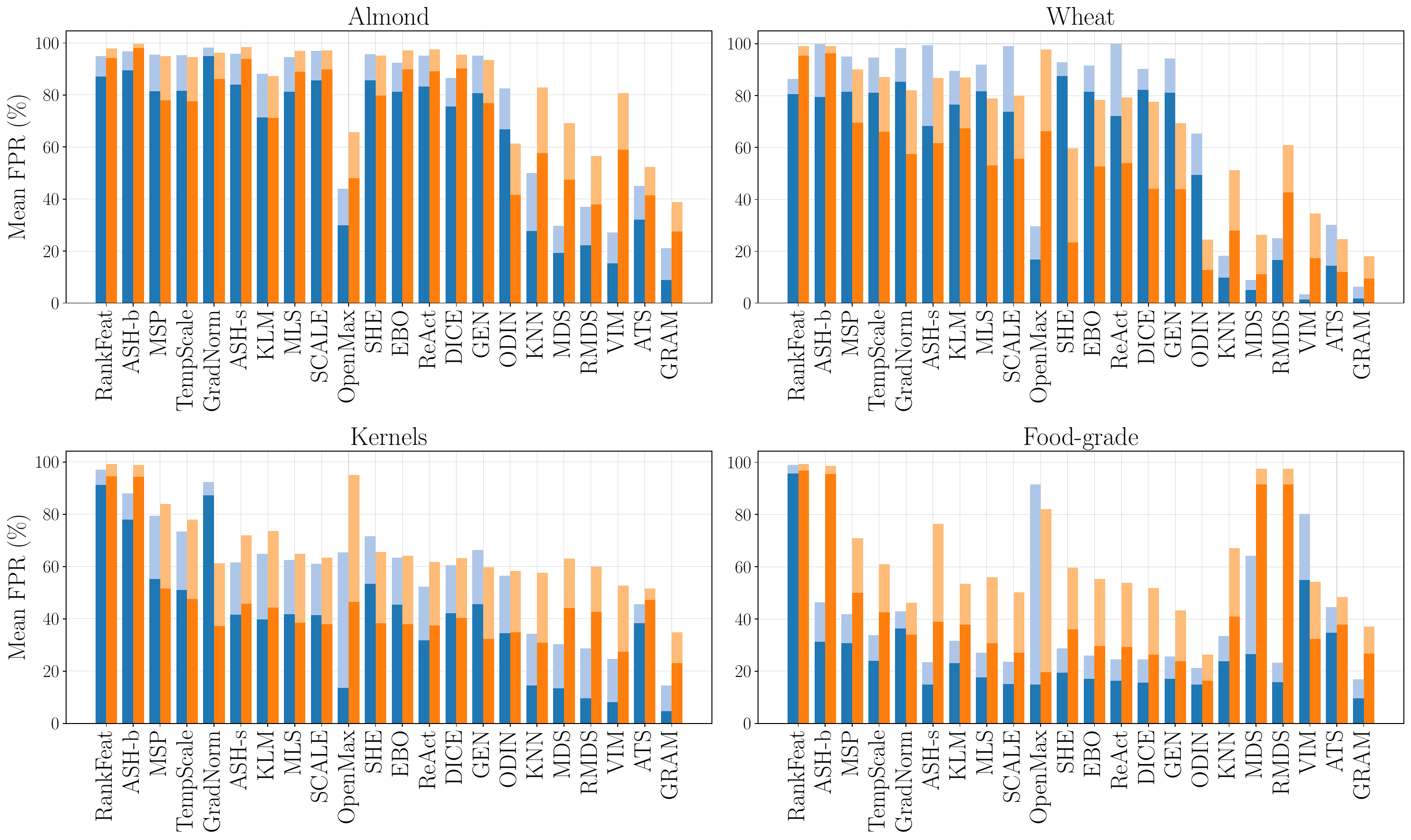}
\caption{
Comparison of mean FPR at $95\%$ and $99\%$ TPR across all methods for each individual task (Almond, Wheat, Kernels, and Food-grade) and architectures (ResNet18 and CCT), averaged over three runs and our four OOD categories (near, far, extreme and synthetic).
The blue and orange bars represent the ResNet18 and CCT architectures, respectively.
For each method, the dark-shaded portion of the bar indicates the mean FPR95, and the light-shaded segment on top represents the additional increase up to FPR99.
Methods are ordered by their overall mean FPR99 score across tasks, reflecting their relative performance ranking.
}
\label{fig:detailed_results_consider_task_and_arch}
\end{figure*}

\paragraph{Extreme-OOD.}
We add samples from four commonly known datasets---ImageNet~\cite{Nguyen2014}, iNaturalist~\cite{Horn_2018_inat}, Places365~\cite{zhou2017places}, and Textures~\cite{cimpoi14dtd}---to our OOD test set and categorize them as extreme-OOD. 
These samples are coarse-grained OOD examples compared to the ICONIC-444 data (and thus to all considered ID tasks), allowing our benchmark to test OOD detection performance on samples with no semantic correlation to the ID data. 
\cref{fig:extreme_ood_samples} shows sample images from all four datasets.

\paragraph{Synthetic-OOD.}
We generate 500 samples each for 25 distinct synthetic OOD types.
Out of these, 16 types are taken from literature, \ie uniform noise~\cite{hendrycks2016_ood_msp}, Gaussian noise~\cite{hendrycks2016_ood_msp}, Rademacher noise~\cite{hendrycks2019oe}, IN pixel permutations~\cite{hein2019relu}, black~\cite{bitterwolf2023_ninco}, white~\cite{bitterwolf2023_ninco}, monochrome~\cite{bitterwolf2023_ninco}, tricolour~\cite{bitterwolf2023_ninco}, horizontal stripes~\cite{bitterwolf2023_ninco}, vertical stripes~\cite{bitterwolf2023_ninco}, smooth noise~\cite{hein2019relu, bitterwolf2020_card, Meinke2022_provability_ard}, smooth noise+~\cite{bitterwolf2023_ninco}, smooth color~\cite{bitterwolf2023_ninco}, smooth IN pixel permutations~\cite{hein2019relu}, and blobs~\cite{hendrycks2019oe}. For these, we adhere to the configuration settings detailed by Bitterwolf \etal~\cite{bitterwolf2023_ninco} to ensure consistency and reproducibility in our comparisons.
Additionally, we extend the synthetic OOD dataset with variants more closely aligned with the ICONIC-444 dataset. 
Below, we describe these newly introduced types in further detail:

\begin{itemize}
\item \textbf{IN channel permutations:} We choose a random image from the ID data and randomly shuffle its color channels.
\item \textbf{red:} The red channel is set to $1.0$ the others to $0$.
\item \textbf{green} The green channel is set to $1.0$ the others to $0$.
\item \textbf{blue:} The blue channel is set to $1.0$ the others to $0$.
\item \textbf{random shape:} The background of the image is set to blue (with the blue channel at $1.0$ and the others at $0.0$). 
A random RGB color is uniformly sampled from the range $[0.4, 0.78]^3$, and the red and blue channels are shifted by $0.15$ and $-0.3$, respectively. 
A random shape (triangle, quadrangle, pentagon, octagon, or ellipse) is then drawn using the pre-generated color.
\end{itemize}
\cref{fig:synthetic_ood_samples} provides examples of each of the 25 synthetic OOD types used in this study.

\begin{figure}[t]
\centering
\includegraphics[width=0.92\linewidth]{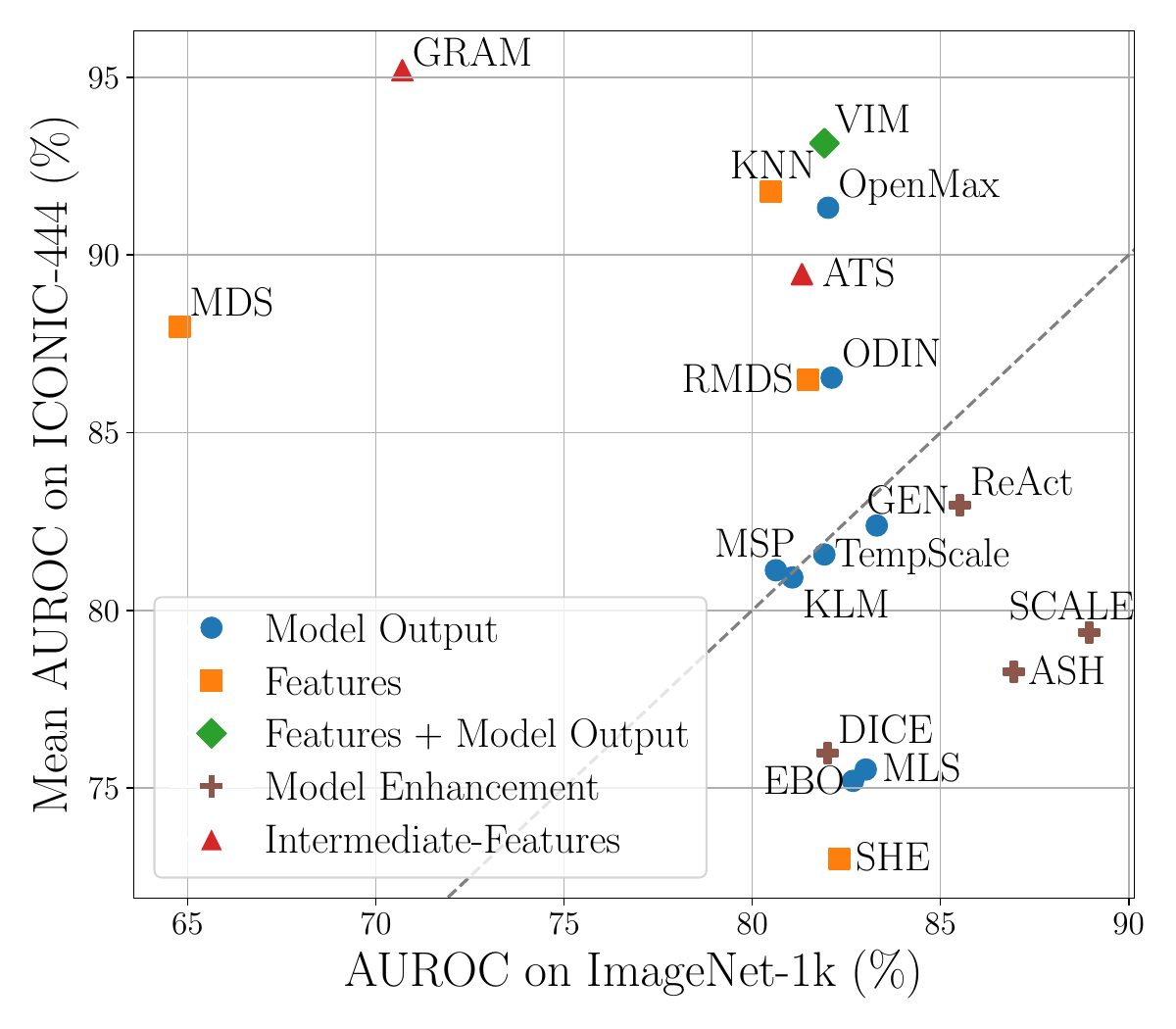}
\caption{
Comparison of the OOD detection methods based on the AUROC across the ICONIC-444 and the ImageNet benchmark.
}
\label{fig:iconic_vs_imagenet_auroc}
\end{figure}

\section{Detailed Results}
\label{sec:supp:detailed_results}
In the following, we (i) summarize per-task trends on the baseline backbones (ResNet-18, CCT; \cref{sec:supp:detailed_results:a}), (ii) assess robustness under stricter operating points (FPR95$\rightarrow$FPR99; \cref{sec:supp:detailed_results:b}), and (iii) analyze feature spaces and contrast ICONIC-444 with ImageNet-1k (\cref{sec:supp:detailed_results:c}).

\subsection{Details on Baseline Architectures}
\label{sec:supp:detailed_results:a}

A detailed breakdown of results on individual tasks is presented in \cref{fig:detailed_results_consider_task_and_arch}, where the mean FPR at $95\%$ and $99\%$ TPR for each method and architecture is shown. 
The general outperformance of ResNet18 over the more recent CCT architecture aligns with findings from other benchmarks~\cite{zhang2024openood}, where this is attributed to many post-hoc OOD methods being overfit to the representational style of traditional CNNs after being implicitly developed and tuned on ResNet backbones.
Methods that leverage information from layers beyond the model output, such as GRAM, ViM, KNN, and MDS, consistently achieve low FPRs across all tasks, with this effect being more pronounced for the ResNet architecture.
However, the substantial performance gap between these methods and model enhancement methods (\eg, ReAct, DICE, ASH), as well as baseline methods (\eg, MSP, MLS, and EBO), progressively narrows from the Almond task to the more complex (in terms of ID classification) Food-grade task.

Interestingly, model enhancement methods begin to outperform methods like MDS, KNN, and ViM on the Food-grade task, which involves a larger number of ID classes. 
This shift suggests that an increasing number of ID classes may favor model enhancement approaches, possibly because methods like ReAct, DICE, and ASH were developed on diverse datasets like ImageNet, which aligns more closely with the Food-grade task’s 324 ID classes. 
GRAM remains a high-performing method across all tasks and retains its advantage even as the number of classes and class granularity increase.

All methods demonstrate task-dependent performance variations, with a relatively consistent degradation from FPR95 to FPR99 within each task. 
OpenMax, however, deviates from this trend, showing a notably larger increase in mean FPR on the Kernels and Food-grade tasks, where its FPR more than doubles.

Furthermore, the results suggest that as the number of ID classes increases, the overall OOD detection performance across all methods improves. 
This observation contrasts with findings by Huang \etal ~\cite{huang2021mos}, who show that the FPR of baseline methods increases as the number of classes grows, particularly on ImageNet where classes are randomly sampled. 
In our dataset, however, the controlled environment, with a uniform background and many fine-grained classes that transition smoothly to more coarse-grained ones, appears to positively affect OOD detection performance. 
Nevertheless, even for state-of-the-art approaches, ICONIC-444 is still far from being saturated, as demonstrated by the high FPR rates.

In summary, our results reveal that all OOD methods are highly sensitive to the complexity and granularity of ID tasks, as well as the architecture used, underscoring the importance of diverse OOD detection benchmarks. 
Such benchmarks are crucial for accurately evaluating each method's strengths and weaknesses, enabling more informed selection for real-world applications, and ultimately enhancing the reliability and safety of OOD detection methods in practical settings.

\begin{figure*}[ht]
    \centering
    \begin{subfigure}[b]{0.49\linewidth}
        \centering
        \includegraphics[width=\linewidth]{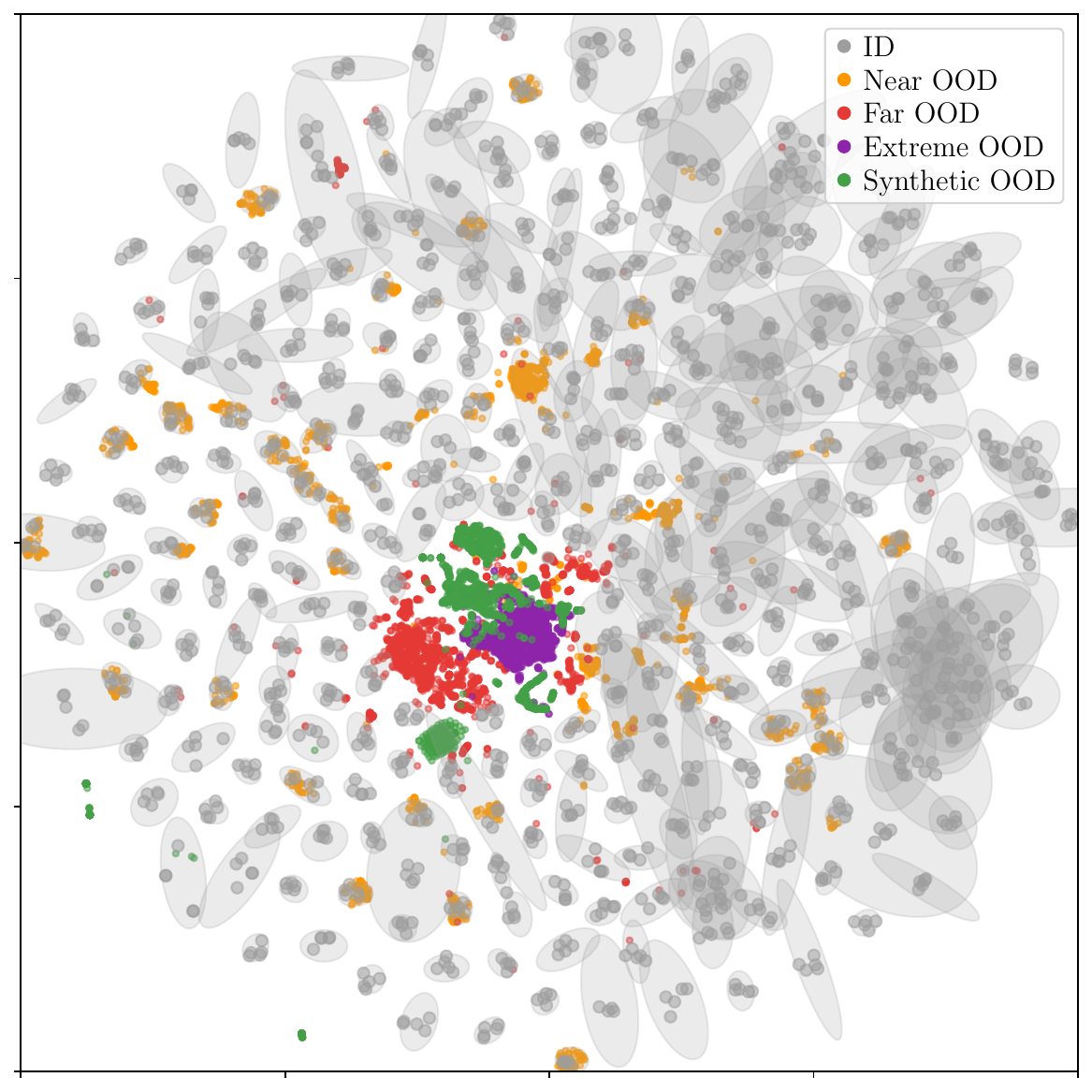}
        \caption{
        ICONIC-444 (Food-grade Task)
        }
    \end{subfigure}
    \hfill
    \begin{subfigure}[b]{0.49\linewidth}
        \centering
        \includegraphics[width=\linewidth]{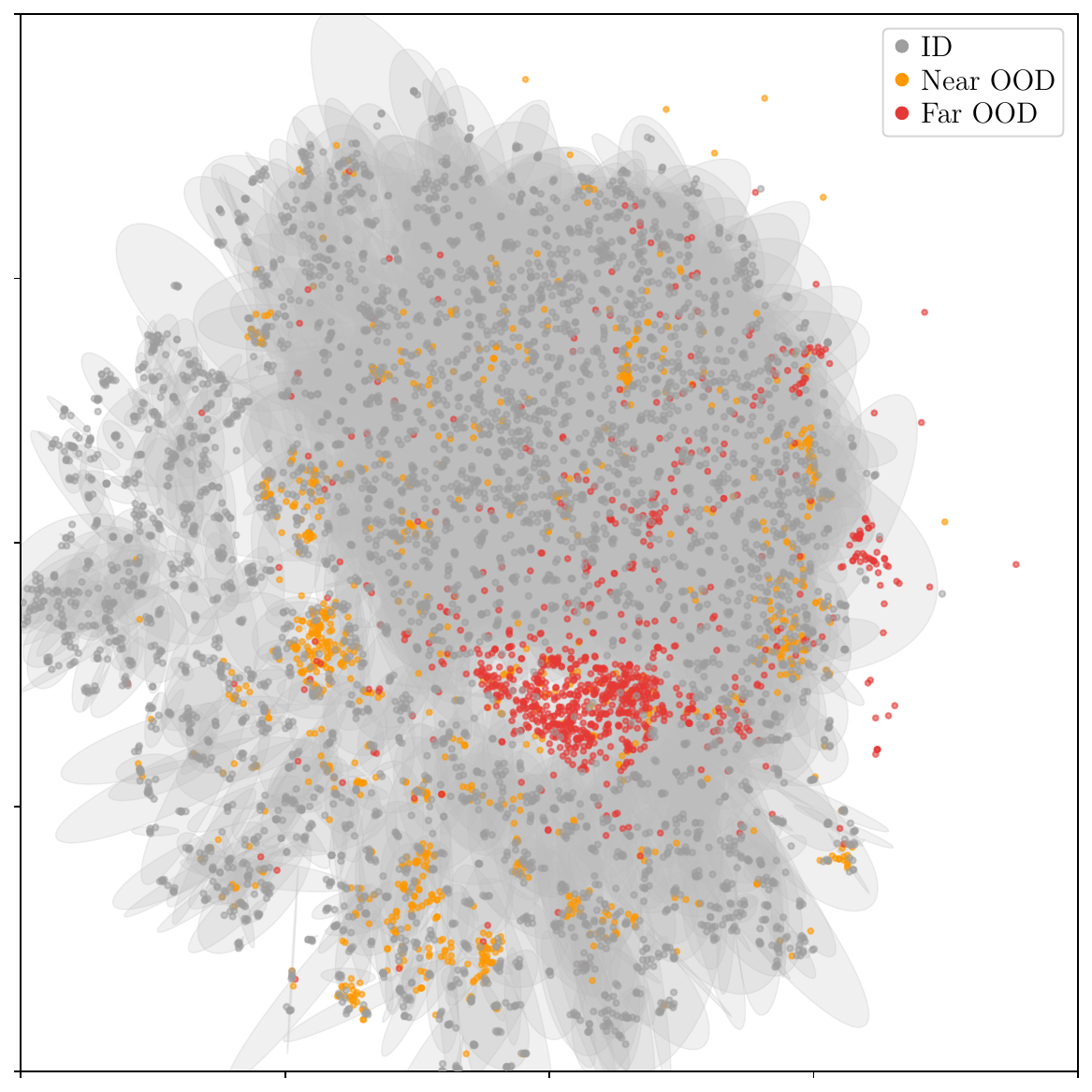}
        \caption{
        ImageNet-1k (OpenOOD ImageNet-1k Benchmark~\cite{zhang2024openood})
        }
    \end{subfigure}
    
    \caption{
    t-SNE visualization of penultimate layer feature embeddings from (a) the ICONIC-444 Food-grade task and (b) the ImageNet-1k benchmark. Features for ICONIC-444 were extracted from a ResNet18 model, while ImageNet-1k features were extracted from a ResNet50. 
    The t-SNE visualization includes only a subset of samples to enhance readability.
    For ID data, 5 samples per class are shown; for OOD data, only a small subset of classes is depicted.
    }
    \label{fig:tsne_iconic_vs_imagenet}
\end{figure*}

\subsection{Robustness under Stricter Evaluation}
\label{sec:supp:detailed_results:b}

\cref{tab:ood_bm_avg_performance_fpr95_fpr99} shows the average OOD detection performance along with the standard deviation for each method, evaluated over three random seeds and two architectures (ResNet18 and CCT) across the four OOD categories.
As expected, performance degrades significantly for all methods across all OOD categories when moving from $95\%$ to $99\%$ TPR. 
This trend is particularly pronounced in the near- and far-OOD categories and remains evident in the extreme- and synthetic-OOD samples.

Among all methods, GRAM and ATS stand out as the only two that effectively handle extreme- and synthetic-OOD samples, at least in the FPR95 setting.
GRAM shows significantly better performance in the FPR99 setting in these categories, likely due to its use of higher-order moments for modeling ID data, whereas ATS relies solely on the mean.
This more comprehensive statistical modeling allows GRAM to capture finer distinctions between ID and OOD samples, making it the more robust approach also for the more strict evaluation with 99\% TPR.

These results emphasize the importance of considering both AUROC and FPR when evaluating OOD detection methods.
For example, MSP achieves an average AUROC ranking, yet at the critical FPR99 operating point, it performs worse than all methods except ASH-b and RankFeat.
This highlights the need for stricter evaluation metrics beyond AUROC alone, especially for real-world applications where minimizing false positives is crucial.

\begin{figure*}[ht]
    \centering
    \begin{subfigure}[b]{0.49\linewidth}
        \centering
        \includegraphics[width=\linewidth]{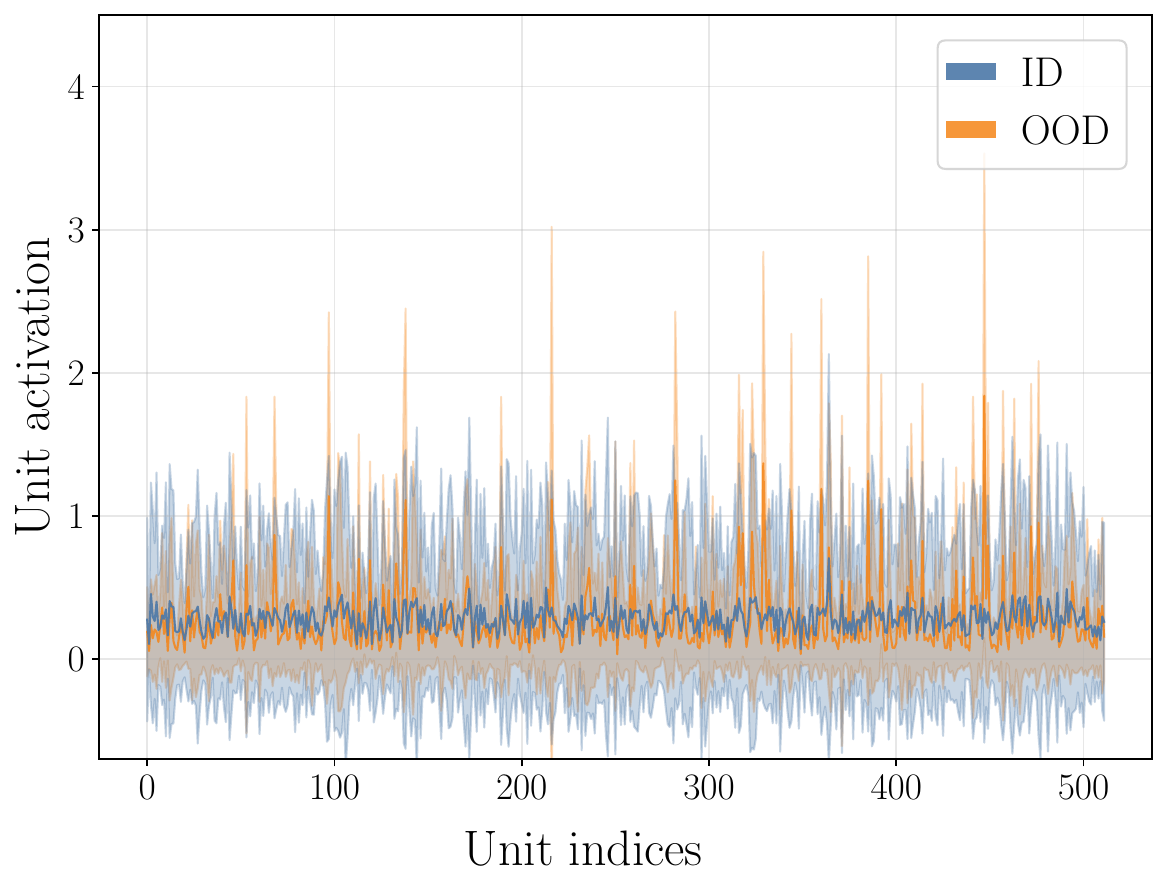}
        \caption{
        ICONIC-444 (Food-grade Task, ResNet18)
        }
    \end{subfigure}
    \hfill
    \begin{subfigure}[b]{0.49\linewidth}
        \centering
        \includegraphics[width=\linewidth]{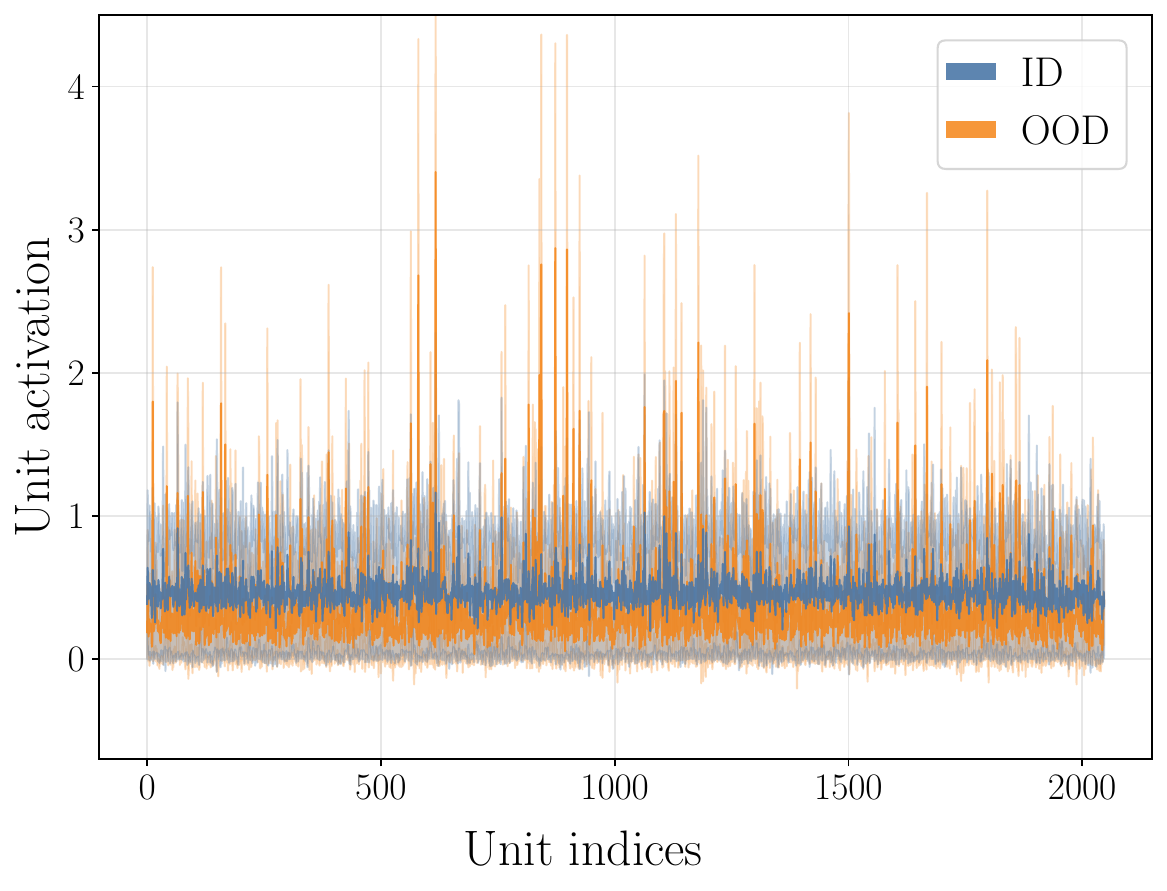}
        \caption{
        ImageNet-1k (OpenOOD ImageNet-1k Benchmark~\cite{zhang2024openood}, ResNet50)
        }
    \end{subfigure}
    
    \caption{
    Per-unit penultimate layer activation profiles for ID and the far-OOD category on a) ICONIC-444 Food-grade with ResNet-18 and (b) ImageNet-1k (OpenOOD benchmark) with ResNet-50.
    The x-axis indexes individual units in the penultimate layer (each position corresponds to one unit), and the y-axis is activation magnitude; the solid line denotes the mean and the shaded region the standard deviation.
    }
    \label{fig:mean_activation_profile_iconic_vs_imagenet}
\end{figure*}

\begin{table*}
\setlength{\extrarowheight}{1.5pt}
\centering\footnotesize
\resizebox{\textwidth}{!}{
\begin{tabular}{llccccccccccc}
\toprule
\multirow{3}{*}{} &
\multirow{3}{*}{\textbf{Method}} &
\multirow{3}{*}{\shortstack[c]{\textbf{Pretrained}\\\textbf{on ImageNet}}} &
\multicolumn{10}{c}{\textbf{OOD-Dataset}} \\
\cmidrule(lr){4-11}
& & &
\multicolumn{2}{c}{\textbf{Near}} &
\multicolumn{2}{c}{\textbf{Far}} &
\multicolumn{2}{c}{\textbf{Extreme}} &
\multicolumn{2}{c}{\textbf{Synthetic}} &
\multicolumn{2}{c}{\textbf{Average}} \\
\cmidrule(lr){4-5}\cmidrule(lr){6-7}\cmidrule(lr){8-9}\cmidrule(lr){10-11}\cmidrule(lr){12-13}
& & &
FPR95 $\downarrow$ & AUROC $\uparrow$ &
FPR95 $\downarrow$ & AUROC $\uparrow$ &
FPR95 $\downarrow$ & AUROC $\uparrow$ &
FPR95 $\downarrow$ & AUROC $\uparrow$ &
FPR95 $\downarrow$ & AUROC $\uparrow$ \\
\midrule
\multirow{6}{*}{{\rotatebox[origin=c]{90}{\textbf{ConvNeXt-P}}}} & GRAM~\cite{sastry20a_gram} & \xmark & $\cellcolor[HTML]{CBDBED} 34.99^{\scriptsize\pm \phantom{0}2.74}$ & $\cellcolor[HTML]{B3CAE4} 91.42^{\scriptsize\pm \phantom{0}1.07}$ & $\cellcolor[HTML]{B1C8E3} 22.44^{\scriptsize\pm \phantom{0}5.58}$ & $\cellcolor[HTML]{9DBBDC} 94.75^{\scriptsize\pm \phantom{0}1.47}$ & $\cellcolor[HTML]{749ECE} \phantom{0}\textbf{0.00}^{\scriptsize\pm \phantom{0}0.00}$ & $\cellcolor[HTML]{749ECE} \textbf{99.99}^{\scriptsize\pm \phantom{0}0.00}$ & $\cellcolor[HTML]{749ECE} \phantom{0}\textbf{0.00}^{\scriptsize\pm \phantom{0}0.00}$ & $\cellcolor[HTML]{749ECE} \textbf{99.95}^{\scriptsize\pm \phantom{0}0.03}$ & $\cellcolor[HTML]{9DBBDC} 14.36^{\scriptsize\pm 17.36}$ & $\cellcolor[HTML]{90B1D8} 96.53^{\scriptsize\pm \phantom{0}4.20}$ \\
 & ATS~\cite{Krumpl2024_ats} & \xmark & $\cellcolor[HTML]{F1F5FA} 61.39^{\scriptsize\pm 10.83}$ & $\cellcolor[HTML]{E3EBF5} 81.29^{\scriptsize\pm \phantom{0}3.98}$ & $\cellcolor[HTML]{C7D8EB} 32.74^{\scriptsize\pm 12.62}$ & $\cellcolor[HTML]{B3CAE4} 91.50^{\scriptsize\pm \phantom{0}3.60}$ & $\cellcolor[HTML]{749ECE} \phantom{0}\underline{0.00}^{\scriptsize\pm \phantom{0}0.00}$ & $\cellcolor[HTML]{759FCE} 99.83^{\scriptsize\pm \phantom{0}0.25}$ & $\cellcolor[HTML]{95B5DA} 11.29^{\scriptsize\pm 10.68}$ & $\cellcolor[HTML]{97B6DA} 95.58^{\scriptsize\pm \phantom{0}4.82}$ & $\cellcolor[HTML]{B9CEE6} 26.35^{\scriptsize\pm 27.02}$ & $\cellcolor[HTML]{AFC7E3} 92.05^{\scriptsize\pm \phantom{0}7.94}$ \\
 & VIM~\cite{wang2022_ood_vim} & \xmark & $\cellcolor[HTML]{FEFFFF} 84.09^{\scriptsize\pm \phantom{0}2.76}$ & $\cellcolor[HTML]{FFFFFF} 62.21^{\scriptsize\pm \phantom{0}2.98}$ & $\cellcolor[HTML]{FFFFFF} 86.74^{\scriptsize\pm \phantom{0}6.26}$ & $\cellcolor[HTML]{FFFFFF} 61.77^{\scriptsize\pm \phantom{0}5.76}$ & $\cellcolor[HTML]{89ADD5} \phantom{0}7.23^{\scriptsize\pm \phantom{0}7.80}$ & $\cellcolor[HTML]{86ABD4} 97.83^{\scriptsize\pm \phantom{0}3.10}$ & $\cellcolor[HTML]{BBD0E7} 27.11^{\scriptsize\pm 13.24}$ & $\cellcolor[HTML]{C6D7EB} 88.08^{\scriptsize\pm \phantom{0}5.08}$ & $\cellcolor[HTML]{E5EDF6} 51.29^{\scriptsize\pm 40.24}$ & $\cellcolor[HTML]{EEF3F9} 77.47^{\scriptsize\pm 18.32}$ \\
 & KNN~\cite{sun2022knnood} & \xmark & $\cellcolor[HTML]{F1F5FA} 61.83^{\scriptsize\pm \phantom{0}1.20}$ & $\cellcolor[HTML]{D9E5F2} 83.77^{\scriptsize\pm \phantom{0}0.89}$ & $\cellcolor[HTML]{F1F5FA} 60.78^{\scriptsize\pm 10.10}$ & $\cellcolor[HTML]{C5D6EA} 88.29^{\scriptsize\pm \phantom{0}2.81}$ & $\cellcolor[HTML]{CADAEC} 34.07^{\scriptsize\pm 10.48}$ & $\cellcolor[HTML]{B1C9E4} 91.76^{\scriptsize\pm \phantom{0}3.73}$ & $\cellcolor[HTML]{D8E4F1} 42.14^{\scriptsize\pm 10.34}$ & $\cellcolor[HTML]{C0D3E9} 89.08^{\scriptsize\pm \phantom{0}4.29}$ & $\cellcolor[HTML]{E3ECF5} 49.71^{\scriptsize\pm 13.80}$ & $\cellcolor[HTML]{C5D6EA} 88.22^{\scriptsize\pm \phantom{0}3.32}$ \\
 & SCALE~\cite{xu2024scaling} & \xmark & $\cellcolor[HTML]{FEFEFF} 81.75^{\scriptsize\pm \phantom{0}2.69}$ & $\cellcolor[HTML]{FCFDFE} 69.35^{\scriptsize\pm \phantom{0}0.19}$ & $\cellcolor[HTML]{FEFFFF} 82.93^{\scriptsize\pm \phantom{0}5.51}$ & $\cellcolor[HTML]{FAFCFD} 70.88^{\scriptsize\pm \phantom{0}2.64}$ & $\cellcolor[HTML]{FEFEFF} 81.53^{\scriptsize\pm 12.97}$ & $\cellcolor[HTML]{FFFFFF} 50.24^{\scriptsize\pm 12.92}$ & $\cellcolor[HTML]{FAFBFD} 71.45^{\scriptsize\pm 18.25}$ & $\cellcolor[HTML]{FFFFFF} 59.97^{\scriptsize\pm 10.62}$ & $\cellcolor[HTML]{FDFEFE} 79.41^{\scriptsize\pm \phantom{0}5.35}$ & $\cellcolor[HTML]{FFFFFF} 62.61^{\scriptsize\pm \phantom{0}9.55}$ \\
 & MSP~\cite{hendrycks2016_ood_msp} & \xmark & $\cellcolor[HTML]{FEFFFF} 83.54^{\scriptsize\pm \phantom{0}0.20}$ & $\cellcolor[HTML]{F7F9FC} 73.14^{\scriptsize\pm \phantom{0}0.48}$ & $\cellcolor[HTML]{FEFFFF} 84.08^{\scriptsize\pm \phantom{0}1.86}$ & $\cellcolor[HTML]{F4F7FB} 74.64^{\scriptsize\pm \phantom{0}2.70}$ & $\cellcolor[HTML]{FFFFFF} 87.91^{\scriptsize\pm \phantom{0}4.77}$ & $\cellcolor[HTML]{FFFFFF} 57.50^{\scriptsize\pm \phantom{0}7.50}$ & $\cellcolor[HTML]{FFFFFF} 85.07^{\scriptsize\pm \phantom{0}4.17}$ & $\cellcolor[HTML]{FFFFFF} 61.05^{\scriptsize\pm \phantom{0}6.95}$ & $\cellcolor[HTML]{FFFFFF} 85.15^{\scriptsize\pm \phantom{0}1.94}$ & $\cellcolor[HTML]{FEFEFF} 66.58^{\scriptsize\pm \phantom{0}8.58}$ \\
\midrule
\multirow{6}{*}{{\rotatebox[origin=c]{90}{\textbf{ConvNeXt-P}}}} & GRAM~\cite{sastry20a_gram} & \cmark & $\cellcolor[HTML]{CEDDEE} 36.73^{\scriptsize\pm \phantom{0}2.06}$ & $\cellcolor[HTML]{A5C0DF} 93.57^{\scriptsize\pm \phantom{0}0.33}$ & $\cellcolor[HTML]{D4E1F0} 40.16^{\scriptsize\pm \phantom{0}2.03}$ & $\cellcolor[HTML]{A2BEDE} 94.08^{\scriptsize\pm \phantom{0}0.35}$ & $\cellcolor[HTML]{749ECE} \phantom{0}\textbf{0.00}^{\scriptsize\pm \phantom{0}0.00}$ & $\cellcolor[HTML]{759ECE} 99.90^{\scriptsize\pm \phantom{0}0.11}$ & $\cellcolor[HTML]{76A0CF} \phantom{0}\underline{0.77}^{\scriptsize\pm \phantom{0}0.28}$ & $\cellcolor[HTML]{77A0CF} 99.57^{\scriptsize\pm \phantom{0}0.12}$ & $\cellcolor[HTML]{A9C3E1} 19.41^{\scriptsize\pm 22.02}$ & $\cellcolor[HTML]{8EB0D7} 96.78^{\scriptsize\pm \phantom{0}3.42}$ \\
 & ATS~\cite{Krumpl2024_ats} & \cmark & $\cellcolor[HTML]{DAE6F2} 43.63^{\scriptsize\pm \phantom{0}5.79}$ & $\cellcolor[HTML]{CBDBED} 86.92^{\scriptsize\pm \phantom{0}3.34}$ & $\cellcolor[HTML]{BBD0E7} 27.30^{\scriptsize\pm \phantom{0}6.01}$ & $\cellcolor[HTML]{A9C3E1} 92.94^{\scriptsize\pm \phantom{0}0.98}$ & $\cellcolor[HTML]{749ECE} \phantom{0}\textbf{0.00}^{\scriptsize\pm \phantom{0}0.00}$ & $\cellcolor[HTML]{749ECE} 99.99^{\scriptsize\pm \phantom{0}0.01}$ & $\cellcolor[HTML]{9BB9DC} 13.67^{\scriptsize\pm \phantom{0}0.75}$ & $\cellcolor[HTML]{95B5DA} 95.83^{\scriptsize\pm \phantom{0}0.70}$ & $\cellcolor[HTML]{AEC6E2} 21.15^{\scriptsize\pm 18.68}$ & $\cellcolor[HTML]{A3BFDF} 93.92^{\scriptsize\pm \phantom{0}5.49}$ \\
 & VIM~\cite{wang2022_ood_vim} & \cmark & $\cellcolor[HTML]{C2D5EA} \underline{30.70}^{\scriptsize\pm \phantom{0}4.19}$ & $\cellcolor[HTML]{A8C2E0} 93.17^{\scriptsize\pm \phantom{0}1.13}$ & $\cellcolor[HTML]{C4D6EA} 31.12^{\scriptsize\pm \phantom{0}4.89}$ & $\cellcolor[HTML]{A1BDDE} 94.20^{\scriptsize\pm \phantom{0}0.90}$ & $\cellcolor[HTML]{759ECE} \phantom{0}0.19^{\scriptsize\pm \phantom{0}0.26}$ & $\cellcolor[HTML]{76A0CF} 99.74^{\scriptsize\pm \phantom{0}0.25}$ & $\cellcolor[HTML]{8BAED6} \phantom{0}7.92^{\scriptsize\pm \phantom{0}0.95}$ & $\cellcolor[HTML]{81A7D2} 98.50^{\scriptsize\pm \phantom{0}0.38}$ & $\cellcolor[HTML]{A5C0DF} 17.48^{\scriptsize\pm 15.82}$ & $\cellcolor[HTML]{91B2D8} 96.40^{\scriptsize\pm \phantom{0}3.21}$ \\
 & KNN~\cite{sun2022knnood} & \cmark & $\cellcolor[HTML]{DFE9F4} 47.05^{\scriptsize\pm \phantom{0}1.08}$ & $\cellcolor[HTML]{B8CDE6} 90.60^{\scriptsize\pm \phantom{0}0.41}$ & $\cellcolor[HTML]{DDE7F3} 45.21^{\scriptsize\pm \phantom{0}2.08}$ & $\cellcolor[HTML]{A9C3E1} 92.93^{\scriptsize\pm \phantom{0}0.36}$ & $\cellcolor[HTML]{AAC4E1} 19.53^{\scriptsize\pm 15.60}$ & $\cellcolor[HTML]{89ADD5} 97.42^{\scriptsize\pm \phantom{0}1.34}$ & $\cellcolor[HTML]{B7CDE6} 25.18^{\scriptsize\pm \phantom{0}3.57}$ & $\cellcolor[HTML]{91B3D8} 96.36^{\scriptsize\pm \phantom{0}0.77}$ & $\cellcolor[HTML]{CADAEC} 34.24^{\scriptsize\pm 13.94}$ & $\cellcolor[HTML]{A0BDDE} 94.33^{\scriptsize\pm \phantom{0}3.14}$ \\
 & SCALE~\cite{xu2024scaling} & \cmark & $\cellcolor[HTML]{EFF4F9} 59.22^{\scriptsize\pm 10.60}$ & $\cellcolor[HTML]{DEE8F3} 82.53^{\scriptsize\pm \phantom{0}7.40}$ & $\cellcolor[HTML]{EBF1F8} 55.46^{\scriptsize\pm \phantom{0}2.20}$ & $\cellcolor[HTML]{D7E3F1} 84.24^{\scriptsize\pm \phantom{0}6.15}$ & $\cellcolor[HTML]{D2DFEF} 38.69^{\scriptsize\pm 19.02}$ & $\cellcolor[HTML]{A3BFDF} 93.87^{\scriptsize\pm \phantom{0}2.79}$ & $\cellcolor[HTML]{C6D7EB} 32.12^{\scriptsize\pm \phantom{0}8.05}$ & $\cellcolor[HTML]{B4CBE5} 91.22^{\scriptsize\pm \phantom{0}3.72}$ & $\cellcolor[HTML]{DEE8F3} 46.37^{\scriptsize\pm 13.03}$ & $\cellcolor[HTML]{C6D7EB} 87.97^{\scriptsize\pm \phantom{0}5.44}$ \\
 & MSP~\cite{hendrycks2016_ood_msp} & \cmark & $\cellcolor[HTML]{EFF4F9} 59.48^{\scriptsize\pm \phantom{0}0.73}$ & $\cellcolor[HTML]{C7D8EB} 87.71^{\scriptsize\pm \phantom{0}1.23}$ & $\cellcolor[HTML]{ECF2F8} 56.42^{\scriptsize\pm \phantom{0}4.48}$ & $\cellcolor[HTML]{BFD2E8} 89.42^{\scriptsize\pm \phantom{0}0.31}$ & $\cellcolor[HTML]{DEE8F3} 46.01^{\scriptsize\pm 15.93}$ & $\cellcolor[HTML]{A3BFDF} 93.92^{\scriptsize\pm \phantom{0}2.14}$ & $\cellcolor[HTML]{CCDCED} 35.64^{\scriptsize\pm \phantom{0}1.86}$ & $\cellcolor[HTML]{A3BFDF} 93.87^{\scriptsize\pm \phantom{0}0.85}$ & $\cellcolor[HTML]{E3EBF5} 49.39^{\scriptsize\pm 10.83}$ & $\cellcolor[HTML]{B4CBE5} 91.23^{\scriptsize\pm \phantom{0}3.15}$ \\
\midrule
\multirow{6}{*}{{\rotatebox[origin=c]{90}{\textbf{ConvNeXt-T}}}} & GRAM~\cite{sastry20a_gram} & \cmark & $\cellcolor[HTML]{CEDDEE} 36.62^{\scriptsize\pm \phantom{0}2.04}$ & $\cellcolor[HTML]{A3BFDF} \underline{93.91}^{\scriptsize\pm \phantom{0}0.51}$ & $\cellcolor[HTML]{D7E3F1} 41.68^{\scriptsize\pm \phantom{0}1.52}$ & $\cellcolor[HTML]{A5C0DF} 93.62^{\scriptsize\pm \phantom{0}0.53}$ & $\cellcolor[HTML]{749ECE} \phantom{0}\textbf{0.00}^{\scriptsize\pm \phantom{0}0.00}$ & $\cellcolor[HTML]{749ECE} 99.93^{\scriptsize\pm \phantom{0}0.02}$ & $\cellcolor[HTML]{84A9D4} \phantom{0}5.30^{\scriptsize\pm \phantom{0}1.57}$ & $\cellcolor[HTML]{7AA2D0} 99.27^{\scriptsize\pm \phantom{0}0.06}$ & $\cellcolor[HTML]{ADC6E2} 20.90^{\scriptsize\pm 21.29}$ & $\cellcolor[HTML]{8FB1D7} 96.68^{\scriptsize\pm \phantom{0}3.38}$ \\
 & ATS~\cite{Krumpl2024_ats} & \cmark & $\cellcolor[HTML]{E0EAF4} 47.68^{\scriptsize\pm 13.12}$ & $\cellcolor[HTML]{CCDBED} 86.85^{\scriptsize\pm \phantom{0}4.72}$ & $\cellcolor[HTML]{C7D8EB} 32.84^{\scriptsize\pm \phantom{0}5.35}$ & $\cellcolor[HTML]{B0C8E3} 91.94^{\scriptsize\pm \phantom{0}0.86}$ & $\cellcolor[HTML]{749ECE} \phantom{0}\textbf{0.00}^{\scriptsize\pm \phantom{0}0.00}$ & $\cellcolor[HTML]{749ECE} 99.96^{\scriptsize\pm \phantom{0}0.03}$ & $\cellcolor[HTML]{95B5DA} 11.62^{\scriptsize\pm \phantom{0}3.55}$ & $\cellcolor[HTML]{89ADD5} 97.43^{\scriptsize\pm \phantom{0}0.69}$ & $\cellcolor[HTML]{B2C9E4} 23.03^{\scriptsize\pm 21.33}$ & $\cellcolor[HTML]{A2BEDE} 94.04^{\scriptsize\pm \phantom{0}5.85}$ \\
 & VIM~\cite{wang2022_ood_vim} & \cmark & $\cellcolor[HTML]{B3CAE4} \textbf{23.40}^{\scriptsize\pm \phantom{0}1.70}$ & $\cellcolor[HTML]{9EBBDD} \textbf{94.65}^{\scriptsize\pm \phantom{0}0.43}$ & $\cellcolor[HTML]{AFC7E3} \underline{21.64}^{\scriptsize\pm \phantom{0}5.32}$ & $\cellcolor[HTML]{95B5DA} \underline{95.89}^{\scriptsize\pm \phantom{0}1.29}$ & $\cellcolor[HTML]{749ECE} \phantom{0}0.00^{\scriptsize\pm \phantom{0}0.01}$ & $\cellcolor[HTML]{749ECE} 99.98^{\scriptsize\pm \phantom{0}0.01}$ & $\cellcolor[HTML]{84A9D4} \phantom{0}5.41^{\scriptsize\pm \phantom{0}0.87}$ & $\cellcolor[HTML]{7BA3D0} 99.12^{\scriptsize\pm \phantom{0}0.17}$ & $\cellcolor[HTML]{98B7DB} \textbf{12.61}^{\scriptsize\pm 11.67}$ & $\cellcolor[HTML]{89ADD5} \textbf{97.41}^{\scriptsize\pm \phantom{0}2.55}$ \\
 & KNN~\cite{sun2022knnood} & \cmark & $\cellcolor[HTML]{D4E1F0} 39.70^{\scriptsize\pm \phantom{0}2.72}$ & $\cellcolor[HTML]{ADC6E2} 92.34^{\scriptsize\pm \phantom{0}0.25}$ & $\cellcolor[HTML]{D3E0EF} 39.07^{\scriptsize\pm \phantom{0}1.80}$ & $\cellcolor[HTML]{A3BFDF} 93.91^{\scriptsize\pm \phantom{0}0.43}$ & $\cellcolor[HTML]{91B2D8} \phantom{0}9.89^{\scriptsize\pm \phantom{0}2.91}$ & $\cellcolor[HTML]{85AAD4} 97.90^{\scriptsize\pm \phantom{0}0.48}$ & $\cellcolor[HTML]{B1C8E3} 22.43^{\scriptsize\pm \phantom{0}6.00}$ & $\cellcolor[HTML]{93B4D9} 96.14^{\scriptsize\pm \phantom{0}1.27}$ & $\cellcolor[HTML]{BCD1E8} 27.77^{\scriptsize\pm 14.36}$ & $\cellcolor[HTML]{9BB9DC} 95.07^{\scriptsize\pm \phantom{0}2.45}$ \\
 & SCALE~\cite{xu2024scaling} & \cmark & $\cellcolor[HTML]{F5F8FC} 65.91^{\scriptsize\pm \phantom{0}7.32}$ & $\cellcolor[HTML]{E4ECF5} 80.84^{\scriptsize\pm \phantom{0}1.95}$ & $\cellcolor[HTML]{F4F7FB} 64.61^{\scriptsize\pm \phantom{0}4.13}$ & $\cellcolor[HTML]{DFE9F4} 82.35^{\scriptsize\pm \phantom{0}1.76}$ & $\cellcolor[HTML]{F6F9FC} 66.50^{\scriptsize\pm 22.32}$ & $\cellcolor[HTML]{E4ECF5} 80.88^{\scriptsize\pm 11.43}$ & $\cellcolor[HTML]{DFE9F4} 46.69^{\scriptsize\pm \phantom{0}5.77}$ & $\cellcolor[HTML]{DCE6F3} 83.25^{\scriptsize\pm \phantom{0}1.27}$ & $\cellcolor[HTML]{F1F5FA} 60.93^{\scriptsize\pm \phantom{0}9.52}$ & $\cellcolor[HTML]{E0EAF4} 81.83^{\scriptsize\pm \phantom{0}1.18}$ \\
 & MSP~\cite{hendrycks2016_ood_msp} & \cmark & $\cellcolor[HTML]{EBF1F8} 55.88^{\scriptsize\pm \phantom{0}5.41}$ & $\cellcolor[HTML]{CADAEC} 87.16^{\scriptsize\pm \phantom{0}2.13}$ & $\cellcolor[HTML]{EBF1F8} 56.09^{\scriptsize\pm \phantom{0}3.50}$ & $\cellcolor[HTML]{C1D4E9} 88.95^{\scriptsize\pm \phantom{0}1.09}$ & $\cellcolor[HTML]{E4ECF5} 50.05^{\scriptsize\pm 16.83}$ & $\cellcolor[HTML]{C5D6EA} 88.29^{\scriptsize\pm \phantom{0}5.49}$ & $\cellcolor[HTML]{CCDBED} 35.46^{\scriptsize\pm \phantom{0}3.82}$ & $\cellcolor[HTML]{C4D6EA} 88.35^{\scriptsize\pm \phantom{0}3.40}$ & $\cellcolor[HTML]{E3EBF5} 49.37^{\scriptsize\pm \phantom{0}9.69}$ & $\cellcolor[HTML]{C5D7EB} 88.19^{\scriptsize\pm \phantom{0}0.75}$ \\
\midrule
\multirow{6}{*}{{\rotatebox[origin=c]{90}{\textbf{ViT-S/16}}}} & GRAM~\cite{sastry20a_gram} & \cmark & $\cellcolor[HTML]{C7D8EB} 33.06^{\scriptsize\pm \phantom{0}4.72}$ & $\cellcolor[HTML]{A6C1E0} 93.41^{\scriptsize\pm \phantom{0}0.72}$ & $\cellcolor[HTML]{AAC4E1} \textbf{19.66}^{\scriptsize\pm \phantom{0}4.00}$ & $\cellcolor[HTML]{92B3D9} \textbf{96.29}^{\scriptsize\pm \phantom{0}0.52}$ & $\cellcolor[HTML]{749ECE} \phantom{0}\textbf{0.00}^{\scriptsize\pm \phantom{0}0.00}$ & $\cellcolor[HTML]{749ECE} \underline{99.99}^{\scriptsize\pm \phantom{0}0.00}$ & $\cellcolor[HTML]{78A1D0} \phantom{0}1.59^{\scriptsize\pm \phantom{0}0.17}$ & $\cellcolor[HTML]{76A0CF} \underline{99.69}^{\scriptsize\pm \phantom{0}0.03}$ & $\cellcolor[HTML]{9BB9DC} \underline{13.57}^{\scriptsize\pm 15.76}$ & $\cellcolor[HTML]{8AADD6} \underline{97.35}^{\scriptsize\pm \phantom{0}3.11}$ \\
 & ATS~\cite{Krumpl2024_ats} & \cmark & $\cellcolor[HTML]{EEF3F9} 57.84^{\scriptsize\pm \phantom{0}3.46}$ & $\cellcolor[HTML]{D3E1F0} 85.15^{\scriptsize\pm \phantom{0}0.78}$ & $\cellcolor[HTML]{C0D3E9} 29.27^{\scriptsize\pm \phantom{0}3.13}$ & $\cellcolor[HTML]{A8C3E0} 93.10^{\scriptsize\pm \phantom{0}0.73}$ & $\cellcolor[HTML]{749ECE} \phantom{0}0.04^{\scriptsize\pm \phantom{0}0.05}$ & $\cellcolor[HTML]{749ECE} 99.98^{\scriptsize\pm \phantom{0}0.01}$ & $\cellcolor[HTML]{96B6DA} 11.85^{\scriptsize\pm \phantom{0}0.34}$ & $\cellcolor[HTML]{8EB0D7} 96.80^{\scriptsize\pm \phantom{0}0.53}$ & $\cellcolor[HTML]{B6CCE5} 24.75^{\scriptsize\pm 25.11}$ & $\cellcolor[HTML]{A4BFDF} 93.76^{\scriptsize\pm \phantom{0}6.39}$ \\
 & VIM~\cite{wang2022_ood_vim} & \cmark & $\cellcolor[HTML]{EDF2F9} 57.27^{\scriptsize\pm \phantom{0}5.09}$ & $\cellcolor[HTML]{D6E2F0} 84.73^{\scriptsize\pm \phantom{0}1.94}$ & $\cellcolor[HTML]{E0EAF4} 47.55^{\scriptsize\pm \phantom{0}5.86}$ & $\cellcolor[HTML]{B7CCE5} 90.82^{\scriptsize\pm \phantom{0}1.84}$ & $\cellcolor[HTML]{80A6D2} \phantom{0}4.04^{\scriptsize\pm \phantom{0}3.12}$ & $\cellcolor[HTML]{7CA3D1} 99.06^{\scriptsize\pm \phantom{0}0.56}$ & $\cellcolor[HTML]{B3CAE4} 23.64^{\scriptsize\pm \phantom{0}4.25}$ & $\cellcolor[HTML]{A1BDDE} 94.24^{\scriptsize\pm \phantom{0}1.17}$ & $\cellcolor[HTML]{C7D8EB} 33.13^{\scriptsize\pm 23.99}$ & $\cellcolor[HTML]{AEC6E2} 92.21^{\scriptsize\pm \phantom{0}6.03}$ \\
 & KNN~\cite{sun2022knnood} & \cmark & $\cellcolor[HTML]{F2F6FA} 62.27^{\scriptsize\pm \phantom{0}3.50}$ & $\cellcolor[HTML]{E5EDF6} 80.60^{\scriptsize\pm \phantom{0}0.37}$ & $\cellcolor[HTML]{EEF3F9} 58.61^{\scriptsize\pm \phantom{0}4.30}$ & $\cellcolor[HTML]{CCDBED} 86.89^{\scriptsize\pm \phantom{0}0.48}$ & $\cellcolor[HTML]{D4E1F0} 40.20^{\scriptsize\pm \phantom{0}8.41}$ & $\cellcolor[HTML]{B0C8E3} 91.88^{\scriptsize\pm \phantom{0}2.38}$ & $\cellcolor[HTML]{EEF3F9} 57.80^{\scriptsize\pm \phantom{0}2.39}$ & $\cellcolor[HTML]{DFE9F4} 82.37^{\scriptsize\pm \phantom{0}3.33}$ & $\cellcolor[HTML]{EAF0F8} 54.72^{\scriptsize\pm \phantom{0}9.87}$ & $\cellcolor[HTML]{D2E0EF} 85.43^{\scriptsize\pm \phantom{0}5.05}$ \\
 & SCALE~\cite{xu2024scaling} & \cmark & $\cellcolor[HTML]{EBF1F8} 55.76^{\scriptsize\pm 10.88}$ & $\cellcolor[HTML]{D6E2F0} 84.62^{\scriptsize\pm \phantom{0}5.43}$ & $\cellcolor[HTML]{E8EFF7} 53.08^{\scriptsize\pm 11.27}$ & $\cellcolor[HTML]{C7D8EB} 87.89^{\scriptsize\pm \phantom{0}4.46}$ & $\cellcolor[HTML]{A1BEDE} 16.10^{\scriptsize\pm \phantom{0}8.48}$ & $\cellcolor[HTML]{8BAED6} 97.18^{\scriptsize\pm \phantom{0}1.59}$ & $\cellcolor[HTML]{D6E2F0} 40.82^{\scriptsize\pm 10.20}$ & $\cellcolor[HTML]{C2D4E9} 88.79^{\scriptsize\pm \phantom{0}3.98}$ & $\cellcolor[HTML]{D7E3F1} 41.44^{\scriptsize\pm 18.10}$ & $\cellcolor[HTML]{BED1E8} 89.62^{\scriptsize\pm \phantom{0}5.35}$ \\
 & MSP~\cite{hendrycks2016_ood_msp} & \cmark & $\cellcolor[HTML]{FAFBFD} 71.50^{\scriptsize\pm \phantom{0}3.57}$ & $\cellcolor[HTML]{DAE6F2} 83.40^{\scriptsize\pm \phantom{0}1.05}$ & $\cellcolor[HTML]{F9FBFD} 70.68^{\scriptsize\pm \phantom{0}3.53}$ & $\cellcolor[HTML]{D0DEEE} 85.94^{\scriptsize\pm \phantom{0}0.78}$ & $\cellcolor[HTML]{E7EEF7} 52.30^{\scriptsize\pm \phantom{0}7.75}$ & $\cellcolor[HTML]{AAC4E1} 92.83^{\scriptsize\pm \phantom{0}1.55}$ & $\cellcolor[HTML]{F9FBFD} 70.68^{\scriptsize\pm \phantom{0}3.16}$ & $\cellcolor[HTML]{D8E4F1} 84.05^{\scriptsize\pm \phantom{0}1.51}$ & $\cellcolor[HTML]{F6F9FC} 66.29^{\scriptsize\pm \phantom{0}9.33}$ & $\cellcolor[HTML]{CDDCEE} 86.56^{\scriptsize\pm \phantom{0}4.32}$ \\
\midrule
\multirow{6}{*}{{\rotatebox[origin=c]{90}{\textbf{ViT-B/16}}}} & GRAM~\cite{sastry20a_gram} & \cmark & $\cellcolor[HTML]{CBDBED} 35.01^{\scriptsize\pm \phantom{0}2.57}$ & $\cellcolor[HTML]{ABC4E1} 92.77^{\scriptsize\pm \phantom{0}0.72}$ & $\cellcolor[HTML]{B5CBE5} 24.29^{\scriptsize\pm \phantom{0}3.68}$ & $\cellcolor[HTML]{97B6DA} 95.58^{\scriptsize\pm \phantom{0}0.54}$ & $\cellcolor[HTML]{749ECE} \phantom{0}\textbf{0.00}^{\scriptsize\pm \phantom{0}0.00}$ & $\cellcolor[HTML]{749ECE} 99.97^{\scriptsize\pm \phantom{0}0.02}$ & $\cellcolor[HTML]{7CA3D1} \phantom{0}2.63^{\scriptsize\pm \phantom{0}0.37}$ & $\cellcolor[HTML]{78A1CF} 99.54^{\scriptsize\pm \phantom{0}0.07}$ & $\cellcolor[HTML]{A0BCDD} 15.48^{\scriptsize\pm 16.97}$ & $\cellcolor[HTML]{8DAFD7} 96.97^{\scriptsize\pm \phantom{0}3.42}$ \\
 & ATS~\cite{Krumpl2024_ats} & \cmark & $\cellcolor[HTML]{F7FAFC} 68.38^{\scriptsize\pm \phantom{0}4.66}$ & $\cellcolor[HTML]{DBE6F2} 83.35^{\scriptsize\pm \phantom{0}0.61}$ & $\cellcolor[HTML]{E3EBF5} 49.41^{\scriptsize\pm \phantom{0}8.85}$ & $\cellcolor[HTML]{C6D7EB} 88.02^{\scriptsize\pm \phantom{0}2.03}$ & $\cellcolor[HTML]{77A0CF} \phantom{0}1.03^{\scriptsize\pm \phantom{0}1.07}$ & $\cellcolor[HTML]{759FCE} 99.82^{\scriptsize\pm \phantom{0}0.18}$ & $\cellcolor[HTML]{A3BFDF} 16.86^{\scriptsize\pm \phantom{0}0.61}$ & $\cellcolor[HTML]{A6C1E0} 93.39^{\scriptsize\pm \phantom{0}0.20}$ & $\cellcolor[HTML]{C9D9EC} 33.92^{\scriptsize\pm 30.55}$ & $\cellcolor[HTML]{B5CBE5} 91.14^{\scriptsize\pm \phantom{0}7.09}$ \\
 & VIM~\cite{wang2022_ood_vim} & \cmark & $\cellcolor[HTML]{E8EFF7} 52.99^{\scriptsize\pm \phantom{0}8.99}$ & $\cellcolor[HTML]{CCDBED} 86.82^{\scriptsize\pm \phantom{0}1.84}$ & $\cellcolor[HTML]{DCE7F3} 44.87^{\scriptsize\pm 15.37}$ & $\cellcolor[HTML]{B1C9E4} 91.73^{\scriptsize\pm \phantom{0}1.92}$ & $\cellcolor[HTML]{749ECE} \phantom{0}0.00^{\scriptsize\pm \phantom{0}0.01}$ & $\cellcolor[HTML]{759ECE} 99.91^{\scriptsize\pm \phantom{0}0.06}$ & $\cellcolor[HTML]{A1BDDE} 15.84^{\scriptsize\pm \phantom{0}1.40}$ & $\cellcolor[HTML]{97B6DA} 95.61^{\scriptsize\pm \phantom{0}0.72}$ & $\cellcolor[HTML]{BED2E8} 28.43^{\scriptsize\pm 24.77}$ & $\cellcolor[HTML]{A6C1DF} 93.52^{\scriptsize\pm \phantom{0}5.57}$ \\
 & KNN~\cite{sun2022knnood} & \cmark & $\cellcolor[HTML]{E3ECF5} 49.44^{\scriptsize\pm \phantom{0}1.49}$ & $\cellcolor[HTML]{C2D5EA} 88.71^{\scriptsize\pm \phantom{0}0.16}$ & $\cellcolor[HTML]{E6EEF6} 51.65^{\scriptsize\pm \phantom{0}6.21}$ & $\cellcolor[HTML]{B9CEE6} 90.31^{\scriptsize\pm \phantom{0}0.46}$ & $\cellcolor[HTML]{E6EEF6} 51.56^{\scriptsize\pm 10.62}$ & $\cellcolor[HTML]{B0C8E3} 91.90^{\scriptsize\pm \phantom{0}2.34}$ & $\cellcolor[HTML]{D1DFEF} 37.92^{\scriptsize\pm \phantom{0}1.55}$ & $\cellcolor[HTML]{B9CEE6} 90.34^{\scriptsize\pm \phantom{0}1.05}$ & $\cellcolor[HTML]{E0EAF4} 47.64^{\scriptsize\pm \phantom{0}6.56}$ & $\cellcolor[HTML]{B9CEE6} 90.31^{\scriptsize\pm \phantom{0}1.30}$ \\
 & SCALE~\cite{xu2024scaling} & \cmark & $\cellcolor[HTML]{EEF3F9} 57.86^{\scriptsize\pm \phantom{0}9.06}$ & $\cellcolor[HTML]{CFDEEE} 86.10^{\scriptsize\pm \phantom{0}1.99}$ & $\cellcolor[HTML]{E9EFF7} 53.71^{\scriptsize\pm 11.75}$ & $\cellcolor[HTML]{C4D6EA} 88.33^{\scriptsize\pm \phantom{0}2.79}$ & $\cellcolor[HTML]{FDFDFE} 78.39^{\scriptsize\pm \phantom{0}9.34}$ & $\cellcolor[HTML]{E4ECF5} 80.83^{\scriptsize\pm \phantom{0}4.50}$ & $\cellcolor[HTML]{F8FAFD} 70.33^{\scriptsize\pm \phantom{0}8.80}$ & $\cellcolor[HTML]{F8FAFD} 71.90^{\scriptsize\pm \phantom{0}4.12}$ & $\cellcolor[HTML]{F5F8FB} 65.07^{\scriptsize\pm 11.34}$ & $\cellcolor[HTML]{E0EAF4} 81.79^{\scriptsize\pm \phantom{0}7.31}$ \\
 & MSP~\cite{hendrycks2016_ood_msp} & \cmark & $\cellcolor[HTML]{FEFEFF} 80.41^{\scriptsize\pm \phantom{0}0.46}$ & $\cellcolor[HTML]{EAF0F8} 78.87^{\scriptsize\pm \phantom{0}0.43}$ & $\cellcolor[HTML]{FEFEFF} 80.77^{\scriptsize\pm \phantom{0}0.30}$ & $\cellcolor[HTML]{E6EEF6} 80.02^{\scriptsize\pm \phantom{0}0.19}$ & $\cellcolor[HTML]{FFFFFF} 85.91^{\scriptsize\pm \phantom{0}2.34}$ & $\cellcolor[HTML]{EFF4F9} 76.90^{\scriptsize\pm \phantom{0}2.89}$ & $\cellcolor[HTML]{FFFFFF} 86.03^{\scriptsize\pm \phantom{0}1.66}$ & $\cellcolor[HTML]{FCFDFE} 69.02^{\scriptsize\pm \phantom{0}3.13}$ & $\cellcolor[HTML]{FEFFFF} 83.28^{\scriptsize\pm \phantom{0}3.11}$ & $\cellcolor[HTML]{F1F5FA} 76.20^{\scriptsize\pm \phantom{0}4.96}$ \\

\bottomrule
\end{tabular}
}
\caption{
Average OOD detection performance for the Almond task, evaluated across four architectures (ConvNeXt-P, ConvNeXt-T, ViT-S/16, and ViT-B/16) and reported per OOD category (near, far, extreme, and synthetic).
Results are reported as mean $\pm$ standard deviation over three random seeds.
Arrows ($\uparrow$/$\downarrow$) indicate whether higher or lower values are better.
Cells are color-coded from \textcolor{best_blue}{\textbf{blue}} (high performance) to white (low performance). 
Additionally, the \textbf{best} and \underline{second-best} results in each column are highlighted in bold and underlined, respectively. 
All values are reported as percentages.
}
\label{tab:modern_arch}
\end{table*}

\subsection{Need for Complementary Benchmarks}
\label{sec:supp:detailed_results:c}

\cref{fig:iconic_vs_imagenet_auroc} compares the OOD detection performance (mean AUROC) of evaluated methods on ICONIC-444 against their reported performance on the ImageNet-1k benchmark from OpenOOD~\cite{zhang2024openood}. 
The comparison reveals that no single class of methods is universally superior; instead, the optimal approach is highly dependent on the dataset's intrinsic properties. 
Notably, feature-based methods (\eg, GRAM, ViM, KNN, ATS) significantly outperform other approaches on ICONIC-444, while model enhancement techniques (\eg, ASH, SCALE, ReAct) are the top performers on the ImageNet benchmark.

We hypothesize that this divergence is caused by fundamental structural differences between the datasets' feature spaces, an idea supported by both visual and quantitative analysis. 
Unless stated otherwise, all feature space analyses use penultimate layer activations: ResNet18 for ICONIC-444 (Food-grade task) and ResNet50 for ImageNet-1k.
The t-SNE visualizations in \cref{fig:tsne_iconic_vs_imagenet} provide a clear visual intuition, showing that ICONIC-444's features form more compact and distinct clusters compared to the diffuse distribution typical of ImageNet.
This observation is confirmed quantitatively. 
ICONIC-444's controlled acquisition process (object-centric) yields a lower mean variance of pixel intensity ($0.154$ vs.~$0.226$ for ImageNet), a consistent background, and finer-grained class distinctions compared to ImageNet's broader semantic and textural variability (scene-centric). 
At the feature level, we observe substantially higher activation sparsity, which is the fraction of near-zero feature values ($0.081$ vs.~$0.005$ for ImageNet); and stronger class separation (inter/intra $\approx 3.67$ vs.\ $1.31$ on ImageNet).
These indicators are consistent with a more statistically coherent representation where distance/statistics-based detectors like GRAM and KNN are highly effective at modeling feature distributions.

However, this controlled environment does not imply a simpler task; it creates a different one. 
The challenge shifts from handling nuisance variables common in \emph{in-the-wild} data---such as cluttered backgrounds---to the more focused task of detecting subtle semantic differences between visually similar classes.
Our results, both in the main paper and this supplementary material, show that \textbf{even top methods yield enormously high false positive rates} on our near- and far-OOD categories, proving the benchmark's difficulty. 

As further evidence, the penultimate layer activation profiles in \cref{fig:mean_activation_profile_iconic_vs_imagenet} show that the ReAct~\cite{sun2021_ood_react} signature observed on ImageNet---near-constant ID means with lower, more variable, positively skewed OOD activations---attenuates on ICONIC-444, where the ID and OOD profiles largely overlap across units. 
Consequently, activation-shaping heuristics (\eg, ReAct) offer limited gains on ICONIC-444, whereas statistic/distance-based detectors (GRAM, ViM, KNN) remain effective by exploiting the structure of intermediate features.

This analysis empirically demonstrates ICONIC-444's value as a diagnostic tool that proves \textbf{the optimal OOD detection strategy is highly task dependent.}
It complements large-scale OOD benchmarks like ImageNet by providing a challenging benchmark for fine-grained OOD, which is essential for progress in safety- and sustainability-critical applications like industrial sorting and recycling.

\section{Evaluation on Complex Architectures}
\label{sec:supp:eval_modern_arch}

In this section, we analyze the benchmark results on more complex architectures, specifically ConvNeXt and Vision Transformers (ViTs). 
For OOD detection, we evaluate six post-hoc methods: the three top-performing methods from our main benchmark (GRAM, ATS, and ViM, selected based on average FPR99), KNN (which utilizes penultimate layer features), SCALE (one of the most recent model-enhancement methods), and MSP (the initial OOD detection baseline). 
\cref{tab:modern_arch} reports the average OOD detection performance for each method over three random seeds on the Almond task, broken down by near, far, extreme, and synthetic OOD categories.

Despite higher capacity and potentially richer feature representations (see model details in ~\cref{tab:model_overview}), ConvNeXt and ViT do not consistently outperform ResNet in OOD detection, aligning with observations from Zhang \etal~\cite{zhang2024openood}.
A possible explanation is that most OOD detection methods have been primarily designed and tuned for ResNet-based backbones, suggesting they may be implicitly optimized for the representational style of traditional CNNs.

\begin{table*}
\setlength{\extrarowheight}{1.5pt}
\centering\footnotesize
\resizebox{\textwidth}{!}{
\begin{tabular}{lcccccccccccc}
\toprule
\multirow{3}{*}{\textbf{Method}} & 
\multirow{3}{*}{\shortstack[c]{\textbf{ID Image} \\ \textbf{Availability}}} &
\multirow{3}{*}{\shortstack[c]{\textbf{Training} \\ \textbf{Type}}} &
\multicolumn{8}{c}{\textbf{OOD-Dataset}}   \\
\cmidrule(lr){4-13}
& & & \multicolumn{2}{c}{\textbf{Near}} & 
      \multicolumn{2}{c}{\textbf{Far}} & 
      \multicolumn{2}{c}{\textbf{Extreme}} & 
      \multicolumn{2}{c}{\textbf{Synthetic}} &
      \multicolumn{2}{c}{\textbf{Average}}\\
\cmidrule(lr){4-5} \cmidrule(lr){6-7} \cmidrule(lr){8-9} \cmidrule(lr){10-11} \cmidrule(lr){12-13}
& & & FPR95 $\downarrow$ & AUROC $\uparrow$ & 
      FPR95 $\downarrow$ & AUROC $\uparrow$ & 
      FPR95 $\downarrow$ & AUROC $\uparrow$ & 
      FPR95 $\downarrow$ & AUROC $\uparrow$ & 
      FPR95 $\downarrow$ & AUROC $\uparrow$  \\
\midrule
MCM~\cite{ming2022mcm} & Zero-shot & Training-free          & \underline{96.10} & \underline{43.72} & \underline{98.65} & \underline{37.53} & \underline{99.01} & \underline{22.94} & 99.41 & \phantom{0}9.60 & \underline{98.29} & 27.70\\
GL-MCM~\cite{Miyai2023GLMCMGA} & Zero-shot & Training-free  & 96.31             & 43.44             & 98.91             & 35.96 & 99.77 & 18.39 & \underline{99.14} & \underline{13.82} & 98.53 & \underline{27.90}  \\
GalLop~\cite{lafon2024gallop} & Few-shot & ID Training  & \textbf{89.91} & \textbf{73.67} & \textbf{90.65} & \textbf{72.77} & \textbf{97.88} & \textbf{68.79} & \textbf{66.36} & \textbf{87.25} & \textbf{86.20} & \textbf{75.62} \\
\bottomrule
\end{tabular}
}
\caption{
CLIP-based (CLIP-B/16) OOD detection performance on the Almond task, measured in FPR95 and AUROC.  
Arrows ($\uparrow$/$\downarrow$) indicate whether higher or lower values are better.
All values are reported as percentages.
}
\label{tab:clip}
\end{table*}

Focusing on ConvNeXt-P, we compare models trained from scratch versus those fine-tuned from ImageNet-pretrained weights. 
Pretraining generally improves extreme-OOD detection, likely because the semantics learned from ImageNet (\eg, generic object shapes or textures) transfer well to OOD examples that share partial visual characteristics with those seen during pretraining.
Also, this improvement is partially explained by dataset overlap: our extreme-OOD category includes ImageNet images for consistency across benchmarks, meaning that pretrained models have already been exposed to similar data.
Although we strictly use the ImageNet test set for OOD evaluation, the semantic structures of these samples remain familiar to the model, potentially inflating performance.
This underscores the importance of either training models from scratch or ensuring that test-time OOD samples are entirely absent from pretraining data, as this is crucial for strict and robust OOD evaluation.
Additionally, this benefit is not uniform across all methods; for instance, GRAM exhibits a notable performance drop on the FPR95 metric when applied to pre-trained models.

In summary, \textbf{scaling up model capacity or adopting newer architectures does not automatically yield better OOD detection.} 
Most importantly, our results confirm that ICONIC-444 remains a challenging benchmark \textbf{even for larger and more complex architectures}, underscoring its value in advancing OOD detection research; nevertheless, FPR remains prohibitively high, limiting real-world applicability.

\section{VLM-based OOD Detection Methods}
\label{sec:supp:eval_clip}

To link ICONIC-444 to recent developments in Image Classification and OOD research, we evaluate how foundation models perform on our benchmark.
Foundation models are large, pre-trained models that can be adapted to a wide range of downstream tasks, often outperforming more specialized supervised approaches \cite{radford2021_clip}. 
In particular, vision language models (VLMs) such as CLIP \cite{radford2021_clip} have demonstrated impressive zero-shot classification capabilities, inspiring several CLIP-based OOD detection methods \cite{miyai2024ood_survey_vlm}.

\begin{figure}
\centering
\includegraphics[width=\linewidth]{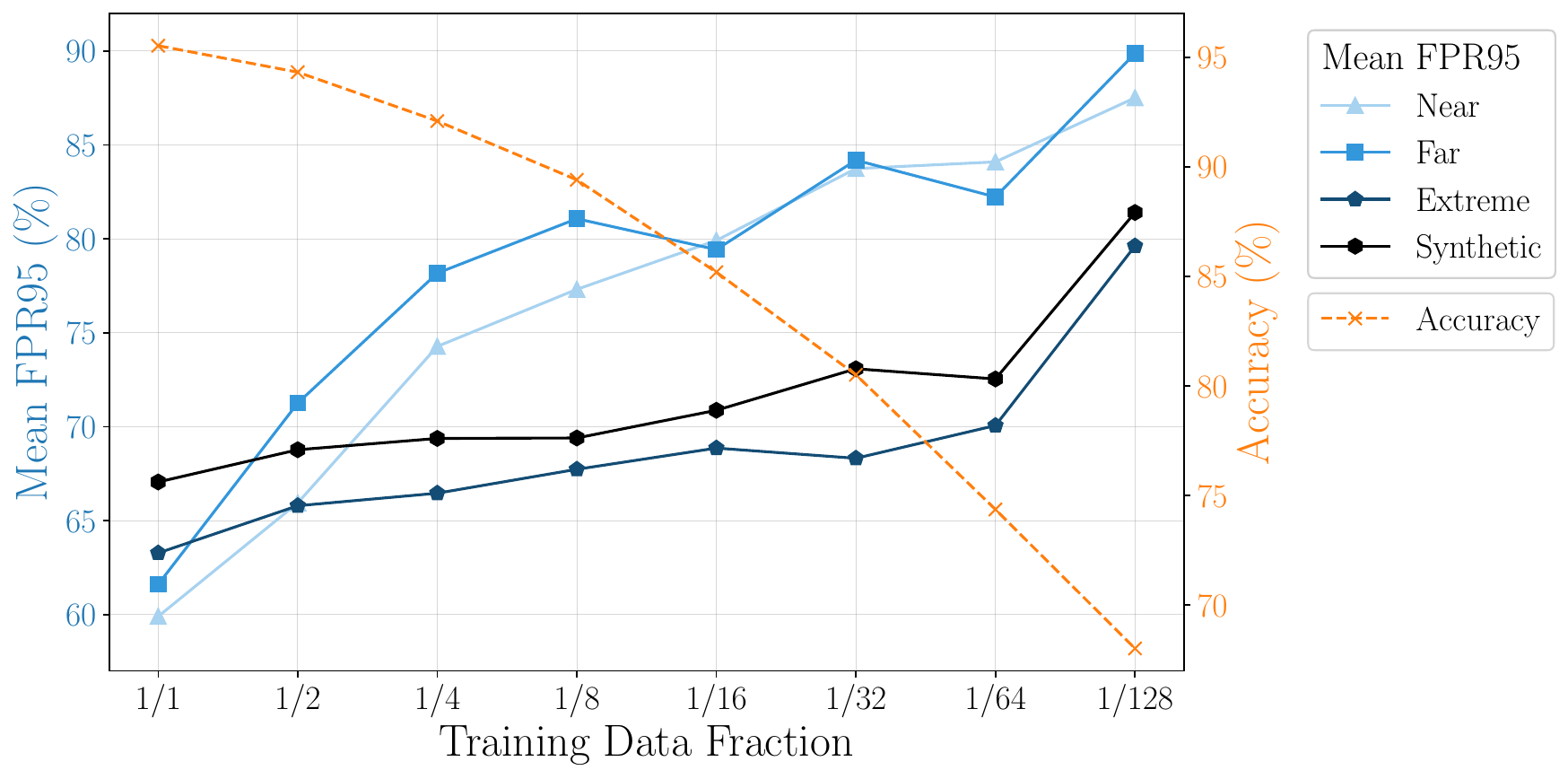}
\caption{
Mean FPR95 for the ResNet18 model across four OOD categories and accuracy for the Almond task, averaged over three seeds and all OOD detection methods, with progressively reduced training set size while keeping validation and test sets fixed.
}
\label{fig:data_reduction_mean_fpr}
\end{figure}

\begin{figure*}
\centering
\includegraphics[width=\linewidth]{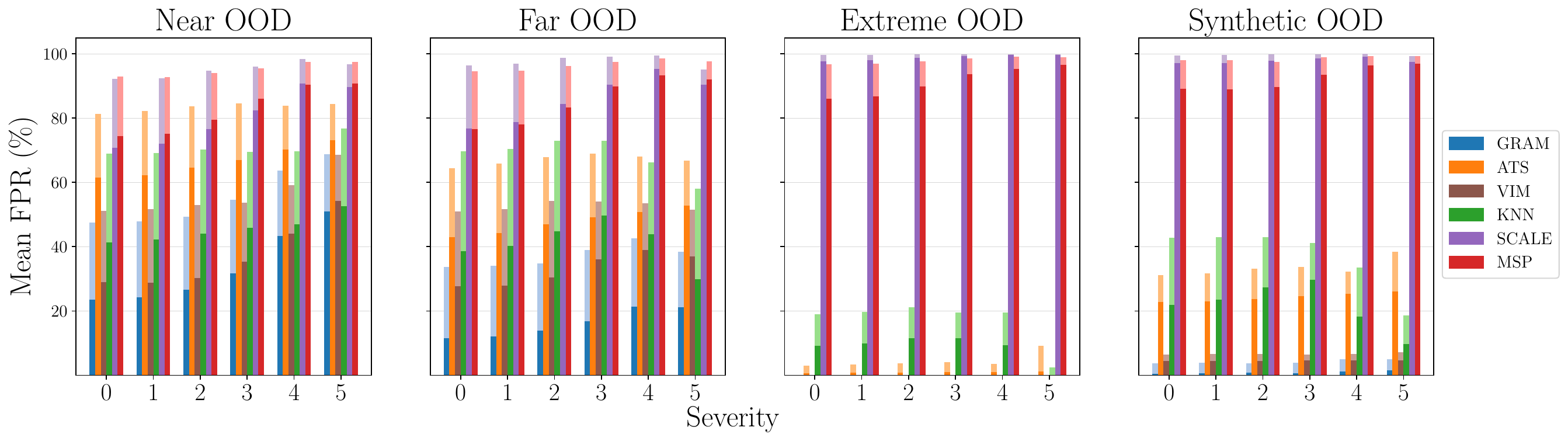}
\caption{
Average OOD detection performance on the Almond task using ResNet18 over three random seeds, measured as FPR at 95\% (FPR95) and 99\% (FPR99) TPR. 
Results are reported for each method, OOD category (near, far, extreme, and synthetic), and Gaussian noise corruption with five different severity levels. 
The dark-shaded portion of each bar represents the mean FPR95, while the light-shaded extension indicates the additional increase up to FPR99.
}
\label{fig:almond_gaussian_noise_rn18}
\end{figure*}

We evaluate three CLIP-based OOD detection methods on the Almond task, selected for its simplicity in terms of in-distribution and near-OOD granularity. 
We consider MCM \cite{ming2022mcm} and GL-MCM \cite{Miyai2023GLMCMGA} (both zero-shot and training-free), as well as GalLop \cite{lafon2024gallop}, which employs a few-shot approach using 16 ID training images.
The standard parametrization of zero-shot CLIP is insufficient, achieving only 20.34\% ID accuracy.
Thus, we fine-tuned the text prompts and class descriptions (on the ID validation set) and could increase the accuracy to 34.70\%.
Nonetheless, even with few-shot fine-tuning, GalLop achieves only 78.45\% accuracy---noticeably lower than our ResNet baselines’ 95.53\%.

\cref{tab:clip} shows each method’s OOD detection performance across near-, far-, extreme-, and synthetic-OOD categories.
As expected, these CLIP-based approaches generally underperform specialized OOD detection methods.
As demonstrated by Radford \etal~\cite{radford2021_clip} (on the MNIST dataset), CLIP does not directly address the fundamental deep learning’s generalization challenge; instead, it aims to cover a broad diversity of data such that most real-world samples become effectively ID.
However, ICONIC-444 contains high-resolution, industrial-domain images that are rarely---if at all---present in publicly available web collections, placing them firmly OOD for CLIP. 

Our findings show that, within our benchmark, CLIP underperforms in both ID classification and OOD detection.
Moreover, most existing CLIP-based OOD detection methods are evaluated on benchmarks where ID performance is comparable to supervised approaches and where only coarse-grained OOD samples are analyzed \cite{miyai2024ood_survey_vlm}. 
Therefore, our results further underscore the value of ICONIC-444 as a challenging benchmark for VLMs in both classification and OOD detection contexts.

More broadly, our results suggest that CLIP and similar foundation models cannot serve as a universal solution for OOD detection, as they inevitably fail on specialized datasets outside their training distribution.
This aligns with the limitations identified by Radford \etal~\cite{radford2021_clip}, particularly in the context of zero-shot classification. 
ICONIC-444 demonstrates such domain-specific data, highlighting the limitations of current foundation models in handling fine-grained, high-resolution, and industrial-domain OOD detection tasks, where their pre-training coverage is insufficient.

\section{ID-OOD Performance Correlation Ablation}
\label{sec:supp:io_ood_cor}
\cref{fig:data_reduction_mean_fpr} shows how progressively reducing training data impacts ID accuracy and OOD detection performance (FPR95) across our four OOD categories. 
As the training data is reduced, the accuracy steadily declines, and a corresponding decrease in OOD detection performance is observed. 
This trend is more pronounced for the near- and far-OOD samples, which show a sharper degradation, likely due to their higher semantic correlation with the ID data. 
In contrast, the extreme- and synthetic-OOD categories, which are less semantically related, exhibit a smaller drop in performance, requiring a more drastic reduction in training data to show a significant decline. 

This finding aligns with recent discussions in our field: 
Vaze \etal~\cite{vaze2022openset} show a positive correlation between classifier accuracy and OOD detection performance. 
Recent work~\cite{Humblot-Renaux_2024_CVPR} further analyzes this correlation, showing that OOD detection performance relies largely on the method’s ability to separate misclassified ID samples from true OOD samples.
Our results extend this understanding by highlighting that the complexity and semantic similarity of OOD samples play a significant role in this correlation.

\section{Ablation on Corruptions}
\label{sec:supp:data_corruption}

A key objective in the acquisition of ICONIC-444 was to ensure high-resolution, high-quality images, providing a strong benchmark for OOD detection research.
While the dataset maintains a high level of visual clarity, minor imperfections naturally occur due to the free-fall acquisition setup. 
Variations in object speed and positioning during scanning introduce slight illumination inconsistencies, motion blur, or occasional dust accumulation, reflecting real-world environmental challenges.
Importantly, the intrinsically high quality of ICONIC-444 allows for controlled degradation, enabling systematic robustness testing for OOD detection models.

\cref{fig:almond_gaussian_noise_rn18} shows a detailed analysis of OOD detection performance on the Almond task using ResNet18 under inference-time image corruption. 
Specifically, we introduce Gaussian noise with five severity levels $\sigma=\{0.01,0.02,0.03,0.05,0.1\}$, following the corruption methodology proposed by Hendrycks \etal~\cite{hendrycks2018benchmarking}.
The clean, uncorrupted images correspond to severity level zero.

As expected, increased corruption generally degrades OOD detection performance. 
However, an exception is observed for KNN on far-, extreme-, and synthetic-OOD categories, where at severity levels four and five, OOD detection performance improves. 
This suggests that noise may introduce additional cues that help distinguish OOD samples in the case of KNN, similar to the effect exploited by ODIN~\cite{shiyu17}, where input perturbations enhance OOD separability.

\begin{landscape}
\begin{table}[p]
\centering\footnotesize
\setlength{\extrarowheight}{1.5pt}
\resizebox{\linewidth}{!}{
\begin{tabular}{lccccccccccccccc}
\toprule
\multirow{3}{*}{\textbf{Method}} & \multicolumn{15}{c}{\textbf{OOD-Dataset}} \\
\cmidrule(lr){2-16}
 & \multicolumn{3}{c}{\textbf{Near}} &
   \multicolumn{3}{c}{\textbf{Far}} &
   \multicolumn{3}{c}{\textbf{Extreme}} &
   \multicolumn{3}{c}{\textbf{Synthetic}} &
   \multicolumn{3}{c}{\textbf{Average}} \\
\cmidrule(lr){2-4}  \cmidrule(lr){5-7} \cmidrule(lr){8-10} \cmidrule(lr){11-13} \cmidrule(lr){14-16}
 & FPR95 $\downarrow$ & FPR99 $\downarrow$ & AUROC $\uparrow$
 & FPR95 $\downarrow$ & FPR99 $\downarrow$ & AUROC $\uparrow$
 & FPR95 $\downarrow$ & FPR99 $\downarrow$ & AUROC $\uparrow$
 & FPR95 $\downarrow$ & FPR99 $\downarrow$ & AUROC $\uparrow$
 & FPR95 $\downarrow$ & FPR99 $\downarrow$ & AUROC $\uparrow$ \\
\midrule
GRAM~\cite{sastry20a_gram} & $\cellcolor[HTML]{CCDCED} \textbf{35.74}^{\scriptsize\pm 23.66}$ & $\cellcolor[HTML]{EAF0F8} \textbf{54.59}^{\scriptsize\pm 24.59}$ & $\cellcolor[HTML]{C5D7EB} \textbf{88.13}^{\scriptsize\pm \phantom{0}8.84}$ & $\cellcolor[HTML]{ABC4E1} \textbf{19.75}^{\scriptsize\pm 16.47}$ & $\cellcolor[HTML]{CEDDEE} \textbf{36.89}^{\scriptsize\pm 21.11}$ & $\cellcolor[HTML]{A2BEDE} \textbf{94.03}^{\scriptsize\pm \phantom{0}5.42}$ & $\cellcolor[HTML]{749ECE} \phantom{0}\textbf{0.00}^{\scriptsize\pm \phantom{0}0.00}$ & $\cellcolor[HTML]{749ECE} \phantom{0}\textbf{0.00}^{\scriptsize\pm \phantom{0}0.00}$ & $\cellcolor[HTML]{749ECE} \textbf{99.99}^{\scriptsize\pm \phantom{0}0.01}$ & $\cellcolor[HTML]{759FCE} \phantom{0}\textbf{0.44}^{\scriptsize\pm \phantom{0}0.61}$ & $\cellcolor[HTML]{7BA3D0} \phantom{0}\textbf{2.40}^{\scriptsize\pm \phantom{0}2.43}$ & $\cellcolor[HTML]{759ECE} \textbf{99.88}^{\scriptsize\pm \phantom{0}0.12}$ & $\cellcolor[HTML]{9CBADC} \textbf{13.98}^{\scriptsize\pm 17.18}$ & $\cellcolor[HTML]{B3CAE4} \textbf{23.47}^{\scriptsize\pm 26.73}$ & $\cellcolor[HTML]{97B7DA} \textbf{95.51}^{\scriptsize\pm \phantom{0}5.65}$ \\
ATS~\cite{Krumpl2024_ats} & $\cellcolor[HTML]{F6F9FC} 66.27^{\scriptsize\pm 26.96}$ & $\cellcolor[HTML]{FEFEFF} 80.76^{\scriptsize\pm 19.56}$ & $\cellcolor[HTML]{F0F5FA} 76.35^{\scriptsize\pm 13.17}$ & $\cellcolor[HTML]{E2EBF5} 48.90^{\scriptsize\pm 21.86}$ & $\cellcolor[HTML]{F7F9FC} 67.99^{\scriptsize\pm 17.83}$ & $\cellcolor[HTML]{D0DEEE} 86.04^{\scriptsize\pm 10.09}$ & $\cellcolor[HTML]{76A0CF} \phantom{0}\underline{0.81}^{\scriptsize\pm \phantom{0}2.76}$ & $\cellcolor[HTML]{7BA3D0} \phantom{0}\underline{2.25}^{\scriptsize\pm \phantom{0}6.07}$ & $\cellcolor[HTML]{759FCE} \underline{99.84}^{\scriptsize\pm \phantom{0}0.38}$ & $\cellcolor[HTML]{99B8DB} \underline{13.07}^{\scriptsize\pm \phantom{0}8.09}$ & $\cellcolor[HTML]{ABC4E1} \underline{20.02}^{\scriptsize\pm 10.52}$ & $\cellcolor[HTML]{97B6DA} 95.59^{\scriptsize\pm \phantom{0}3.51}$ & $\cellcolor[HTML]{C6D7EB} 32.26^{\scriptsize\pm 30.50}$ & $\cellcolor[HTML]{D9E4F2} \underline{42.76}^{\scriptsize\pm 37.59}$ & $\cellcolor[HTML]{BED2E8} 89.45^{\scriptsize\pm 10.47}$ \\
VIM~\cite{wang2022_ood_vim} & $\cellcolor[HTML]{DDE7F3} 45.37^{\scriptsize\pm 28.34}$ & $\cellcolor[HTML]{F6F9FC} 66.79^{\scriptsize\pm 29.74}$ & $\cellcolor[HTML]{D2DFEF} 85.53^{\scriptsize\pm 11.91}$ & $\cellcolor[HTML]{C6D7EB} 32.42^{\scriptsize\pm 21.74}$ & $\cellcolor[HTML]{ECF2F8} 56.64^{\scriptsize\pm 28.32}$ & $\cellcolor[HTML]{A5C0DF} \underline{93.57}^{\scriptsize\pm \phantom{0}5.45}$ & $\cellcolor[HTML]{A1BDDE} 15.92^{\scriptsize\pm 24.39}$ & $\cellcolor[HTML]{C4D6EA} 31.44^{\scriptsize\pm 35.47}$ & $\cellcolor[HTML]{8BAED6} 97.12^{\scriptsize\pm \phantom{0}4.48}$ & $\cellcolor[HTML]{9CBADC} 14.07^{\scriptsize\pm 17.73}$ & $\cellcolor[HTML]{B4CBE4} 23.87^{\scriptsize\pm 24.75}$ & $\cellcolor[HTML]{91B3D8} \underline{96.35}^{\scriptsize\pm \phantom{0}5.52}$ & $\cellcolor[HTML]{BBCFE7} \underline{26.94}^{\scriptsize\pm 14.79}$ & $\cellcolor[HTML]{DCE7F3} 44.69^{\scriptsize\pm 20.33}$ & $\cellcolor[HTML]{A8C2E0} \underline{93.14}^{\scriptsize\pm \phantom{0}5.30}$ \\
RMDS~\cite{Ren2021_rmds} & $\cellcolor[HTML]{DDE7F3} 45.27^{\scriptsize\pm 24.47}$ & $\cellcolor[HTML]{F4F7FB} \underline{64.29}^{\scriptsize\pm 18.18}$ & $\cellcolor[HTML]{DDE7F3} 82.96^{\scriptsize\pm 15.25}$ & $\cellcolor[HTML]{D8E4F1} 42.53^{\scriptsize\pm 28.08}$ & $\cellcolor[HTML]{F0F5FA} 60.64^{\scriptsize\pm 24.79}$ & $\cellcolor[HTML]{D6E2F0} 84.61^{\scriptsize\pm 15.03}$ & $\cellcolor[HTML]{B6CCE5} 24.66^{\scriptsize\pm 35.25}$ & $\cellcolor[HTML]{C7D8EB} 33.01^{\scriptsize\pm 36.45}$ & $\cellcolor[HTML]{BED1E8} 89.60^{\scriptsize\pm 17.48}$ & $\cellcolor[HTML]{BBCFE7} 27.07^{\scriptsize\pm 31.30}$ & $\cellcolor[HTML]{CEDDEE} 36.65^{\scriptsize\pm 32.67}$ & $\cellcolor[HTML]{C2D5EA} 88.71^{\scriptsize\pm 17.54}$ & $\cellcolor[HTML]{CBDBED} 34.88^{\scriptsize\pm 10.52}$ & $\cellcolor[HTML]{E2EAF5} 48.65^{\scriptsize\pm 16.09}$ & $\cellcolor[HTML]{CDDCEE} 86.47^{\scriptsize\pm \phantom{0}3.19}$ \\
MDS~\cite{lee2018_mahala} & $\cellcolor[HTML]{E1EAF4} 48.23^{\scriptsize\pm 28.66}$ & $\cellcolor[HTML]{F7FAFC} 68.30^{\scriptsize\pm 27.44}$ & $\cellcolor[HTML]{DFE9F4} 82.33^{\scriptsize\pm 16.47}$ & $\cellcolor[HTML]{CEDDEE} 36.92^{\scriptsize\pm 31.46}$ & $\cellcolor[HTML]{EEF3F9} 58.31^{\scriptsize\pm 29.59}$ & $\cellcolor[HTML]{C7D8EB} 87.75^{\scriptsize\pm 16.03}$ & $\cellcolor[HTML]{ACC5E2} 20.35^{\scriptsize\pm 31.57}$ & $\cellcolor[HTML]{CCDCED} 35.52^{\scriptsize\pm 37.63}$ & $\cellcolor[HTML]{AEC7E3} 92.19^{\scriptsize\pm 16.18}$ & $\cellcolor[HTML]{B4CBE4} 23.74^{\scriptsize\pm 33.01}$ & $\cellcolor[HTML]{C6D7EB} 32.54^{\scriptsize\pm 35.37}$ & $\cellcolor[HTML]{BDD1E8} 89.65^{\scriptsize\pm 17.93}$ & $\cellcolor[HTML]{C6D7EB} 32.31^{\scriptsize\pm 12.80}$ & $\cellcolor[HTML]{E2EBF5} 48.67^{\scriptsize\pm 17.43}$ & $\cellcolor[HTML]{C6D7EB} 87.98^{\scriptsize\pm \phantom{0}4.18}$ \\
KNN~\cite{sun2022knnood} & $\cellcolor[HTML]{DAE6F2} \underline{43.75}^{\scriptsize\pm 18.84}$ & $\cellcolor[HTML]{F5F8FC} 65.54^{\scriptsize\pm 18.16}$ & $\cellcolor[HTML]{D7E3F1} 84.32^{\scriptsize\pm 11.10}$ & $\cellcolor[HTML]{C8D9EC} 33.64^{\scriptsize\pm 17.93}$ & $\cellcolor[HTML]{EEF3F9} 57.85^{\scriptsize\pm 21.02}$ & $\cellcolor[HTML]{ACC5E2} 92.47^{\scriptsize\pm \phantom{0}4.01}$ & $\cellcolor[HTML]{A8C3E0} 18.93^{\scriptsize\pm 20.73}$ & $\cellcolor[HTML]{D2E0EF} 38.97^{\scriptsize\pm 30.67}$ & $\cellcolor[HTML]{91B3D8} 96.32^{\scriptsize\pm \phantom{0}4.96}$ & $\cellcolor[HTML]{ACC5E2} 20.26^{\scriptsize\pm 17.05}$ & $\cellcolor[HTML]{CBDBED} 35.14^{\scriptsize\pm 22.69}$ & $\cellcolor[HTML]{A2BEDE} 93.96^{\scriptsize\pm \phantom{0}6.82}$ & $\cellcolor[HTML]{BFD2E9} 29.14^{\scriptsize\pm 11.79}$ & $\cellcolor[HTML]{E3EBF5} 49.37^{\scriptsize\pm 14.65}$ & $\cellcolor[HTML]{B1C9E4} 91.77^{\scriptsize\pm \phantom{0}5.21}$ \\
ODIN~\cite{shiyu17} & $\cellcolor[HTML]{E0EAF4} 47.70^{\scriptsize\pm 10.58}$ & $\cellcolor[HTML]{F6F9FC} 66.19^{\scriptsize\pm 13.16}$ & $\cellcolor[HTML]{D9E4F2} 83.84^{\scriptsize\pm \phantom{0}3.85}$ & $\cellcolor[HTML]{C4D6EA} \underline{31.44}^{\scriptsize\pm 17.51}$ & $\cellcolor[HTML]{E6EEF6} \underline{51.61}^{\scriptsize\pm 22.26}$ & $\cellcolor[HTML]{B5CBE5} 91.14^{\scriptsize\pm \phantom{0}5.25}$ & $\cellcolor[HTML]{BBD0E7} 27.29^{\scriptsize\pm 29.87}$ & $\cellcolor[HTML]{D5E2F0} 40.63^{\scriptsize\pm 32.22}$ & $\cellcolor[HTML]{CCDCED} 86.73^{\scriptsize\pm 21.94}$ & $\cellcolor[HTML]{BFD2E9} 29.01^{\scriptsize\pm 26.07}$ & $\cellcolor[HTML]{D3E1F0} 39.51^{\scriptsize\pm 29.16}$ & $\cellcolor[HTML]{D7E3F1} 84.45^{\scriptsize\pm 18.82}$ & $\cellcolor[HTML]{C9D9EC} 33.86^{\scriptsize\pm \phantom{0}9.38}$ & $\cellcolor[HTML]{E3ECF5} 49.49^{\scriptsize\pm 12.40}$ & $\cellcolor[HTML]{CDDCEE} 86.54^{\scriptsize\pm \phantom{0}3.31}$ \\
GEN~\cite{Liu2023_GEN} & $\cellcolor[HTML]{EDF2F9} 57.27^{\scriptsize\pm 14.18}$ & $\cellcolor[HTML]{FDFDFE} 77.53^{\scriptsize\pm 12.38}$ & $\cellcolor[HTML]{E1EAF4} 81.66^{\scriptsize\pm \phantom{0}6.05}$ & $\cellcolor[HTML]{DEE8F3} 46.29^{\scriptsize\pm 27.56}$ & $\cellcolor[HTML]{F7F9FC} 68.15^{\scriptsize\pm 25.93}$ & $\cellcolor[HTML]{C9D9EC} 87.36^{\scriptsize\pm \phantom{0}9.29}$ & $\cellcolor[HTML]{E1EAF4} 48.21^{\scriptsize\pm 36.55}$ & $\cellcolor[HTML]{F3F7FB} 63.07^{\scriptsize\pm 37.99}$ & $\cellcolor[HTML]{E2EAF5} 81.57^{\scriptsize\pm 22.06}$ & $\cellcolor[HTML]{E2EBF5} 48.92^{\scriptsize\pm 30.05}$ & $\cellcolor[HTML]{F4F7FB} 64.82^{\scriptsize\pm 27.21}$ & $\cellcolor[HTML]{EAF0F8} 78.94^{\scriptsize\pm 20.11}$ & $\cellcolor[HTML]{E4ECF5} 50.17^{\scriptsize\pm \phantom{0}4.86}$ & $\cellcolor[HTML]{F7FAFC} 68.39^{\scriptsize\pm \phantom{0}6.44}$ & $\cellcolor[HTML]{DFE9F4} 82.38^{\scriptsize\pm \phantom{0}3.55}$ \\
DICE~\cite{sun2022dice} & $\cellcolor[HTML]{F0F4FA} 60.00^{\scriptsize\pm 16.80}$ & $\cellcolor[HTML]{FDFDFE} 77.82^{\scriptsize\pm 13.28}$ & $\cellcolor[HTML]{ECF2F8} 77.87^{\scriptsize\pm \phantom{0}9.36}$ & $\cellcolor[HTML]{E0EAF4} 47.64^{\scriptsize\pm 27.05}$ & $\cellcolor[HTML]{F8FAFD} 69.66^{\scriptsize\pm 24.90}$ & $\cellcolor[HTML]{D3E1F0} 85.13^{\scriptsize\pm 10.76}$ & $\cellcolor[HTML]{E6EEF6} 51.73^{\scriptsize\pm 41.07}$ & $\cellcolor[HTML]{F5F8FB} 65.41^{\scriptsize\pm 37.60}$ & $\cellcolor[HTML]{FBFCFD} 70.42^{\scriptsize\pm 32.97}$ & $\cellcolor[HTML]{E2EBF5} 48.95^{\scriptsize\pm 32.60}$ & $\cellcolor[HTML]{F2F6FA} 62.30^{\scriptsize\pm 27.89}$ & $\cellcolor[HTML]{FBFCFD} 70.41^{\scriptsize\pm 28.31}$ & $\cellcolor[HTML]{E6EEF6} 52.08^{\scriptsize\pm \phantom{0}5.55}$ & $\cellcolor[HTML]{F7FAFC} 68.80^{\scriptsize\pm \phantom{0}6.73}$ & $\cellcolor[HTML]{F1F5FA} 75.96^{\scriptsize\pm \phantom{0}7.05}$ \\
ReAct~\cite{sun2021_ood_react} & $\cellcolor[HTML]{F0F5FA} 60.46^{\scriptsize\pm 16.93}$ & $\cellcolor[HTML]{FDFEFE} 80.07^{\scriptsize\pm 14.68}$ & $\cellcolor[HTML]{E3ECF5} 81.04^{\scriptsize\pm \phantom{0}6.51}$ & $\cellcolor[HTML]{E0EAF4} 47.69^{\scriptsize\pm 25.93}$ & $\cellcolor[HTML]{FAFBFD} 71.59^{\scriptsize\pm 26.45}$ & $\cellcolor[HTML]{C9D9EC} 87.41^{\scriptsize\pm \phantom{0}8.52}$ & $\cellcolor[HTML]{E2EBF5} 48.98^{\scriptsize\pm 39.01}$ & $\cellcolor[HTML]{F4F7FB} 64.46^{\scriptsize\pm 38.05}$ & $\cellcolor[HTML]{E0E9F4} 82.04^{\scriptsize\pm 20.73}$ & $\cellcolor[HTML]{E3ECF5} 49.52^{\scriptsize\pm 31.56}$ & $\cellcolor[HTML]{F5F8FC} 66.02^{\scriptsize\pm 29.53}$ & $\cellcolor[HTML]{E3EBF5} 81.27^{\scriptsize\pm 19.19}$ & $\cellcolor[HTML]{E6EEF6} 51.66^{\scriptsize\pm \phantom{0}5.92}$ & $\cellcolor[HTML]{F9FBFD} 70.54^{\scriptsize\pm \phantom{0}7.05}$ & $\cellcolor[HTML]{DDE7F3} 82.94^{\scriptsize\pm \phantom{0}3.01}$ \\
EBO~\cite{liu2020_ood_ebo} & $\cellcolor[HTML]{F0F4FA} 59.77^{\scriptsize\pm 14.43}$ & $\cellcolor[HTML]{FDFDFE} 78.72^{\scriptsize\pm 11.67}$ & $\cellcolor[HTML]{EEF3F9} 77.59^{\scriptsize\pm \phantom{0}7.27}$ & $\cellcolor[HTML]{E2EAF5} 48.57^{\scriptsize\pm 27.13}$ & $\cellcolor[HTML]{F8FAFD} 69.76^{\scriptsize\pm 24.72}$ & $\cellcolor[HTML]{D6E2F0} 84.65^{\scriptsize\pm \phantom{0}9.99}$ & $\cellcolor[HTML]{EBF1F8} 55.54^{\scriptsize\pm 40.76}$ & $\cellcolor[HTML]{F6F9FC} 66.70^{\scriptsize\pm 37.37}$ & $\cellcolor[HTML]{FCFDFE} 69.26^{\scriptsize\pm 30.90}$ & $\cellcolor[HTML]{E9EFF7} 53.80^{\scriptsize\pm 30.44}$ & $\cellcolor[HTML]{F7FAFC} 68.90^{\scriptsize\pm 25.29}$ & $\cellcolor[HTML]{FCFDFE} 69.28^{\scriptsize\pm 26.14}$ & $\cellcolor[HTML]{EAF0F8} 54.42^{\scriptsize\pm \phantom{0}4.63}$ & $\cellcolor[HTML]{F9FBFD} 71.02^{\scriptsize\pm \phantom{0}5.29}$ & $\cellcolor[HTML]{F3F7FB} 75.19^{\scriptsize\pm \phantom{0}7.43}$ \\
SHE~\cite{zhang2023_she} & $\cellcolor[HTML]{F0F5FA} 60.57^{\scriptsize\pm 16.33}$ & $\cellcolor[HTML]{FDFEFE} 80.02^{\scriptsize\pm 11.15}$ & $\cellcolor[HTML]{F6F9FC} 73.77^{\scriptsize\pm 13.33}$ & $\cellcolor[HTML]{E6EEF6} 51.46^{\scriptsize\pm 27.72}$ & $\cellcolor[HTML]{FAFCFD} 72.99^{\scriptsize\pm 22.69}$ & $\cellcolor[HTML]{E3EBF5} 81.28^{\scriptsize\pm 17.64}$ & $\cellcolor[HTML]{E4ECF5} 49.99^{\scriptsize\pm 39.29}$ & $\cellcolor[HTML]{F7F9FC} 67.52^{\scriptsize\pm 34.44}$ & $\cellcolor[HTML]{FCFDFE} 69.09^{\scriptsize\pm 36.60}$ & $\cellcolor[HTML]{E3ECF5} 49.78^{\scriptsize\pm 32.98}$ & $\cellcolor[HTML]{F4F7FB} 63.97^{\scriptsize\pm 27.75}$ & $\cellcolor[HTML]{FDFDFE} 67.88^{\scriptsize\pm 31.08}$ & $\cellcolor[HTML]{E8EFF7} 52.95^{\scriptsize\pm \phantom{0}5.13}$ & $\cellcolor[HTML]{F9FBFD} 71.13^{\scriptsize\pm \phantom{0}7.00}$ & $\cellcolor[HTML]{F7F9FC} 73.00^{\scriptsize\pm \phantom{0}6.07}$ \\
OpenMax~\cite{Bendale2016_openmax} & $\cellcolor[HTML]{DBE6F2} 44.06^{\scriptsize\pm 11.86}$ & $\cellcolor[HTML]{FDFEFE} 80.23^{\scriptsize\pm 14.83}$ & $\cellcolor[HTML]{CDDCEE} \underline{86.50}^{\scriptsize\pm \phantom{0}4.15}$ & $\cellcolor[HTML]{CEDDEE} 36.38^{\scriptsize\pm 22.15}$ & $\cellcolor[HTML]{FEFEFF} 80.28^{\scriptsize\pm 13.43}$ & $\cellcolor[HTML]{B9CEE6} 90.45^{\scriptsize\pm \phantom{0}5.56}$ & $\cellcolor[HTML]{ABC4E1} 20.10^{\scriptsize\pm 30.63}$ & $\cellcolor[HTML]{F0F5FA} 60.63^{\scriptsize\pm 40.17}$ & $\cellcolor[HTML]{9AB9DB} 95.18^{\scriptsize\pm \phantom{0}6.35}$ & $\cellcolor[HTML]{BBD0E7} 27.29^{\scriptsize\pm 23.36}$ & $\cellcolor[HTML]{F4F7FB} 64.33^{\scriptsize\pm 31.81}$ & $\cellcolor[HTML]{A8C2E0} 93.17^{\scriptsize\pm \phantom{0}5.32}$ & $\cellcolor[HTML]{C5D7EB} 31.96^{\scriptsize\pm 10.46}$ & $\cellcolor[HTML]{FAFBFD} 71.37^{\scriptsize\pm 10.37}$ & $\cellcolor[HTML]{B4CBE4} 91.33^{\scriptsize\pm \phantom{0}3.76}$ \\
SCALE~\cite{xu2024scaling} & $\cellcolor[HTML]{F1F5FA} 60.90^{\scriptsize\pm 16.80}$ & $\cellcolor[HTML]{FEFEFF} 80.74^{\scriptsize\pm 14.60}$ & $\cellcolor[HTML]{E8EFF7} 79.46^{\scriptsize\pm \phantom{0}5.92}$ & $\cellcolor[HTML]{DFE9F4} 46.63^{\scriptsize\pm 28.00}$ & $\cellcolor[HTML]{F8FAFD} 70.20^{\scriptsize\pm 28.34}$ & $\cellcolor[HTML]{CCDCED} 86.74^{\scriptsize\pm \phantom{0}9.87}$ & $\cellcolor[HTML]{E7EEF7} 52.66^{\scriptsize\pm 39.18}$ & $\cellcolor[HTML]{F5F8FB} 65.37^{\scriptsize\pm 38.95}$ & $\cellcolor[HTML]{EFF4F9} 76.83^{\scriptsize\pm 25.80}$ & $\cellcolor[HTML]{E8EFF7} 52.99^{\scriptsize\pm 31.89}$ & $\cellcolor[HTML]{F8FAFD} 69.43^{\scriptsize\pm 27.39}$ & $\cellcolor[HTML]{F4F7FB} 74.47^{\scriptsize\pm 22.94}$ & $\cellcolor[HTML]{E8EFF7} 53.29^{\scriptsize\pm \phantom{0}5.85}$ & $\cellcolor[HTML]{FAFBFD} 71.44^{\scriptsize\pm \phantom{0}6.55}$ & $\cellcolor[HTML]{E9EFF7} 79.38^{\scriptsize\pm \phantom{0}5.32}$ \\
MLS~\cite{hendrycks2019_ood_mls} & $\cellcolor[HTML]{EFF4F9} 59.07^{\scriptsize\pm 14.49}$ & $\cellcolor[HTML]{FDFEFE} 79.12^{\scriptsize\pm 11.78}$ & $\cellcolor[HTML]{EBF1F8} 78.32^{\scriptsize\pm \phantom{0}7.10}$ & $\cellcolor[HTML]{E2EAF5} 48.56^{\scriptsize\pm 26.94}$ & $\cellcolor[HTML]{F9FBFD} 71.14^{\scriptsize\pm 24.65}$ & $\cellcolor[HTML]{D3E1F0} 85.14^{\scriptsize\pm \phantom{0}9.66}$ & $\cellcolor[HTML]{EAF1F8} 55.00^{\scriptsize\pm 40.41}$ & $\cellcolor[HTML]{F6F9FC} 66.89^{\scriptsize\pm 36.94}$ & $\cellcolor[HTML]{FCFDFE} 69.44^{\scriptsize\pm 31.02}$ & $\cellcolor[HTML]{E9F0F7} 54.18^{\scriptsize\pm 30.27}$ & $\cellcolor[HTML]{F8FAFD} 69.32^{\scriptsize\pm 24.98}$ & $\cellcolor[HTML]{FCFDFE} 69.14^{\scriptsize\pm 26.45}$ & $\cellcolor[HTML]{E9F0F7} 54.20^{\scriptsize\pm \phantom{0}4.33}$ & $\cellcolor[HTML]{FAFBFD} 71.62^{\scriptsize\pm \phantom{0}5.30}$ & $\cellcolor[HTML]{F2F6FB} 75.51^{\scriptsize\pm \phantom{0}7.71}$ \\
KLM~\cite{hendrycks2019_ood_mls} & $\cellcolor[HTML]{F1F5FA} 61.36^{\scriptsize\pm \phantom{0}9.13}$ & $\cellcolor[HTML]{FEFEFF} 80.25^{\scriptsize\pm \phantom{0}8.45}$ & $\cellcolor[HTML]{F0F4FA} 76.51^{\scriptsize\pm \phantom{0}7.33}$ & $\cellcolor[HTML]{E6EEF6} 51.45^{\scriptsize\pm 21.07}$ & $\cellcolor[HTML]{FAFBFD} 72.20^{\scriptsize\pm 20.86}$ & $\cellcolor[HTML]{D4E1F0} 84.90^{\scriptsize\pm \phantom{0}8.74}$ & $\cellcolor[HTML]{E2EAF5} 48.52^{\scriptsize\pm 32.51}$ & $\cellcolor[HTML]{F5F8FB} 64.91^{\scriptsize\pm 32.16}$ & $\cellcolor[HTML]{DAE5F2} 83.59^{\scriptsize\pm 15.78}$ & $\cellcolor[HTML]{EAF0F8} 54.58^{\scriptsize\pm 21.71}$ & $\cellcolor[HTML]{F8FAFD} 70.51^{\scriptsize\pm 22.12}$ & $\cellcolor[HTML]{EAF1F8} 78.68^{\scriptsize\pm 15.51}$ & $\cellcolor[HTML]{E9EFF7} 53.98^{\scriptsize\pm \phantom{0}5.51}$ & $\cellcolor[HTML]{FAFBFD} 71.97^{\scriptsize\pm \phantom{0}6.34}$ & $\cellcolor[HTML]{E4ECF5} 80.92^{\scriptsize\pm \phantom{0}3.98}$ \\
ASH-s~\cite{djurisic2023ash} & $\cellcolor[HTML]{F2F6FB} 62.47^{\scriptsize\pm 17.03}$ & $\cellcolor[HTML]{FEFFFF} 83.09^{\scriptsize\pm 14.80}$ & $\cellcolor[HTML]{EAF0F8} 78.92^{\scriptsize\pm \phantom{0}6.08}$ & $\cellcolor[HTML]{E3ECF5} 49.56^{\scriptsize\pm 27.30}$ & $\cellcolor[HTML]{FBFCFE} 74.63^{\scriptsize\pm 28.06}$ & $\cellcolor[HTML]{D1DFEF} 85.70^{\scriptsize\pm 10.10}$ & $\cellcolor[HTML]{EBF1F8} 55.66^{\scriptsize\pm 38.40}$ & $\cellcolor[HTML]{FAFCFD} 72.99^{\scriptsize\pm 33.88}$ & $\cellcolor[HTML]{F2F6FB} 75.31^{\scriptsize\pm 25.83}$ & $\cellcolor[HTML]{ECF2F8} 56.93^{\scriptsize\pm 28.24}$ & $\cellcolor[HTML]{FCFDFE} 76.21^{\scriptsize\pm 26.17}$ & $\cellcolor[HTML]{F7F9FC} 73.08^{\scriptsize\pm 21.99}$ & $\cellcolor[HTML]{EBF1F8} 56.15^{\scriptsize\pm \phantom{0}5.30}$ & $\cellcolor[HTML]{FCFDFE} 76.73^{\scriptsize\pm \phantom{0}4.44}$ & $\cellcolor[HTML]{EBF1F8} 78.25^{\scriptsize\pm \phantom{0}5.52}$ \\
GradNorm~\cite{huang2021_gradnorm} & $\cellcolor[HTML]{FBFCFD} 73.72^{\scriptsize\pm 13.76}$ & $\cellcolor[HTML]{FFFFFF} 88.56^{\scriptsize\pm \phantom{0}8.46}$ & $\cellcolor[HTML]{FBFCFE} 69.71^{\scriptsize\pm 12.64}$ & $\cellcolor[HTML]{F8FAFD} 69.32^{\scriptsize\pm 18.42}$ & $\cellcolor[HTML]{FEFFFF} 84.27^{\scriptsize\pm 15.39}$ & $\cellcolor[HTML]{F6F9FC} 73.69^{\scriptsize\pm 12.54}$ & $\cellcolor[HTML]{EBF1F8} 56.18^{\scriptsize\pm 42.18}$ & $\cellcolor[HTML]{F5F8FC} 65.91^{\scriptsize\pm 40.84}$ & $\cellcolor[HTML]{FFFFFF} 63.75^{\scriptsize\pm 38.58}$ & $\cellcolor[HTML]{F0F4FA} 60.16^{\scriptsize\pm 31.81}$ & $\cellcolor[HTML]{F8FAFD} 69.87^{\scriptsize\pm 29.69}$ & $\cellcolor[HTML]{FFFFFF} 61.34^{\scriptsize\pm 32.68}$ & $\cellcolor[HTML]{F5F8FB} 64.84^{\scriptsize\pm \phantom{0}8.08}$ & $\cellcolor[HTML]{FCFDFE} 77.15^{\scriptsize\pm 10.96}$ & $\cellcolor[HTML]{FDFEFE} 67.12^{\scriptsize\pm \phantom{0}5.62}$ \\
TempScale~\cite{Gua2017_tempscaling} & $\cellcolor[HTML]{F5F8FC} 66.08^{\scriptsize\pm \phantom{0}9.07}$ & $\cellcolor[HTML]{FFFFFF} 84.48^{\scriptsize\pm \phantom{0}9.03}$ & $\cellcolor[HTML]{E5EDF6} 80.44^{\scriptsize\pm \phantom{0}4.90}$ & $\cellcolor[HTML]{ECF2F8} 56.63^{\scriptsize\pm 21.82}$ & $\cellcolor[HTML]{FDFDFE} 77.54^{\scriptsize\pm 20.43}$ & $\cellcolor[HTML]{CDDCEE} 86.50^{\scriptsize\pm \phantom{0}7.35}$ & $\cellcolor[HTML]{EAF0F8} 54.44^{\scriptsize\pm 33.51}$ & $\cellcolor[HTML]{F9FBFD} 71.28^{\scriptsize\pm 32.94}$ & $\cellcolor[HTML]{E3EBF5} 81.24^{\scriptsize\pm 20.07}$ & $\cellcolor[HTML]{EEF3F9} 58.52^{\scriptsize\pm 22.72}$ & $\cellcolor[HTML]{FCFDFE} 75.53^{\scriptsize\pm 23.07}$ & $\cellcolor[HTML]{ECF2F8} 78.08^{\scriptsize\pm 17.67}$ & $\cellcolor[HTML]{EFF4F9} 58.92^{\scriptsize\pm \phantom{0}5.06}$ & $\cellcolor[HTML]{FCFDFE} 77.21^{\scriptsize\pm \phantom{0}5.51}$ & $\cellcolor[HTML]{E2EAF5} 81.57^{\scriptsize\pm \phantom{0}3.55}$ \\
MSP~\cite{hendrycks2016_ood_msp} & $\cellcolor[HTML]{F7FAFC} 68.81^{\scriptsize\pm \phantom{0}7.46}$ & $\cellcolor[HTML]{FFFFFF} 87.78^{\scriptsize\pm \phantom{0}6.59}$ & $\cellcolor[HTML]{E7EEF7} 79.90^{\scriptsize\pm \phantom{0}4.96}$ & $\cellcolor[HTML]{F0F5FA} 60.65^{\scriptsize\pm 17.82}$ & $\cellcolor[HTML]{FEFFFF} 82.50^{\scriptsize\pm 15.60}$ & $\cellcolor[HTML]{D0DEEE} 85.90^{\scriptsize\pm \phantom{0}6.86}$ & $\cellcolor[HTML]{EDF2F9} 57.20^{\scriptsize\pm 31.21}$ & $\cellcolor[HTML]{FCFDFE} 75.84^{\scriptsize\pm 29.75}$ & $\cellcolor[HTML]{E3ECF5} 81.02^{\scriptsize\pm 19.61}$ & $\cellcolor[HTML]{F2F6FA} 62.36^{\scriptsize\pm 20.67}$ & $\cellcolor[HTML]{FDFEFE} 79.69^{\scriptsize\pm 20.80}$ & $\cellcolor[HTML]{EDF2F9} 77.65^{\scriptsize\pm 17.13}$ & $\cellcolor[HTML]{F2F6FA} 62.26^{\scriptsize\pm \phantom{0}4.87}$ & $\cellcolor[HTML]{FEFEFF} 81.45^{\scriptsize\pm \phantom{0}5.02}$ & $\cellcolor[HTML]{E3ECF5} 81.12^{\scriptsize\pm \phantom{0}3.48}$ \\
ASH-b~\cite{djurisic2023ash} & $\cellcolor[HTML]{FFFFFF} 86.50^{\scriptsize\pm 12.01}$ & $\cellcolor[HTML]{FFFFFF} 95.60^{\scriptsize\pm \phantom{0}6.19}$ & $\cellcolor[HTML]{FEFFFF} 64.45^{\scriptsize\pm 12.66}$ & $\cellcolor[HTML]{FEFEFF} 80.47^{\scriptsize\pm 22.16}$ & $\cellcolor[HTML]{FFFFFF} 92.57^{\scriptsize\pm 13.26}$ & $\cellcolor[HTML]{FEFEFF} 66.32^{\scriptsize\pm 17.46}$ & $\cellcolor[HTML]{FEFFFF} 82.30^{\scriptsize\pm 31.94}$ & $\cellcolor[HTML]{FFFFFF} 86.63^{\scriptsize\pm 31.73}$ & $\cellcolor[HTML]{FFFFFF} 56.08^{\scriptsize\pm 24.90}$ & $\cellcolor[HTML]{FEFEFF} 81.82^{\scriptsize\pm 25.97}$ & $\cellcolor[HTML]{FFFFFF} 88.67^{\scriptsize\pm 23.04}$ & $\cellcolor[HTML]{FFFFFF} 53.50^{\scriptsize\pm 20.24}$ & $\cellcolor[HTML]{FEFFFF} 82.77^{\scriptsize\pm \phantom{0}2.60}$ & $\cellcolor[HTML]{FFFFFF} 90.87^{\scriptsize\pm \phantom{0}4.01}$ & $\cellcolor[HTML]{FFFFFF} 60.09^{\scriptsize\pm \phantom{0}6.25}$ \\
RankFeat~\cite{song2022rankfeat} & $\cellcolor[HTML]{FFFFFF} 92.35^{\scriptsize\pm \phantom{0}6.82}$ & $\cellcolor[HTML]{FEFFFF} 97.69^{\scriptsize\pm \phantom{0}3.73}$ & $\cellcolor[HTML]{FFFFFF} 50.49^{\scriptsize\pm \phantom{0}9.07}$ & $\cellcolor[HTML]{FFFFFF} 89.03^{\scriptsize\pm 13.13}$ & $\cellcolor[HTML]{FFFFFF} 95.33^{\scriptsize\pm 10.32}$ & $\cellcolor[HTML]{FFFFFF} 53.74^{\scriptsize\pm 14.28}$ & $\cellcolor[HTML]{FFFFFF} 92.55^{\scriptsize\pm 16.73}$ & $\cellcolor[HTML]{FEFFFF} 96.19^{\scriptsize\pm 11.37}$ & $\cellcolor[HTML]{FFFFFF} 35.92^{\scriptsize\pm 22.19}$ & $\cellcolor[HTML]{FFFFFF} 93.83^{\scriptsize\pm \phantom{0}7.29}$ & $\cellcolor[HTML]{FEFFFF} 97.21^{\scriptsize\pm \phantom{0}5.52}$ & $\cellcolor[HTML]{FFFFFF} 36.05^{\scriptsize\pm 11.24}$ & $\cellcolor[HTML]{FFFFFF} 91.94^{\scriptsize\pm \phantom{0}2.05}$ & $\cellcolor[HTML]{FEFFFF} 96.60^{\scriptsize\pm \phantom{0}1.06}$ & $\cellcolor[HTML]{FFFFFF} 44.05^{\scriptsize\pm \phantom{0}9.41}$ \\

\bottomrule
\end{tabular}
}
\caption{
Average OOD detection performance measured as FPR at $95\%$ and $99\%$ TPR, reported per method and OOD category (near, far, extreme, and synthetic). 
Results are shown as mean $\pm$ standard deviation over three random seeds, two architectures (ResNet18 and CCT), and four tasks (Almond, Wheat, Kernels, and Food-grade).
The arrow ($\downarrow$) indicates that lower values are better. 
Cells are color-coded from \textcolor{best_blue}{\textbf{blue}} (high performance) to white (low performance). 
Additionally, \textbf{best} and \underline{second-best} results in each column are highlighted in bold and underlined, respectively. 
All values are reported as percentages, and methods are sorted based on their average FPR99 score.
}
\label{tab:ood_bm_avg_performance_fpr95_fpr99}
\end{table}
\end{landscape}

\begin{figure*}
\centering
\includegraphics[width=\linewidth]{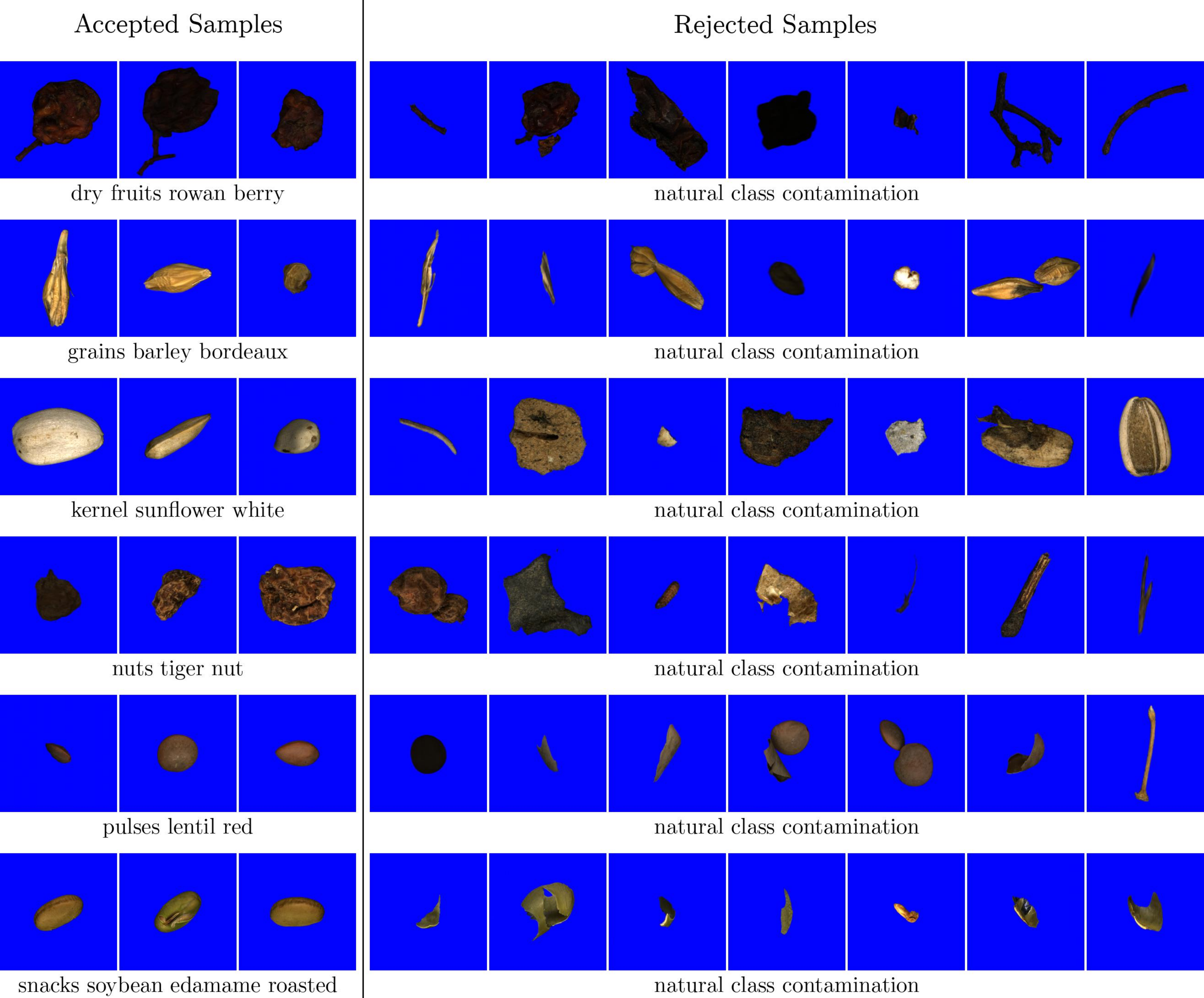}
\caption{
Illustrative examples of the dataset's quality control process across six representative classes. 
Each row displays images from a single class, beginning with three accepted reference samples (left), followed by seven examples of rejected items (right) that were flagged during the cleaning process.
}
\label{fig:contamination_samples}
\end{figure*}

\begin{figure*}
\centering
\includegraphics[width=\linewidth]{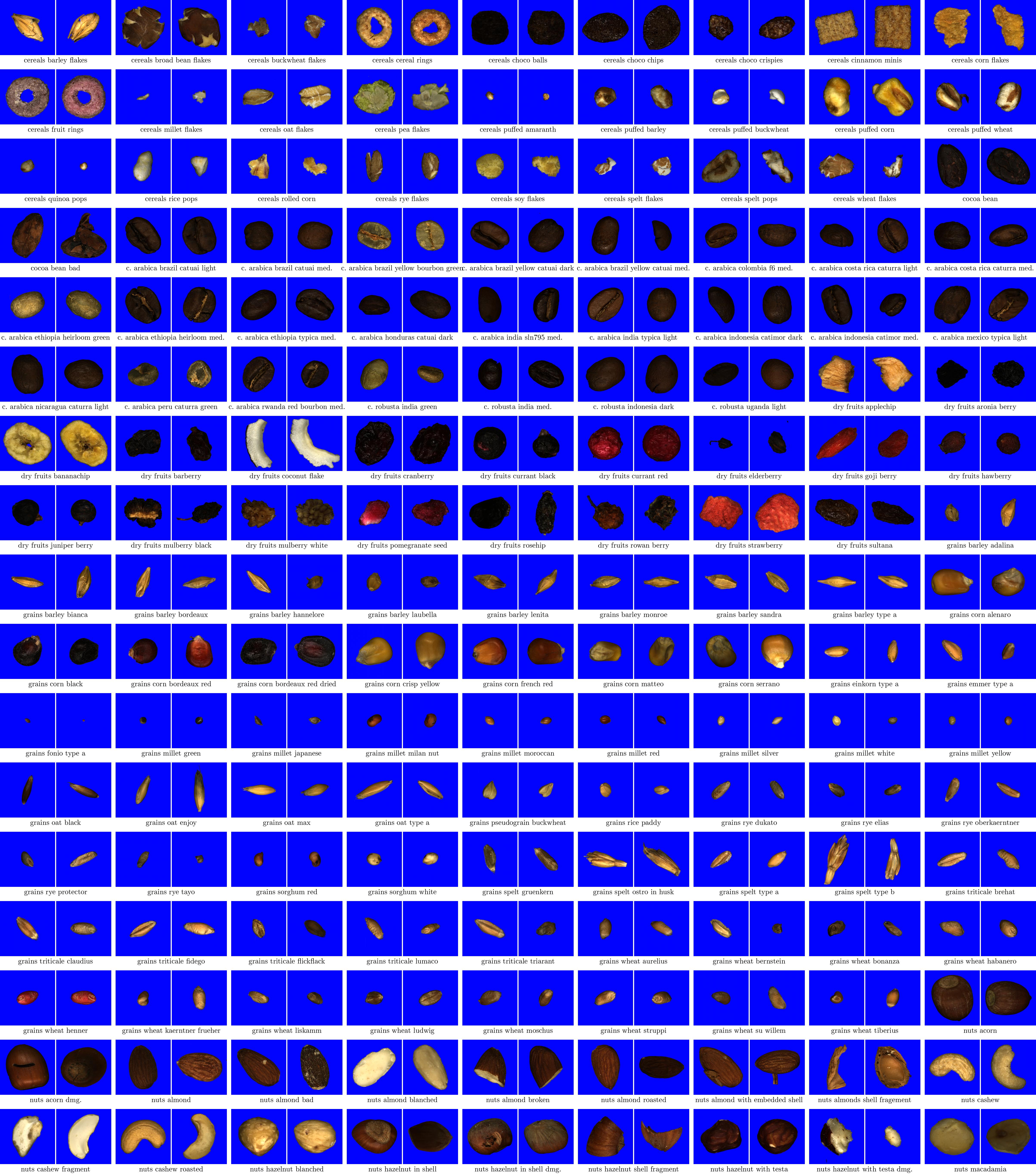}
\caption{
Samples for each class of the ICONIC-444 dataset (1/3), best viewed on screen.
}
\label{fig:iconic_1_3}
\end{figure*}

\begin{figure*}
\centering
\includegraphics[width=\linewidth]{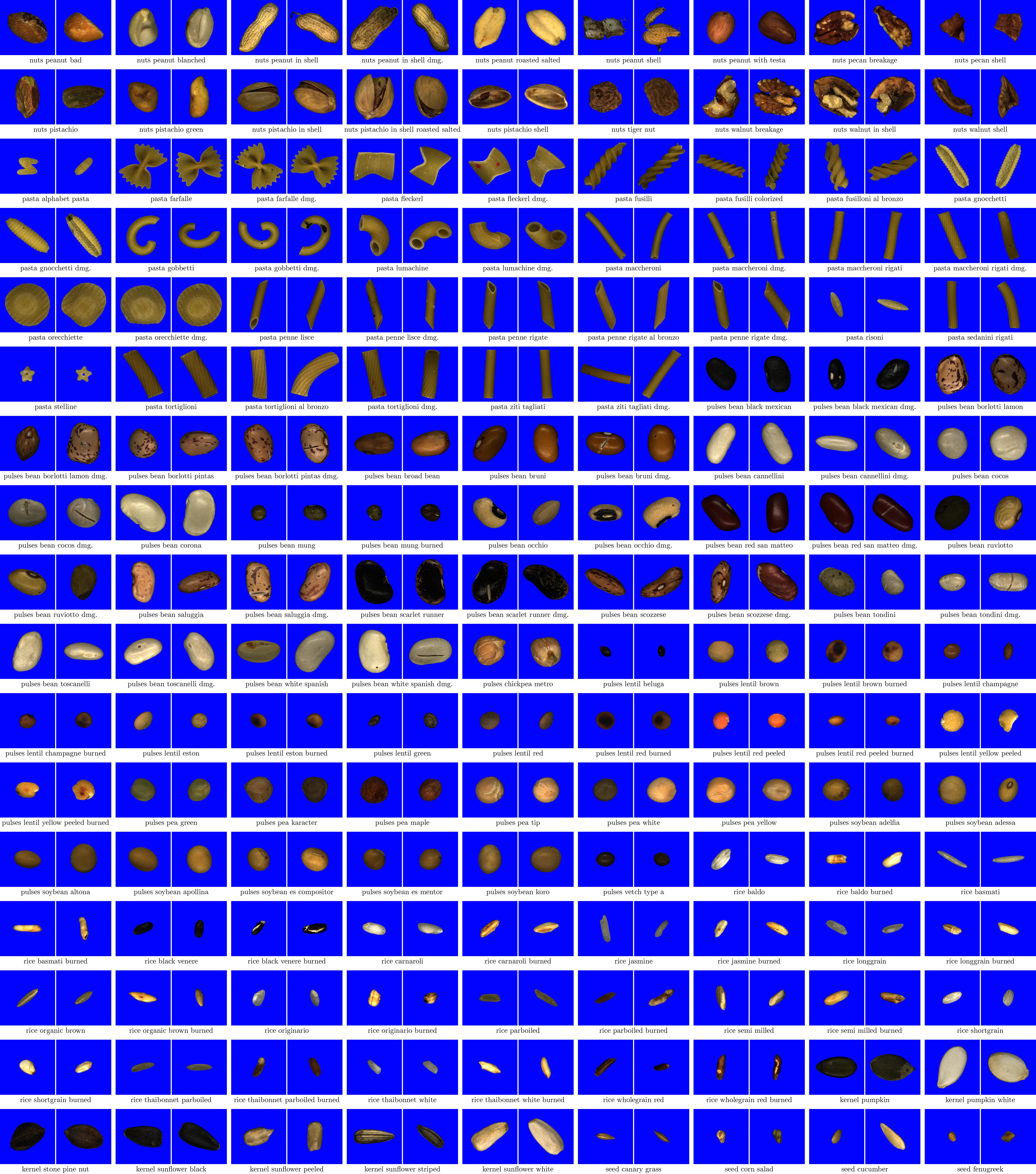}
\caption{
Samples for each class of the ICONIC-444 dataset (2/3), best viewed on screen.
}
\label{fig:iconic_2_3}
\end{figure*}

\begin{figure*}
\centering
\includegraphics[width=\linewidth]{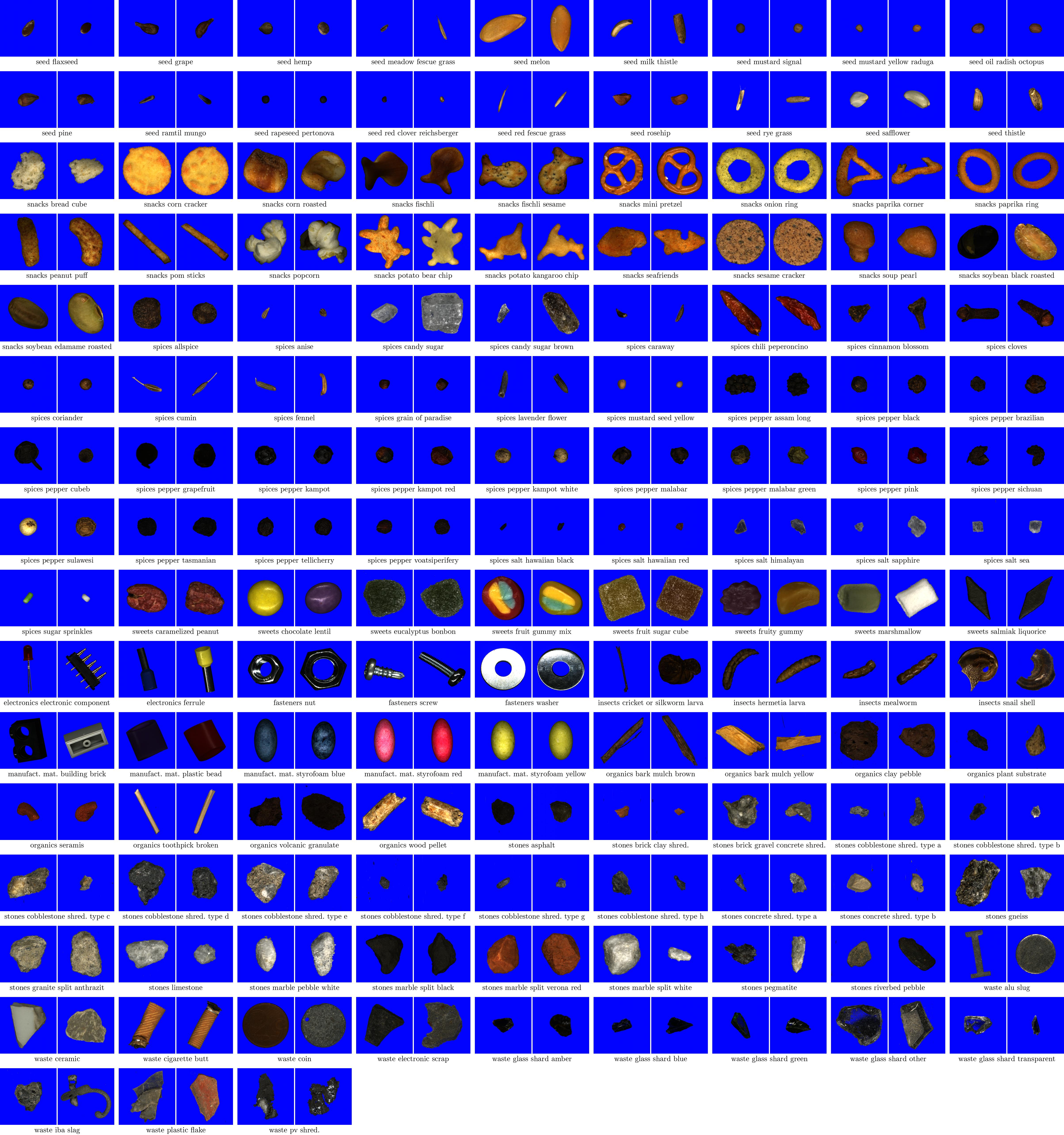}
\caption{
Samples for each class of the ICONIC-444 dataset (3/3), best viewed on screen.
}
\label{fig:iconic_3_3}
\end{figure*}

\begin{figure*}[ht]
    \centering
    \begin{subfigure}[b]{0.48\linewidth}
        \centering
        \includegraphics[width=\linewidth]{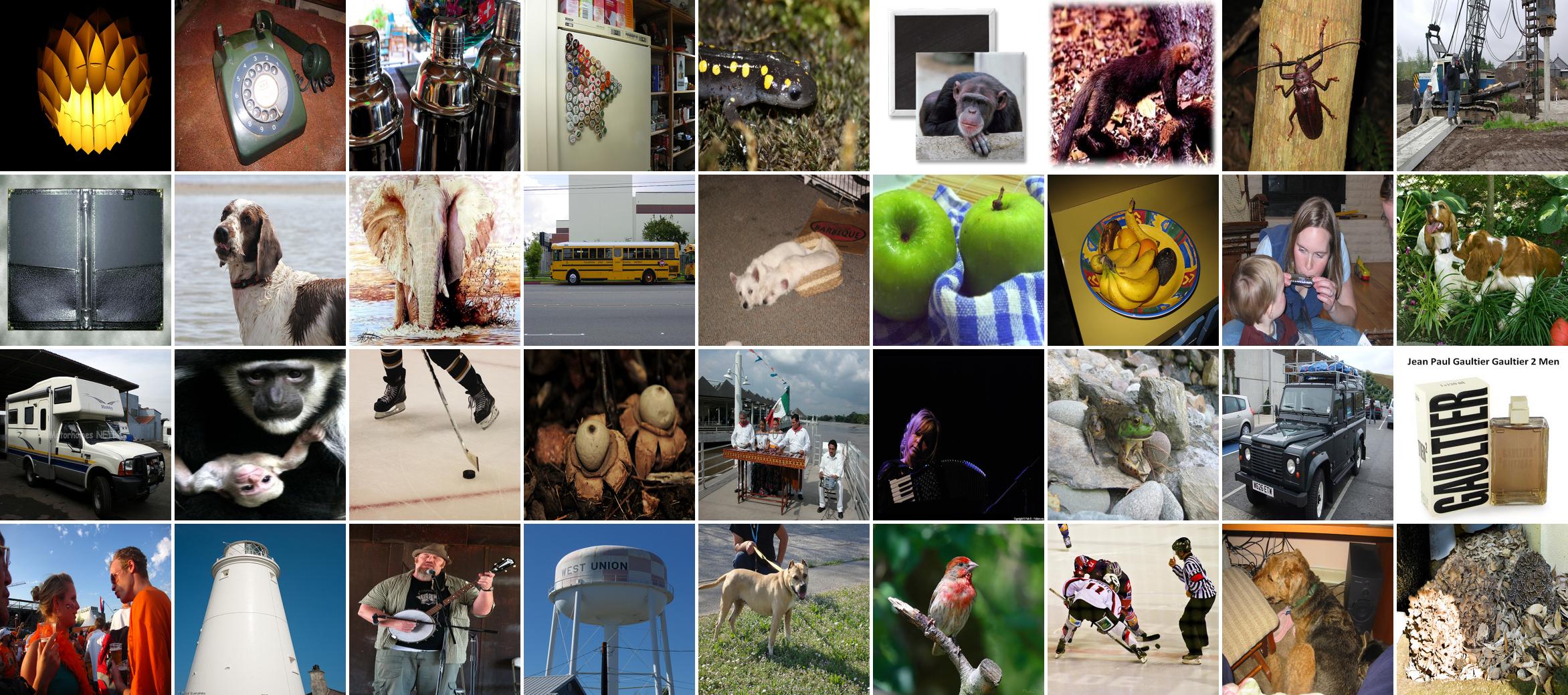}
        \caption{ImageNet~\cite{deng2009_imagenet}}
    \end{subfigure}
    \hfill
    \begin{subfigure}[b]{0.48\linewidth}
        \centering
        \includegraphics[width=\linewidth]{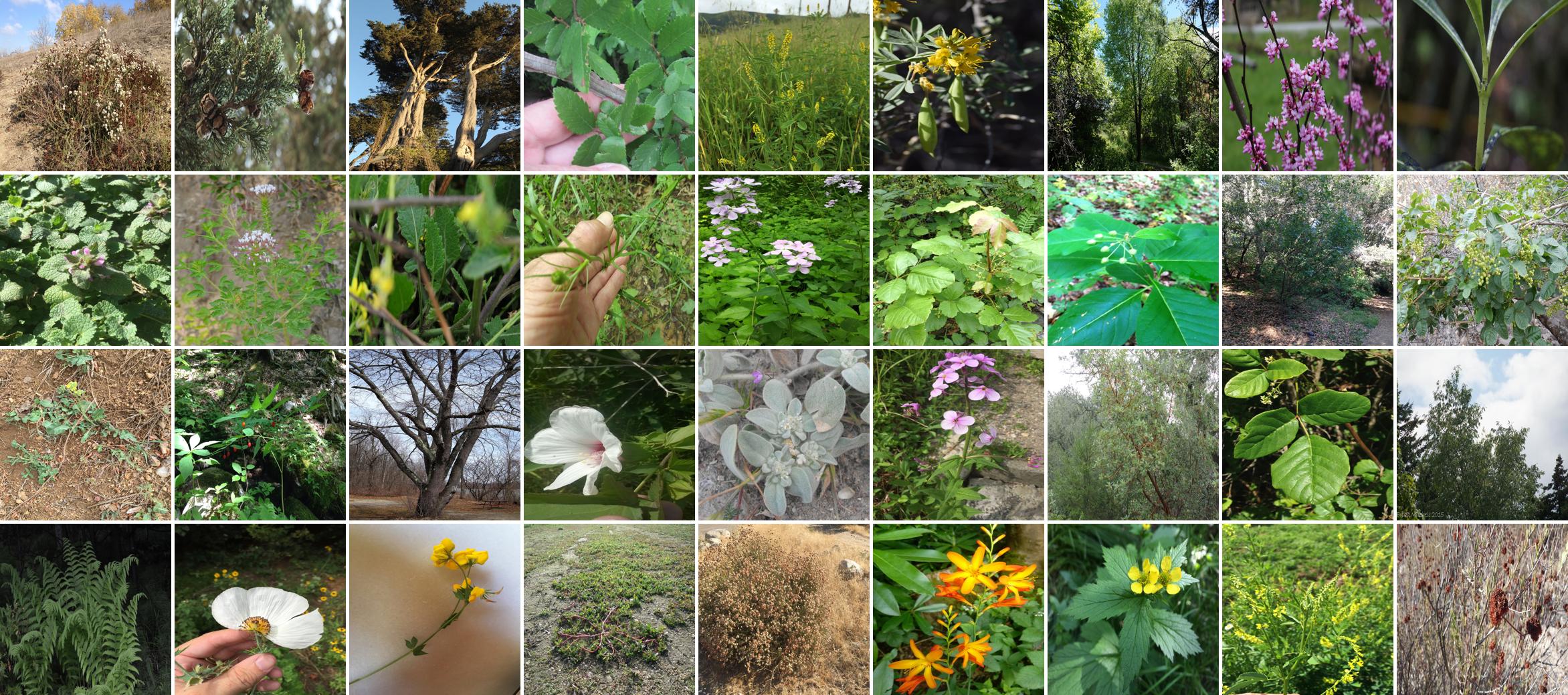}
        \caption{iNaturalist~\cite{Horn_2018_inat}}
    \end{subfigure}
    
    \vspace{10pt} %

    \begin{subfigure}[b]{0.48\linewidth}
        \centering
        \includegraphics[width=\linewidth]{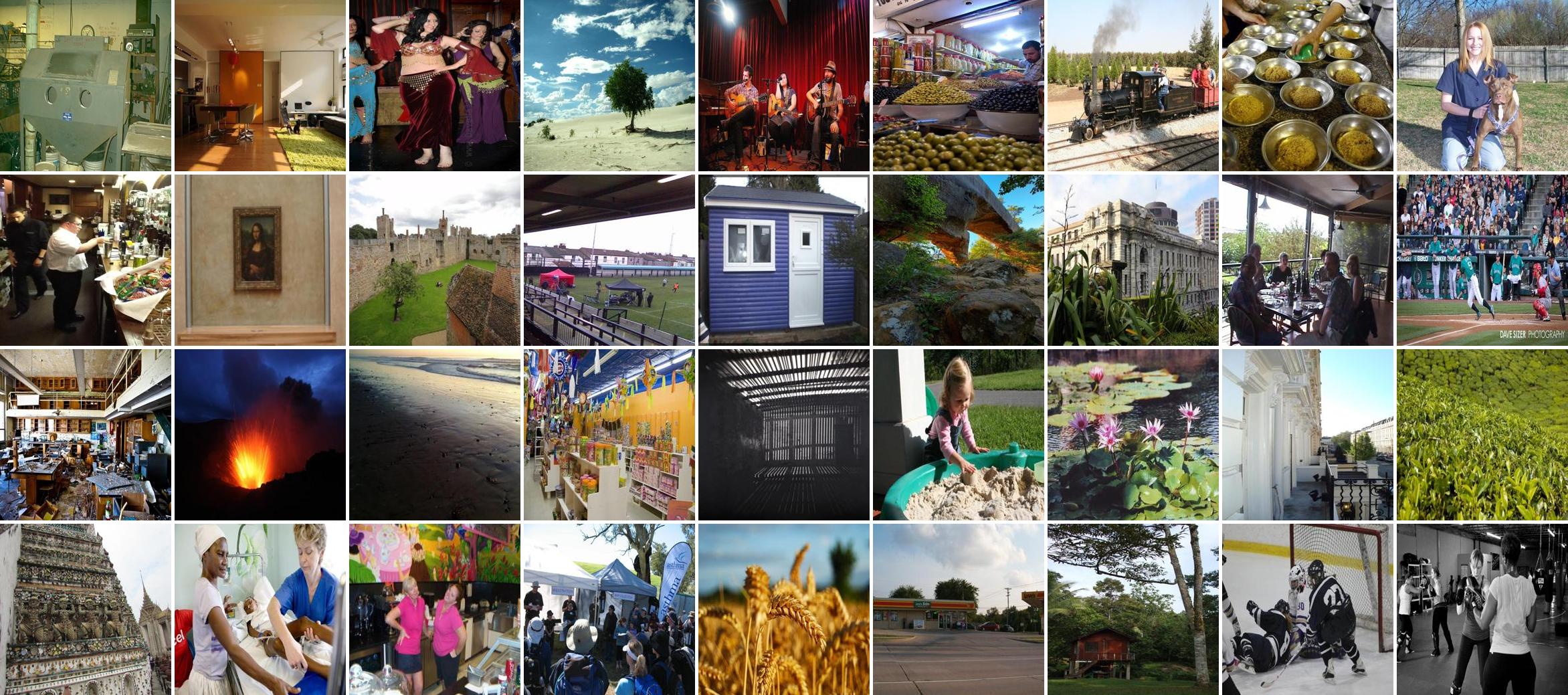}
        \caption{Places365~\cite{zhou2017places}}
    \end{subfigure}
    \hfill
    \begin{subfigure}[b]{0.48\linewidth}
        \centering
        \includegraphics[width=\linewidth]{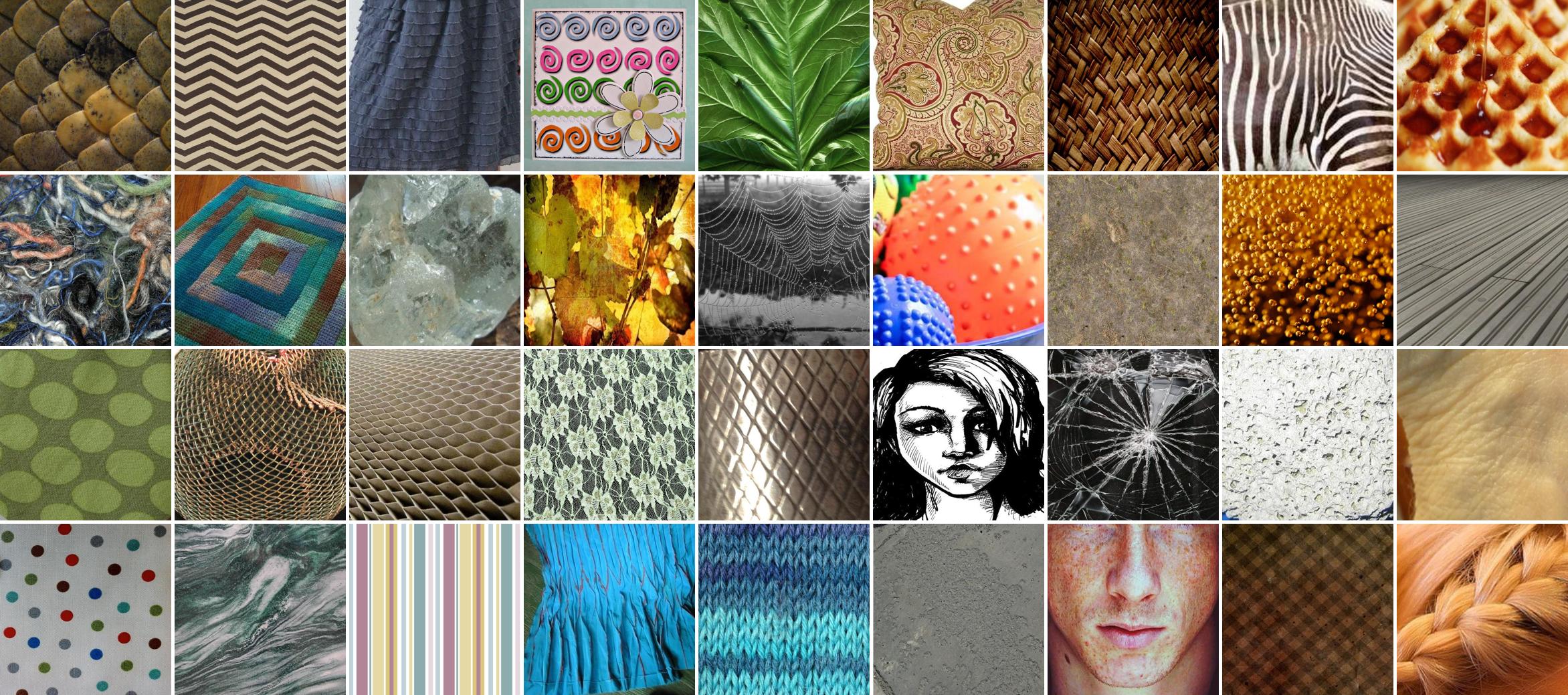}
        \caption{Textures~\cite{cimpoi14dtd}}
    \end{subfigure}

    \caption{Randomly selected samples (36 per dataset) from the Extreme OOD set.}
    \label{fig:extreme_ood_samples}
\end{figure*}

\begin{figure*}
\centering
\includegraphics[width=\linewidth]{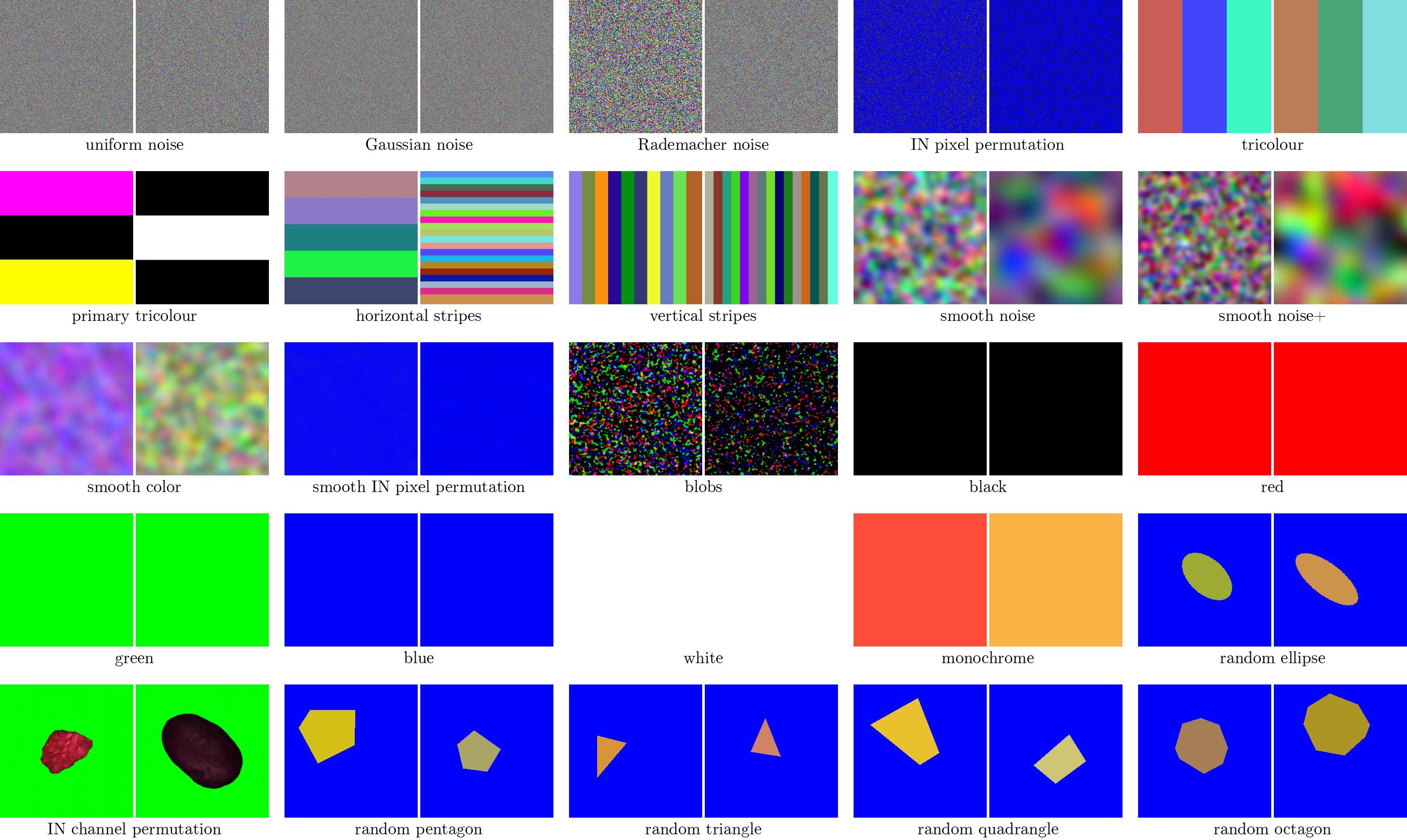}
\caption{
Examples of two randomly selected samples from each of the 25 synthetic OOD classes.
}
\label{fig:synthetic_ood_samples}
\end{figure*}

\end{document}